\documentclass[lettersize,journal]{IEEEtran}
\usepackage{amsmath,amsfonts}
\usepackage{algorithmicx}
\usepackage{algpseudocode}
\usepackage{algorithm}
\usepackage{array}
\usepackage{textcomp}
\usepackage{stfloats}
\usepackage{url}
\usepackage{verbatim}
\usepackage{graphicx}
\usepackage{wrapfig}

\usepackage[numbers]{natbib}

\usepackage[utf8]{inputenc} %
\usepackage[T1]{fontenc}    %
\usepackage{url}            %
\usepackage{booktabs}       %
\usepackage{amsfonts}       %
\usepackage{nicefrac}       %
\usepackage{microtype}      %

\usepackage{color,xcolor}
\usepackage{epsfig}
\usepackage{graphicx}
\usepackage{subfiles}

\usepackage{adjustbox}
\usepackage{array}
\usepackage{booktabs}
\usepackage{colortbl}
\usepackage{float,wrapfig}
\usepackage{hhline}
\usepackage{multirow}
\usepackage{subcaption} %
\usepackage[labelfont={bf},labelsep={period},font={small}]{caption}

\let\llncssubparagraph\subparagraph
\let\subparagraph\paragraph
\usepackage[compact]{titlesec}
\let\subparagraph\llncssubparagraph

\usepackage{amsmath,amsfonts,amssymb}
\usepackage{bm}
\usepackage{nicefrac}
\usepackage{microtype}

\usepackage{changepage}
\usepackage{extramarks}
\usepackage{fancyhdr}
\usepackage{lastpage}
\usepackage{setspace}
\usepackage{soul}
\usepackage{xspace}

\usepackage{enumerate}

\usepackage{times}

\usepackage{pbox}
\usepackage{footnote}
\usepackage{tablefootnote}

\usepackage{paralist}

\usepackage{enumitem}
\setitemize{noitemsep,topsep=0pt,parsep=0pt,partopsep=0pt}

\definecolor{mydarkgreen}{rgb}{0.02,0.6,0.02}

\usepackage{tabu}
\usepackage{pifont}
\usepackage[normalem]{ulem}
\usepackage{nccmath}
\usepackage{dsfont}
\usepackage{makecell}
\usepackage{scrextend}
\usepackage{hyperref}

\newcommand{\myparagraph}[1]{\paragraph{#1}}

\newcommand{\ignorethis}[1]{}

\makeatletter
\DeclareRobustCommand\onedot{\futurelet\@let@token\@onedot}
\def\@onedot{\ifx\@let@token.\else.\null\fi\xspace}

\def\eg{\emph{e.g}\onedot} 
\def\ie{\emph{i.e}\onedot} 
 
\def\etc{\emph{etc}\onedot} \def\vs{\emph{vs}\onedot}
 
\def\etal{\emph{et al}\onedot}
\makeatother

\makeatletter
\newcommand\footnoteref[1]{\protected@xdef\@thefnmark{\ref{#1}}\@footnotemark}
\makeatother

\definecolor{mydarkblue}{rgb}{0,0.08,1}

\definecolor{mydarkred}{rgb}{0.8,0.02,0.02}
\definecolor{mydarkorange}{rgb}{0.40,0.2,0.02}
\definecolor{mypurple}{RGB}{111,0,255}
\definecolor{myred}{rgb}{1.0,0.0,0.0}
\definecolor{mygold}{rgb}{0.75,0.6,0.12}
\definecolor{mydarkgray}{rgb}{0.66, 0.66, 0.66}
\definecolor{mygray}{gray}{0.9}

\def\method{MCUNet\xspace}

\begin{document}

\title{Tiny Machine Learning: Progress and Futures}

\author{
Ji Lin
\space
Ligeng Zhu
\space
Wei-Ming Chen
\space
Wei-Chen Wang
\space
Song Han \\
Massachusetts Institute of Technology
\\
\url{https://tinyml.mit.edu}
\thanks{This paper is published by IEEE Circuits and Systems Magazine. © 2023 IEEE. Personal use of this material is permitted.  Permission from IEEE must be obtained for all other uses, in any current or future media, including reprinting/republishing this material for advertising or promotional purposes, creating new collective works, for resale or redistribution to servers or lists, or reuse of any copyrighted component of this work in other works.}
}



\maketitle

\begin{abstract}
Tiny Machine Learning (TinyML) is a new frontier of machine learning. By squeezing deep learning models into billions of IoT devices and microcontrollers (MCUs), we expand the scope of AI applications and enable ubiquitous intelligence. However, TinyML is challenging due to hardware constraints: the tiny memory resource makes it difficult to hold deep learning models designed for cloud and mobile platforms. There is also limited compiler and inference engine support for bare-metal devices. 
Therefore, we need to co-design the algorithm and system stack to enable TinyML.
In this review, we will first discuss the definition, challenges, and applications of TinyML. 
We then survey the recent progress in TinyML and deep learning on MCUs. Next, we will introduce MCUNet, showing how we can achieve ImageNet-scale AI applications on IoT devices with \emph{system-algorithm co-design}. We will further \emph{extend} the solution from \emph{inference to training} and introduce tiny on-device training techniques. 
Finally, we present future directions in this area. Today's ``large'' model might be tomorrow's ``tiny'' model. The scope of TinyML should evolve and adapt over time.

\end{abstract}


\section{Overview of Tiny Machine Learning}
\label{intro}
\begin{figure}
    \centering
    \includegraphics[width=1.0\linewidth]{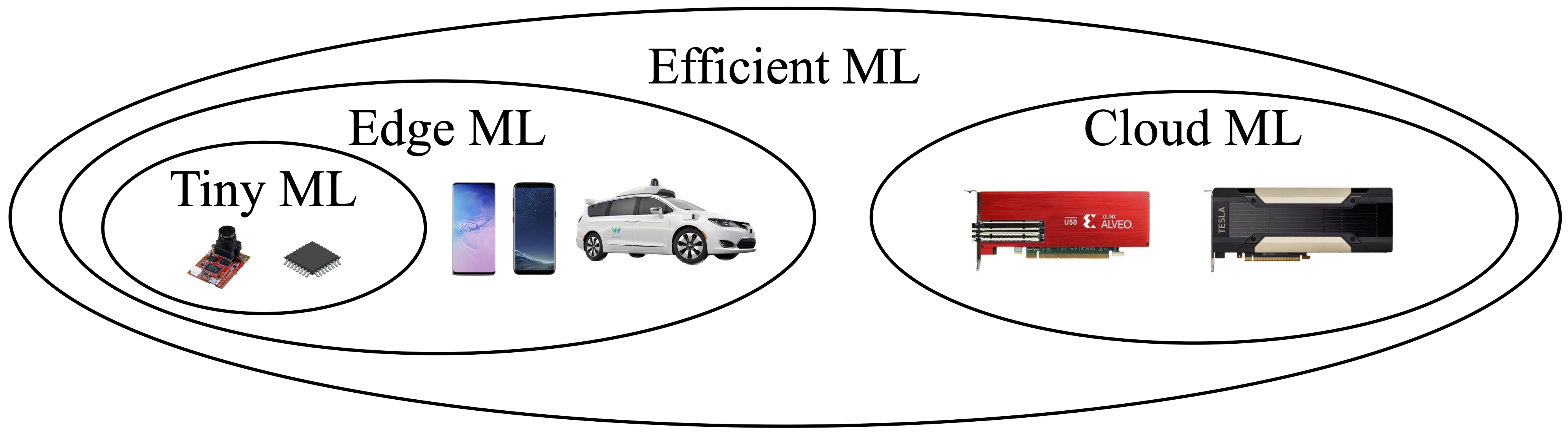}
    \caption{
    Efficiency is critical for CloudML, EdgeML, and TinyML. CloudML targets high-throughput accelerators like GPUs, while EdgeML focuses on portable devices like mobile phones. TinyML further pushes the efficiency boundary, enabling powerful ML models to run on ultra-low-power devices such as microcontrollers. 
    }
    \label{fig:TinyML_concept}
\end{figure}

Machine learning (ML) has made significant impacts on various fields, including vision, language, and audio. However, state-of-the-art models often come at the cost of high computation and memory, making them expensive to deploy.
To address this, researchers have been working on efficient algorithms, systems, and hardware to reduce the cost of machine learning models in various deployment scenarios.
There are two main subdomains of efficient ML: EdgeML and CloudML (Figure~\ref{fig:TinyML_concept}). While CloudML focuses on improving latency and throughput on cloud servers, EdgeML focuses on improving energy efficiency, latency, and privacy on edge devices. These two domains also intersect in areas such as hybrid inference~\cite{liu2020datamix, singh2019detailed}, over-the-air (OTA) updates, and federated learning between the edge and cloud~\cite{konevcny2016federated}.
In recent years, there has been significant progress in extending the scope of EdgeML to ultra-low-power devices such as IoT devices and microcontrollers, known as \textbf{TinyML}.

TinyML has several key advantages. It enables machine learning using only a few hundred kilobytes of memory which greatly reduces the cost. With billions of IoT devices producing more and more data in our daily lives, there is a growing need for low-power, always-on, on-device AI. By performing on-device inference near the sensor, TinyML enables better responsiveness and privacy while reducing the energy cost associated with wireless communication. On-device processing of data can be beneficial for applications where real-time decision-making is crucial, such as autonomous vehicles.

In addition to  inference, we push the frontier of TinyML to enable on-device training on IoT devices. Itrevolutionizes EdgeAI through continuous and lifelong learning. Edge device can finetune the model on itself rather than transmitting data to cloud servers, which protects privacy. On-device learning has numerous benefits and a variety of applications. For example, home cameras can continuously recognize new faces, and email clients can gradually improve their prediction by updating customized language models. 
It also enables IoT applications that do not have a physical connection to the internet to adapt to the environment, such as precision agriculture and ocean sensing.

In this review, we will first discuss the definition and challenges of TinyML, analyzing why we can't directly scale mobile ML or cloud ML models for tinyML. Then we delve into the importance of system-algorithm co-design in TinyML. We will then survey recent literature and the progress of the field, presenting a holistic  survey and comparison in Tables~\ref{tab:open_track} and \ref{tab:closed_track}. Next, we will introduce our TinyML project, MCUNet, which combines efficient system and algorithm design to enable TinyML for both inference to training. Finally, we will discuss several emerging topics for future research directions in the field.

\begin{figure}
    \centering
    \includegraphics[width=0.85\linewidth]{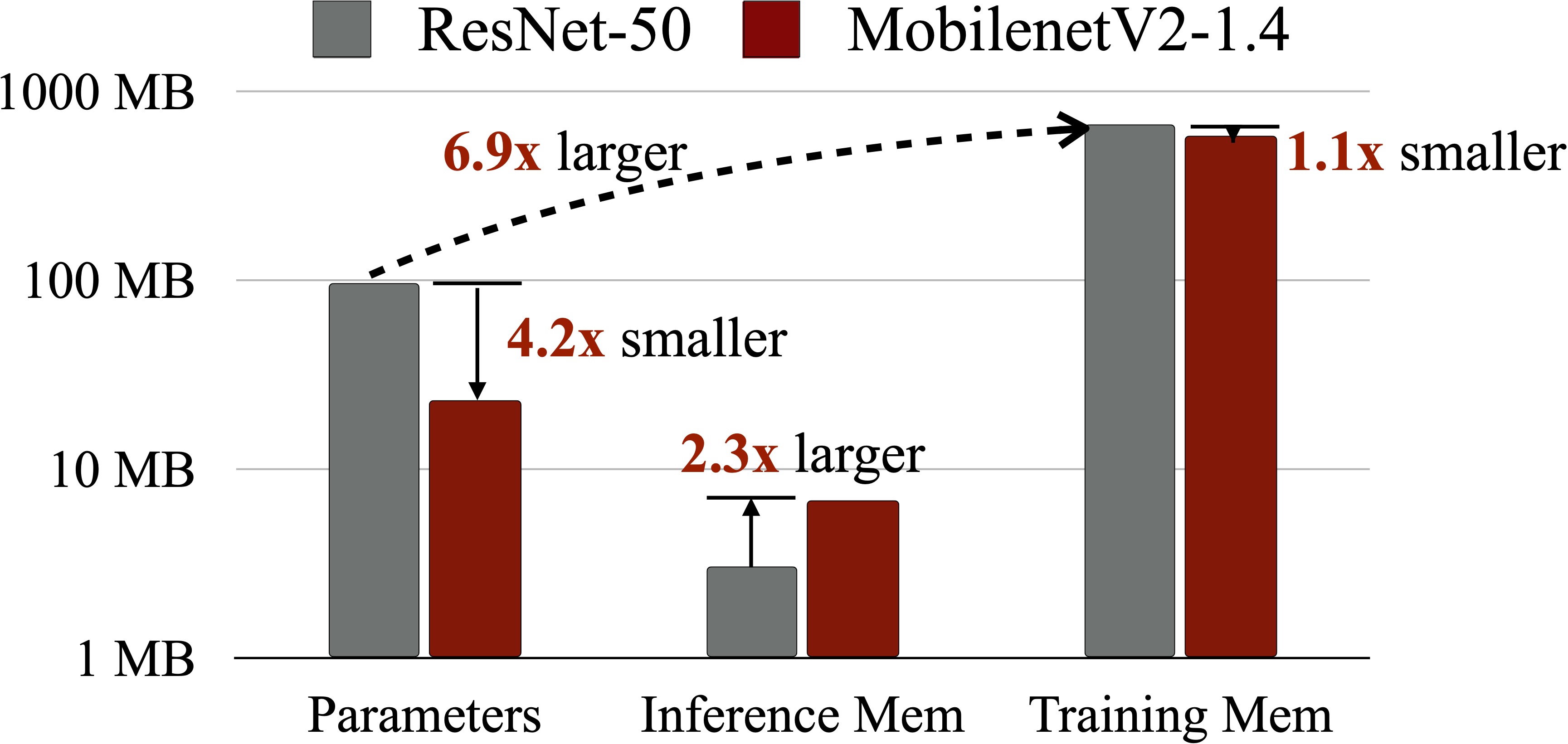}
    \caption{
        We can't directly scale mobile ML or cloud ML models for TinyML. MobilenetV2~\cite{sandler2018mobilenetv2} with a width of 1.4 was used for the experiments. The batch size was set to 1 for inference and 8 for training. While MobilenetV2 reduces the number of parameters by 4.2$\times$ compared to ResNet, the peak memory usage increases by 2.3$\times$ for inference and only improves by 1.1$\times$ for training. Additionally, the total required training memory is 6.9$\times$ larger than the memory needed for inference. These results demonstrate the significant memory bottleneck for TinyML, and the bottleneck is the activation memory, not the number of parameters. 
    }
    \label{fig:memory_comparison}
\end{figure}

\begin{table*}[t]
\caption{\textbf{Left}: Microcontrollers have 3 orders of magnitude \textit{less} memory and storage compared to mobile phones, and 5-6 orders of magnitude less than cloud GPUs. The extremely limited memory makes deep learning deployment difficult. \textbf{Right}: The peak memory and storage usage of widely used deep learning models. ResNet-50 exceeds the resource limit on microcontrollers by $100\times$, MobileNet-V2 exceeds by $20\times$. Even the int8 quantized MobileNetV2 requires $5.3\times$ larger memory and can't fit a microcontroller. }
    \label{tab:hardware_stats}
    \centering
    \includegraphics[width=\textwidth]{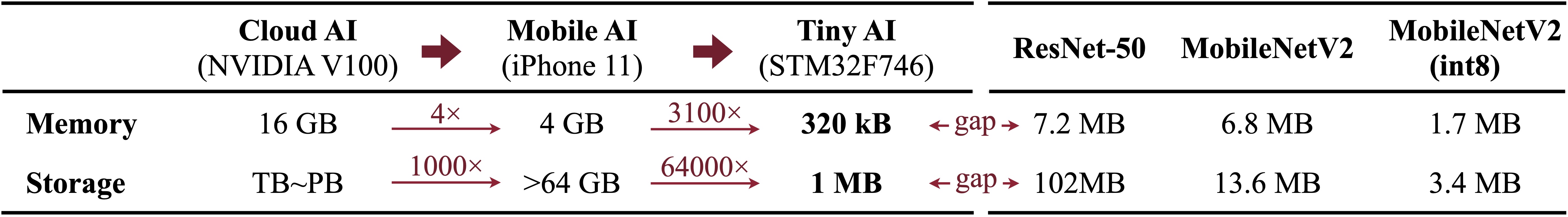}
\end{table*}

\subsection{Challenges of TinyML}
The success of deep learning models often comes at the cost of high computation, which is not feasible for use in TinyML applications due to the strict resource constraints of devices such as microcontrollers. Deploying and training AI models on MCU is extremely hard: No DRAM, no operating systems (OS), and strict memory constraints (SRAM is smaller than 256kB, and FLASH is read-only). The available resources on these devices are orders of magnitude smaller than those available on mobile platforms (see Table~\ref{tab:hardware_stats}). Previous work in the field has either (I) focused on reducing model parameters without addressing the real bottleneck of activations, or (II) only optimized operator kernels without considering improving the network architecture design.  Neither of which considers the problem from a co-design perspective, and this has led to less optimal solutions for TinyML applications. We observe several unique challenges of TinyML and postulate how they might be overcome:

\subsubsection{Models designed for mobile platforms does not fit TinyML}
There has been a lot of effort optimizing deep learning models for mobile platforms like MobileNets~\cite{howard2017mobilenets, sandler2018mobilenetv2} and ShuffleNet~\cite{ma2018shufflenet}. However, since mobile devices have sufficient memory resources (Table~\ref{tab:hardware_stats}), the model designs focus on parameters/FLOPs/latency reduction but not peak memory usage. As shown in Figure~\ref{fig:memory_comparison} Left and Middle, comparing two models with the same level of ImageNet accuracy, MobileNetV2-1.4 has 4.2$\times$ smaller model size compared to ResNet-50, but its peak memory even larger by 2.3$\times$. Using MobileNet designs does not adequately address the SRAM limit, instead, it actually makes the situation even worse compared to ResNet. Therefore, we need to rethink the model design principles for TinyML.

\subsubsection{Directly adapting models for inference does not work for tiny training.}
Training poses an even greater challenge in terms of resource constraints, as intermediate activations must be stored in order to compute backward gradients. When moving from inference to training with full backpropagation, the required memory increases by a factor of 6.9. As shown in Figure~\ref{fig:memory_comparison}, the training memory requirements of MobileNets are not much better than ResNets (improved by only 10\%). Tiny IoT devices such as microcontrollers typically have a limited SRAM size, such as 256KB, which is barely enough for the inference of deep learning models, let alone training. Previous work in the cloud and mobile AI has focused on reducing FLOPs~\cite{howard2017mobilenets, sandler2018mobilenetv2, zhang2018shufflenet} or only optimizing inference memory~\cite{lin2020mcunet, lin2021mcunetv2}. However, even using memory-efficient inference models such as MCUNet~\cite{lin2020mcunet} to bridge the three orders of magnitude gap, training is still too expensive for tiny platforms. If we follow conventional full model update schemes, the model must be scaled down significantly to fit within the tight memory constraints, resulting in low accuracy. This highlights the need to redesign backpropagation schemes and investigate new learning algorithms to reduce the main activation memory bottleneck and enable fast and accurate training on tiny devices. In Section~\ref{tiny_training}, we will discuss this issue in detail and introduce the concept of sparse layer and sparse tensor updates.

\subsubsection{Co-design is necessary for TinyML}
Co-design is necessary for TinyML because it allows us to fully customize the solutions that are optimized for the unique constraints of tiny devices. Previous neural architectures like MobileNets~\cite{howard2017mobilenets, sandler2018mobilenetv2}, and ResNets~\cite{he2016deep} are  designed for mobile/cloud scenarios but not well-suited for tiny hardware. Therefore, we need to design neural architectures that are suitable for TinyML applications. On the other hand, existing deep training frameworks are optimized for cloud servers and lack support for memory-efficient forward and backward, thus cannot fit into tiny devices. The huge gap (>1000$\times$) between the resources of tiny IoT devices and the requirements of current frameworks prohibits the usage. To address these challenges, it is necessary to develop algorithms, systems, and training techniques that are specifically tailored to the settings of these tiny platforms. 


\subsection{Applications of TinyML}
By democratizing costly deep learning models to IoT devices, TinyML has many practical applications. Some example applications include: 

\begin{itemize}
    \item Personalized healthcare: TinyML can allow wearable devices, such as smartwatches, to continuously track the activities and oxygen saturation status of the user in order to provide health suggestions~\cite{tsoukas2021healthcare, rana2022cough, dsouza2022healthcare, shumba2022healthcare}. Body pose estimation is also a crucial application for elderly healthcare~\cite{vuletic2021healthcare}.
    \item Wearable applications: TinyML can assist people with wearable or IoT devices for speech applications, \eg, keyword spotting, automatic speech recognition, and speaker verification~\cite{wong2020tinyspeech, mazumder2021kws, hardy2021voice}.
    \item Smart home: TinyML can enable object detection, image recognition, and face detection on IoT devices to build smart environments, such as smart homes and hospitals~\cite{lu2020edgecameras, giordano2020facedetection, luukkonen2021cough, mohan2021facemaskdetection, wong2020attendnets}.
    \item Human-machine interface: TinyML can enable human-machine interface applications, like hand gesture recognition~\cite{benatti2019handgesture, moin2021handgesture, zhou2021handgesture, bian2021handgesture}. TinyML is also capable of predicting and recognizing sign languages~\cite{paul2020signlanguage}. 
    \item Smart vehicle and transportation: TinyML can perform object detection, lane detection, and decision making without a cloud connection, achieving high-accuracy and low-latency results for autonomous driving scenarios~\cite{de2021autonomous, roshan2021autonomous, bao2021vehicularIoT}.
    \item Anomaly detection: TinyML can equip robots and sensors with the capability to perform anomaly detection to reduce human efforts~\cite{ying2021manufacturing, siang2021anomaly, roth2022anomaly}.
    \item Ecology and agriculture: TinyML can also help with ecological, agricultural, environmental, and phenomics applications so as to conserve endangered species or forecast weather activities~\cite{alongi2020environment, vuppalapati2020agriculture1, vuppalapati2020agriculture2, nakhle2021phenomics, curnick2022smallsats, nicolas2022agriculture}.
\end{itemize}

Overall, the potential applications of TinyML are diverse and numerous, and will expand as the field continues to advance.

\begin{table*}[!t]
    \setlength{\tabcolsep}{4.7pt}
    \caption{Specification and performance comparison of recent progress on TinyML research targeting microcontrollers.}
    \label{tab:open_track}
    \centering
    \small{
    {\renewcommand{\arraystretch}{1.4}%
    \begin{tabular}{l|cccccc}
    \toprule
     & \textbf{\makecell{CMSIS-NN\\ arXiv’18~\cite{lai2018cmsis}}} & \textbf{\makecell{X-Cube-AI\\~\cite{X-Cube-AI}}} & \textbf{\makecell{MicroTVM\\~\cite{microTVM}}} & \textbf{\makecell{Liberis et al.\\ MLSys’20~\cite{liberis2019neural}}} & \textbf{\makecell{Rusci et al.\\ MLSys’20~\cite{rusci2019memory}}} & \textbf{\makecell{CMix-NN\\ TCAS’20~\cite{capotondi2020cmix}}} \\  
    \midrule 
    \makecell[l]{On-Device Training\\ or Inference} & Inference & Inference & Inference & Inference & Inference & Inference \\

    Measured Device & \makecell{MCU\\ (STM32H743)} & \makecell{MCU\\ (STM32H743)} & \makecell{MCU\\ (STM32F746)} & \makecell{MCU\\ (STM32F767)} & \makecell{MCU\\ (STM32H743)} & \makecell{MCU\\ (STM32H743)} \\

    Dataset & ImageNet & ImageNet & CIFAR-10 & VWW & ImageNet & ImageNet \\
    
    Model & MobileNetV1 & MobileNetV1 & SmallCifar & MobileNetV1 & MobileNetV1 & MobileNetV1 \\

    Input Resolution & 192 & 192 & 32 & 96 & 224 & 192 \\

    Width Multiplier & 0.5 & 0.5 & 1.0 & 0.25 & 0.75 & 0.5 \\

    Data Bitwidth & INT8 & INT8 & INT8 & INT8 & Mixed & INT8 \\
    
    Latency & 510 ms$^1$ & 437 ms$^1$ & 157 ms$^3$ & 1325 ms & 1860 ms & 677 ms \\
    
    Peak Memory & < 1 MB$^2$ & < 1 MB$^2$ & 144 KB$^3$ & 55 KB & < 512 KB$^2$ & < 512 KB$^2$ \\
    
    Flash Usage & 1.4 MB$^1$ & 1.4 MB$^1$ & < 1 MB$^2$ & < 2 MB$^2$ & 2 MB & 1.4 MB \\
    
    Energy Consumption & 135 mJ$^1$ & 115 mJ$^1$  & - & 735 mJ & 491 mJ$^4$ & 179 mJ$^4$  \\
    
    Top-1 Accuracy & 59.5\%$^1$ & 59.5\%$^1$ & - & $\sim$76\% & 68.2\% & 62.9\% \\
    \midrule \midrule

     & \textbf{\makecell{MCUNetV1\\ NeurIPS'20~\cite{lin2020mcunet}}} & \textbf{\makecell{TF-Lite Micro\\ MLSys'21~\cite{david2021_tflitemicro}}} & \textbf{\makecell{MicroNets\\ MLSys’21~\cite{banbury2021micronets}}} & \textbf{\makecell{MCUNetV2\\ NeurIPS’21~\cite{lin2021mcunetv2}}} & \textbf{\makecell{TinyOps\\ CVPRW'22~\cite{Sadiq2022TinyOps}}} & \textbf{\makecell{TinyMaix\\~\cite{TinyMaix}}}\\  
    \midrule 
    \makecell[l]{On-Device Training\\ or Inference} & Inference & Inference & Inference & Inference & Inference & Inference \\

    Measured Device & \makecell{MCU\\ (STM32H743)} & \makecell{MCU\\ (STM32F746)} & \makecell{MCU\\ (STM32F746)} & \makecell{MCU\\ (STM32H743)} & \makecell{MCU\\ (STM32F746)} & \makecell{MCU\\ (STM32H750)} \\

    Dataset & ImageNet & ImageNet & VWW & ImageNet & ImageNet & VWW \\
    
    Model & MCUNet & MobileNetV2 & \makecell{MicroNet-\\VWW-1} & MCUNet & MNASNet & MobileNetV1 \\

    Input Resolution & 160 & 64 & 160 & 224 & 96 & 96 \\

    Width Multiplier & N/A & 0.35 & N/A & N/A & 1.0 & 0.25 \\

    Data Bitwidth & INT8 & INT8 & INT8 & INT8 & INT8 & Mixed \\
    
    Latency & 463 ms & 296 ms$^3$ & 1133 ms & 859 ms & 866 ms & 64 ms \\
    
    Peak Memory & 416 KB & 211 KB$^3$ & 285 KB & 434 KB & 397 KB & 54 KB \\
    
    Flash Usage & 1.7 MB & < 1 MB$^2$ & 0.8 MB & 1.8 MB & 4.7 MB & 0.2 MB \\
    
    Energy Consumption & - & - & 479 mJ & - & 546 mJ & - \\
    
    Top-1 Accuracy & 68.0\% & - & 88.0\% & 71.8\% & 64.0\% & $\sim$76\% \\
    \midrule \midrule

     & \textbf{\makecell{UDC\\ NeurIPS'22~\cite{fedorov2022udc}}} & \textbf{\makecell{TinyTL\\ NeurIPS'20~\cite{cai2020tinytl}}} & \textbf{\makecell{TinyOL\\ IJCNN'21~\cite{ren2021tinyol}}} & \textbf{\makecell{POET\\ ICML'22~\cite{patil2022poet}}} & \textbf{\makecell{MiniLearn\\ EWSN’22~\cite{profentzas2022minilearn}}} &  \textbf{\makecell{MCUNetV3\\ NeurIPS’22~\cite{lin2022ondevice_mcunetv3}}} \\  
    \midrule 
    \makecell[l]{On-Device Training\\ or Inference} & Inference & Training & Training & Training & Training & Training \\

    Measured Device & \makecell{N/A\\ (Simulation)} & \makecell{N/A\\ (Simulation)} & \makecell{MCU\\ (nRF52840)} & \makecell{MCU\\ (nRF52840)} & \makecell{MCU\\ (nRF52840)} & \makecell{MCU\\ (STM32F746)} \\

    Dataset & ImageNet & CIFAR-10 & Self-Collected & CIFAR-10 & KWS-subset & VWW \\
    
    Model & UDC & \makecell{ProxylessNAS-\\Mobile} & Autoencoder & ResNet-18 & Customized & MCUNet \\



    Data Bitwidth & Mixed & FP32 & FP32 & FP32 &  Mixed & INT8 \\
    
    Latency & - & - & - & 49 ms & 93 ms & 546 ms \\
    
    Peak Memory & - & 65 MB & < 256 KB$^2$ & 271 KB & 196 KB & 173 KB \\
    
    Flash Usage & 1.27 MB & - & < 1 MB$^2$ & < 1 MB$^2$ & 0.9 MB & 0.7MB \\
    
    Energy Consumption & - & - & - & 868 mJ & 1486 mJ & - \\
    
    Top-1 Accuracy & 72.1\% & 96.1\% & - & 95.5\% & 88.5\% & 89.3\% \\
    \bottomrule
    \end{tabular}
    }
    }
    \footnotesize{$^1$Measured by CMix-NN paper~\cite{capotondi2020cmix}. $^2$Speculated by the specification of the corresponding MCU. $^3$Measured by MCUNet paper~\cite{lin2020mcunet}. $^4$In a private email on Dec. 22, 2022, the authors of the papers replied that the energy consumption should be interpreted as mJ instead of $\mu$J in their papers.}
\end{table*}

\begin{table*}[!t]
    \caption{Performance comparison of various tiny models and inference frameworks on STM32H743, which runs at 480MHz with the resource constraint of 512 KB peak memory and 2 MB storage. 
    }
    \label{tab:closed_track}
    \centering
    \small{
    \begin{tabular}{l||>{\centering\arraybackslash}p{3.5cm}|>{\centering\arraybackslash}p{3.5cm}|>{\centering\arraybackslash}p{3.5cm}|>{\centering\arraybackslash}p{3.5cm}}
    \toprule
     & \textbf{\makecell{CMSIS-NN\\ arXiv’18~\cite{lai2018cmsis}}} & \textbf{\makecell{X-Cube-AI\\~\cite{X-Cube-AI}}} & \textbf{\makecell{TinyEngine\\ NeurIPS’20~\cite{lin2020mcunet}}} & \textbf{\makecell{TF-Lite Micro\\ MLSys'21~\cite{david2021_tflitemicro}}} \\  
    
    \midrule \midrule 

    \multicolumn{5}{c}{\textbf{Dataset}: VWW; \textbf{Model}: mcunet-vww0	; \textbf{Input Resolution}: 64; \textbf{Width Multiplier}: N/A; \textbf{Top-1 Accuracy}: 87.3\%} \\
    \midrule 

    \textbf{Latency} & 53 ms & 32 ms & 27 ms & 587 ms \\
    
    \textbf{Peak Memory} & 163 KB & 88 KB & 59 KB & 163 KB \\
    
    \textbf{Storage usage} & 646 KB & 463 KB & 453 KB & 627 KB \\
    
    \midrule \midrule
    
    \multicolumn{5}{c}{\textbf{Dataset}: VWW; \textbf{Model}: mcunet-vww1	; \textbf{Input Resolution}: 80; \textbf{Width Multiplier}: N/A; \textbf{Top-1 Accuracy}: 88.9\%} \\
    \midrule 

    \textbf{Latency} & 97 ms & 57 ms & 51 ms & 1120 ms \\
    
    \textbf{Peak Memory} & 220 KB & 113 KB & 92 KB & 220 KB \\
    
    \textbf{Storage usage} & 736 KB & 534 KB & 521 KB & 718 KB \\
    
    \midrule \midrule

    \multicolumn{5}{c}{\textbf{Dataset}: VWW; \textbf{Model}: mcunet-vww2	; \textbf{Input Resolution}: 144; \textbf{Width Multiplier}: N/A; \textbf{Top-1 Accuracy}: 91.8\%} \\
    \midrule 

    \textbf{Latency} & 478 ms & 269 ms & 234 ms & 5310 ms \\
    
    \textbf{Peak Memory} & 390 KB & 201 KB & 174 KB & 385 KB \\
    
    \textbf{Storage usage} & 1034 KB & 774 KB & 741 KB & 1016 KB \\
    
    \midrule \midrule
    
    \multicolumn{5}{c}{\textbf{Dataset}: ImageNet; \textbf{Model}: mcunet-in0	; \textbf{Input Resolution}: 48; \textbf{Width Multiplier}: N/A; \textbf{Top-1 Accuracy}: 40.4\%} \\
    \midrule 

    \textbf{Latency} & 51 ms & 35 ms & 25 ms & 596 ms \\
    
    \textbf{Peak Memory} & 161 KB & 69 KB & 49 KB & 161 KB \\
    
    \textbf{Storage usage} & 1090 KB & 856 KB & 842 KB & 1072 KB \\

    \midrule \midrule

    \multicolumn{5}{c}{\textbf{Dataset}: ImageNet; \textbf{Model}: mcunet-in1	; \textbf{Input Resolution}: 96; \textbf{Width Multiplier}: N/A; \textbf{Top-1 Accuracy}: 49.9\%} \\
    \midrule 

    \textbf{Latency} & 103 ms & 63 ms & 56 ms & 1227 ms \\
    
    \textbf{Peak Memory} & 219 KB & 106 KB & 96 KB & 219 KB \\
    
    \textbf{Storage usage} & 956 KB & 737 KB & 727 KB & 937 KB \\
    
    \midrule \midrule

    \multicolumn{5}{c}{\textbf{Dataset}: ImageNet; \textbf{Model}: mcunet-in2	; \textbf{Input Resolution}: 160; \textbf{Width Multiplier}: N/A; \textbf{Top-1 Accuracy}: 60.3\%} \\
    \midrule 

    \textbf{Latency} & 642 ms & 351 ms & 280 ms & 6463 ms \\
    
    \textbf{Peak Memory} & 469 KB & 238 KB & 215 KB & 460 KB \\
    
    \textbf{Storage usage} & 1102 KB & 849 KB & 830 KB & 1084 KB \\
    
    \midrule \midrule

    \multicolumn{5}{c}{\textbf{Dataset}: ImageNet; \textbf{Model}: mcunet-in3	; \textbf{Input Resolution}: 176; \textbf{Width Multiplier}: N/A; \textbf{Top-1 Accuracy}: 61.8\%} \\
    \midrule 

    \textbf{Latency} & 770 ms & 414 ms & 336 ms & 7821 ms \\
    
    \textbf{Peak Memory} & 493 KB & 243 KB & 260 KB & 493 KB \\
    
    \textbf{Storage usage} & 1106 KB & 867 KB & 835 KB & 1091 KB \\
    
    \midrule \midrule

    \multicolumn{5}{c}{\textbf{Dataset}: ImageNet; \textbf{Model}: mcunet-in4	; \textbf{Input Resolution}: 160; \textbf{Width Multiplier}: N/A; \textbf{Top-1 Accuracy}: 68.0\%} \\
    \midrule 

    \textbf{Latency} & OOM & 516 ms & 463 ms & OOM \\
    
    \textbf{Peak Memory} & OOM & 342 KB & 416 KB & OOM \\
    
    \textbf{Storage usage} & OOM & 1843 KB & 1825 KB & OOM \\
    


    
    
    
    \midrule \midrule

    \multicolumn{5}{c}{\textbf{Dataset}: ImageNet; \textbf{Model}: proxyless-w0.3; \textbf{Input Resolution}: 64; \textbf{Width Multiplier}: 0.3; \textbf{Top-1 Accuracy}: 37.0\% } \\
    \midrule 

    \textbf{Latency} & 54 ms & 35 ms & 23 ms & 512 ms \\
    
    \textbf{Peak Memory} & 136 KB & 97 KB & 35 KB & 128 KB \\
    
    \textbf{Storage usage} & 1084 KB & 865 KB & 777 KB & 1065 KB \\

    \midrule \midrule

    \multicolumn{5}{c}{\textbf{Dataset}: ImageNet; \textbf{Model}: proxyless-w0.3; \textbf{Input Resolution}: 176; \textbf{Width Multiplier}: 0.3; \textbf{Top-1 Accuracy}: 56.2\%} \\
    \midrule 

    \textbf{Latency} & 380 ms & 205 ms & 176 ms & 3801 ms \\
    
    \textbf{Peak Memory} & 453 KB & 221 KB & 259 KB & 453 KB \\
    
    \textbf{Storage usage} & 1084 KB & 865 KB & 779 KB & 1065 KB \\

    \midrule \midrule

    \multicolumn{5}{c}{\textbf{Dataset}: ImageNet; \textbf{Model}: mbv2-w0.3; \textbf{Input Resolution}: 64; \textbf{Width Multiplier}: 0.3; \textbf{Top-1 Accuracy}: 34.1\%} \\
    \midrule 

    \textbf{Latency} & 43 ms & 29 ms & 23 ms & 467 ms \\
    
    \textbf{Peak Memory} & 173 KB & 88 KB & 61 KB & 173 KB \\
    
    \textbf{Storage usage} & 959 KB & 768 KB & 690 KB & 940 KB \\
    
    \bottomrule
    \end{tabular}
    }
   \footnotesize{$^1$All the inference frameworks used in this measurement are the latest versions as of Dec. 19, 2022. $^2$The measurement of X-Cube-AI (v7.3.0) is with the default compilation setting, \ie, balanced optimization. $^3$OOM denotes Out Of Memory. $^4$All the models are available on:\url{https://github.com/mit-han-lab/mcunet}.}
    
\end{table*}

\section{Recent Progress in TinyML}
\label{related}

\subsection{Recent Progress on TinyML Inference}
\label{related_inference}
TinyML and deep learning on MCUs have seen rapid growth in industry and academia in recent years. The primary challenge of deploying deep learning models on MCUs for inference is the limited memory and computation available on these devices. For example, a popular ARM Cortex-M7 MCU, the STM32F746, has only 320KB of SRAM and 1MB of flash memory. In deep learning scenarios, SRAM limits the size of activations (read and write) while flash memory limits the size of the model (read-only). In addition, the STM32F746 has a processor with a clock speed of 216 MHz, which is 10 to 20 times lower than laptops. To enable deep learning inference on MCUs, researchers have proposed various designs and solutions to address these issues. Table~\ref{tab:open_track} summarizes the recent related studies on TinyML targeting MCUs, including both algorithm solutions and system solutions. In Table~\ref{tab:closed_track}, we measured three different metrics (\ie, latency, peak memory, and flash usage) of four representative related studies (\ie, CMSIS-NN~\cite{lai2018cmsis}, X-Cube-AI~\cite{X-Cube-AI}, TinyEngine~\cite{lin2020mcunet}, and TF-Lite Micro~\cite{david2021_tflitemicro}) on an identical MCU (STM32H743) and identical datasets (VWW and Imagenet), in order to provide a more accurate and transparent comparison.

\myparagraph{Algorithm Solutions}
The importance of neural network's efficiency to the overall performance of a deep learning system cannot be overstated. Compressing off-the-shelf networks by removing redundancy and reducing complexity through pruning~\cite{han2015learning, he2017channel, lin2017runtime, liu2017learning, he2018amc, liu2019metapruning} and quantization~\cite{han2016deep, zhu2016trained, rastegari2016xnor, zhou2016dorefa, courbariaux2016binarynet, choi2018pact, wang2019haq, langroudi2021tent} are two popular methods to improve network efficiency. Tensor decomposition~\cite{lebedev2014speeding,gong2014compressing,kim2015compression} is also an efficient compression technique. In order to enhance network efficiency, knowledge distillation is also a method to transfer information learned from one teacher model to another student model~\cite{hinton2015distilling, park2019distillation, tung2019distillation, mirzadeh2020distillation, wang2021distillation, yang2022distillation, zhao2022distillation, beyer2022distillation}. Another method is to directly design tiny and efficient network structures~\cite{howard2017mobilenets, sandler2018mobilenetv2, ma2018shufflenet,zhang2018shufflenet}. Recently, neural architecture search (NAS) has dominated the design of efficient networks~\cite{zoph2017neural, zoph2018learning, liu2019darts, cai2019proxylessnas, tan2019mnasnet, wu2019fbnet}.
To make deep learning feasible on MCUs, researchers have proposed various algorithm solutions. Rusci et al. proposed a rule-based quantization strategy that minimizes the bit precision of activations and weights in order to reduce memory usage~\cite{rusci2019memory}. Depending on the memory constraints of various devices, this method can quantize activations and weights with 8 bits, 4 bits, or 2 bits of mixed precision. On the other hand, although neural architecture search (NAS) has been successful in finding efficient network architectures, its effectiveness is highly dependent on the quality of the search space~\cite{radosavovic2020designing}. For MCUs with limited memory, standard model designs and appropriate search spaces are especially lacking. To address this, TinyNAS, proposed as part of MCUNet, employs a two-step NAS strategy that optimizes the search space according to the available resources~\cite{lin2020mcunet}. TinyNAS then specializes network architectures within the optimized search space, allowing it to automatically deal with a variety of constraints (\eg, device, latency, energy, memory) at low search costs. MicroNets observed that the inference latency of networks in the NAS search space for MCUs varies linearly with the number of FLOPs in the model~\cite{banbury2021micronets}. As a result, it proposed differentiated NAS, which treats the FLOPs as a proxy for latency in order to achieve both low memory consumption and high speed. MCUNetV2 identified that the imbalanced memory distribution is the primary memory bottleneck in most convolutional neural network designs, where the memory usage of the first few blocks is an order of magnitude greater than the rest of the network~\cite{lin2021mcunetv2}. As a result, this study proposed receptive field redistribution to shift the receptive field and FLOPs to a later stage, reducing the halo's computation overhead. To minimize the difficulty of manually redistributing the receptive field, this study also automated the neural architecture search process to simultaneously optimize the neural architecture and inference scheduling. UDC explored a broader design search space to generate compressible neural networks with high accuracy for neural processing units (NPUs), which can address the memory problem by exploiting model compression with a broader range of weight quantization and sparsity~\cite{fedorov2022udc}.

\myparagraph{System Solutions}
In recent years, popular training frameworks such as PyTorch~\cite{pytorch2019}, TensorFlow~\cite{abadi2016tensorflow}, MXNet~\cite{chen2015mxnet}, and JAX~\cite{jax2018github} have contributed to the success of deep learning. However, these frameworks typically rely on a host language (\eg, Python) and various runtime systems, which adds significant overhead and makes them incompatible with tiny edge devices. Emerging frameworks such as TVM~\cite{chen2018tvm}, TF-Lite~\cite{tflite}, MNN~\cite{jiang2020mnn}, NCNN~\cite{ncnn}, TensorRT~\cite{tensorRT}, and OpenVino~\cite{vaswani2017attention} offer lightweight runtime systems for edge devices such as mobile phones, but they are not yet small enough for MCUs. These frameworks cannot accommodate IoT devices and MCUs with limited memory.

CMSIS-NN implements optimized kernels to increase inference speed, minimize memory footprint, and enhance the energy efficiency of deep learning models on ARM Cortex-M processors~\cite{lai2018cmsis}. X-Cube-AI, designed by STMicroelectronics, enables the automatic conversion of pre-trained deep learning models to run on STM MCUs with optimized kernel libraries~\cite{X-Cube-AI}. TVM~\cite{chen2018tvm} and AutoTVM~\cite{chen2018learning} also supports microcontrollers (referred to as $\mu$TVM/microTVM~\cite{microTVM}). Compilation techniques can also be employed to reduce memory requirements. For instance, Stoutchinin et al. propose to improve deep learning performance on MCU by optimizing the convolution loop nest~\cite{stoutchinin2019optimally}. Liberis et al. and Ahn et al. present to reorder the operator executions to minimize peak memory~\cite{liberis2019neural, ahn2020ordering}, whereas Miao et al. seek to achieve better memory utilization by temporarily swapping data off SRAM~\cite{miao2021enabling}. With a similar goal of reducing peak memory, other researchers further propose computing partial spatial regions across multiple layers~\cite{alwani2016fused, goetschalckx2019breaking, saha2020rnnpool}. Additionally, CMix-NN supports mixed-precision kernel libraries of quantized activation and weight on MCU to reduce memory footprint~\cite{capotondi2020cmix}. TinyEngine, as part of MCUNet, is proposed as a memory-efficient inference engine for expanding the search space and fitting a larger model~\cite{lin2020mcunet}. TinyEngine transfers the majority of operations from runtime to compile time before generating only the code that will be executed by the TinyNAS model. In addition, TinyEngine adapts memory scheduling to the overall network topology as opposed to layer-by-layer optimization. TensorFlow-Lite Micro (TF-Lite Micro) is among the first deep-learning frameworks to support bare-metal microcontrollers in order to enable deep-learning inference on MCUs with tight memory constraints~\cite{david2021_tflitemicro}. However, the aforementioned frameworks only support per-layer inference, which limits the model capacity that can be executed with only a small amount of memory and makes higher-resolution input impossible. Hence, MCUNetV2 proposes a generic patch-by-patch inference scheduling, which operates on a small spatial region of the feature map and drastically reduces peak memory usage, and thus makes the inference with high-resolution input on MCUs feasible~\cite{lin2021mcunetv2}. TinyOps combines fast internal memory with an additional slow external memory through Direct Memory Access (DMA) peripheral to enlarge memory size and speed up inference~\cite{Sadiq2022TinyOps}. TinyMaix, similar to CMSIS-NN, is an optimized inference kernel library, but it eschews new but rare features and seeks to preserve the readability and simplicity of the codebase~\cite{TinyMaix}.

\subsection{Recent Progress on TinyML Training}
\label{related_training}
On-device training on small devices is gaining popularity, as it enables machine learning models to be trained and refined directly on small and low-power devices. On-device training offers several benefits, including the provision of personalized services and the protection of user privacy, as user data is never transmitted to the cloud. However, on-device training presents additional challenges compared to on-device inference, due to larger memory footprints and increased computing operations needed to store intermediate activations and gradients.

Researchers have been investigating ways to reduce the memory footprint of training deep learning models. One kind of approach is to design lightweight network structures manually or by utilizing NAS~\cite{cai2019proxylessnas, tan2019efficientnet, tan2021efficientnetv2}. Another common approach is to trade computation for memory efficiency, such as freeing up activation during inference and recomputing discarded activation during the backward propagation~\cite{audrunas2016nips, chen2016training}. However, such an approach comes at the expense of increased computation time, which is not affordable for tiny devices with limited computation resources. Another approach is layer-wise training, which can also reduce the memory footprint compared to end-to-end training. However, it is not as effective at achieving high levels of accuracy~\cite{klaus2017iclr}. Another approach reduces the memory footprint by building a dynamic and sparse computation graph for training by activation pruning ~\cite{liu2019dynamic}. Some researchers propose different optimizers~\cite{wang2019e2train}. Quantization is also a common approach that reduces the size of activation during training by reducing the bitwidth of training activation~\cite{wang2018training, sun2019hybrid}.

Due to limited data and computational resources, on-device training usually focuses on transfer learning. In transfer learning, a neural network is first pre-trained on a large-scale dataset, such as ImageNet~\cite{krizhevsky2012imagenet}, and used as a feature extractor~\cite{cui2018large, kornblith2019better, kolesnikov2020big}. Then, only the last layer needs to be fine-tuned on a smaller, task-specific dataset~\cite{chatfield2014return, donahue2014decaf, gan2015devnet, sharif2014cnn}. This approach reduces the memory footprint by eliminating the need to store intermediate activations during training, but due to the limited capacity, the accuracy can be poor when the domain shift is large~\cite{cai2020tinytl}. Fine-tuning all layers can achieve better accuracy but requires large memory to store activation, which is not affordable for tiny devices~\cite{kornblith2019better, cui2018large}. Recently, several memory-friendly on-device training frameworks were proposed~\cite{wang2022melon, xu2022mandheling, gim2022trainingonmobile}, but these frameworks targeted larger edge devices (i.e., mobile devices) and cannot be adopted on MCUs. An alternative approach is only updating the parameters of batch normalization layers~\cite{frankle2020training, mudrakarta2019k}. This reduces the number of trainable parameters, which however does not translate to memory efficiency~\cite{cai2020tinytl} because the intermediate activation of batch normalization layers still needs to be stored in the memory.
It has been shown that the activation of a neural network is the main factor limiting the ability to train on small devices. Tiny-Transfer-Learning (TinyTL) addresses this issue by freezing the weights of the network and only fine-tuning the biases, which allows intermediate activations to be discarded during backward propagation, reducing peak memory usage~\cite{cai2020tinytl}. TinyOL trains only the weights of the final layer, allowing for weight training while keeping the activation small enough to fit on small devices~\cite{ren2021tinyol}. This enables incremental on-device streaming of data for training. However, fine-tuning only the biases or the last layer may not provide sufficient precision.
To train more layers on devices with limited memory, POET (Private Optimal Energy Training)~\cite{patil2022poet} introduces two techniques: rematerialization, which frees up activations early at the cost of recomputation, and paging, which allows activations to be transferred to secondary storage. POET uses an integer linear program to find the energy-optimal schedule for on-device training.
To further reduce the memory required to store trained weights, MiniLearn applies quantization and dequantization techniques to store the weights and intermediate output in integer precision and dequantizes them to floating-point precision during training~\cite{profentzas2022minilearn}. When deployed on tiny devices, deep learning models are often quantized to reduce the memory usage of parameters and activations. However, even after quantization, the parameters may still be too large to fit in the limited hardware resources, preventing full back-propagation. To address these challenges, MCUNetV3 proposes an algorithm-system co-design approach~\cite{lin2022ondevice_mcunetv3}. The algorithm part includes Quantization-Aware Scaling (QAS) and the sparse update. QAS calibrates the gradient scales and stabilizes 8-bit quantized training, while the sparse update skips the gradient computation of less important layers and sub-tensors. The system part includes the Tiny Training Engine (TTE), which has been developed to support both QAS and the sparse update, enabling on-device learning on microcontrollers with limited memory, such as those with 256KB or even less.
\section{Tiny Inference}
\label{tiny_inference}

\begin{figure*}[t]
    \centering
     \includegraphics[width=1\textwidth]{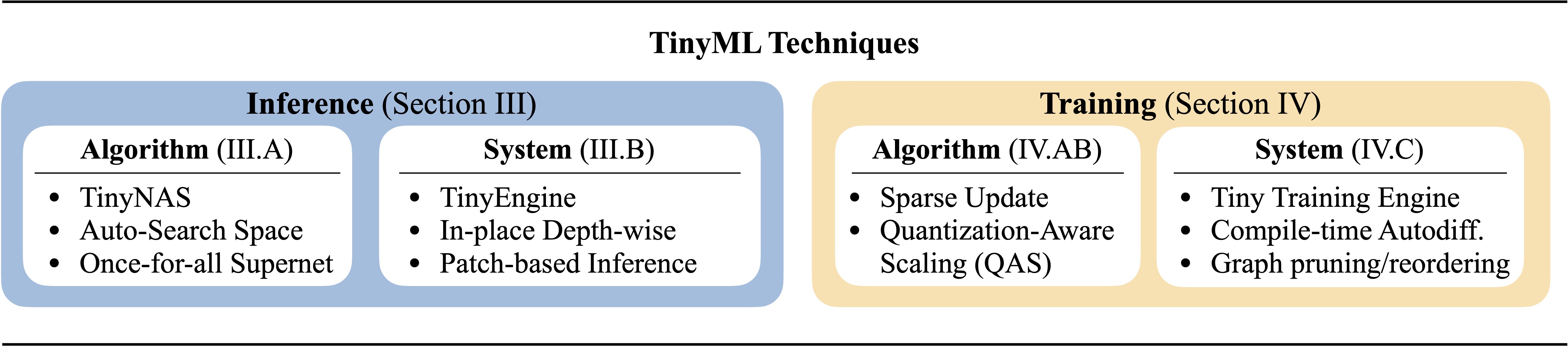}
    \caption{
    Techniques specifically designed for tiny devices. 
    In order to fully leverage the limited available resources, we need to take careful consideration of both the system and the algorithm.
    The co-design approach not only enables practical AI applications on a wide range of IoT platforms (\textit{inference}), but also allows AI to continuously learn over time, adapting to a world that is changing fast (\textit{training}).
    }
    \label{fig:design_space}
\end{figure*}

\begin{figure*}[t]
    \centering
     \includegraphics[width=0.8\textwidth]{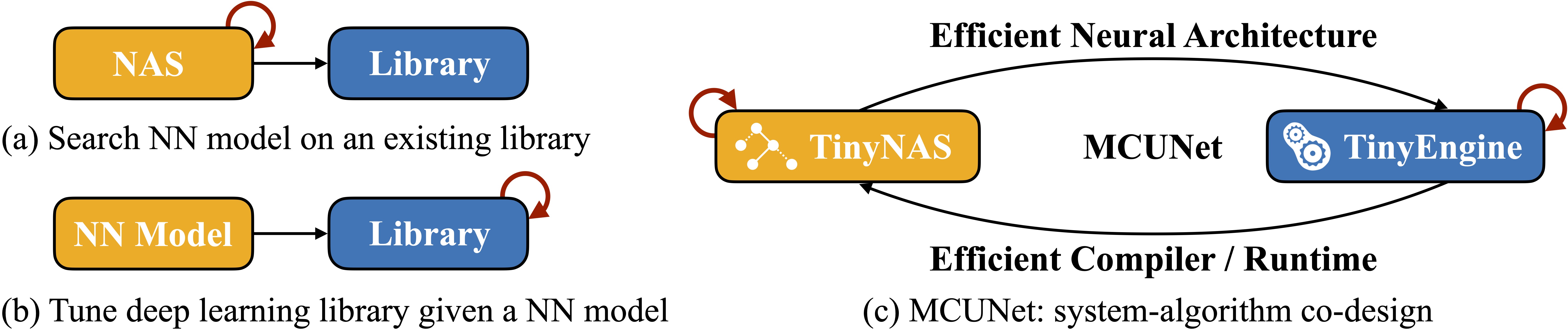}
    \caption{MCUNet jointly designs the neural architecture and the inference scheduling to fit the tight memory resource on microcontrollers. TinyEngine makes full use of the limited resources on MCU, allowing a larger design space for architecture search. With a larger degree of design freedom, TinyNAS is more likely to find a high accuracy model compared to using existing frameworks.
    }
    \label{fig:overview}
\end{figure*}

In this section, we discuss our recent work, MCUNet family~\cite{lin2020mcunet, lin2021mcunetv2}, a system-algorithm co-design framework that jointly optimizes the NN architecture (TinyNAS) and the inference scheduling (TinyEngine) in the same loop (Figure~\ref{fig:overview}). Compared to traditional methods that either (a) optimize the neural network using neural architecture search based on a given deep learning library (\eg, TensorFlow, PyTorch)~\cite{tan2019mnasnet, cai2019proxylessnas, wu2019fbnet}, or (b) tune the library to maximize the inference speed for a given network~\cite{chen2018tvm, chen2018learning}, MCUNet can better utilize the resources by system-algorithm co-design, enabling a better performance on microcontrollers. The design space of the inference part is listed in Figure~\ref{fig:design_space} (left). 

\subsection{TinyNAS: Automated Tiny Model Design}

TinyNAS is a two-stage neural architecture search method that first optimizes the search space to fit the tiny and diverse resource constraints, and then performs neural architecture search within the optimized space. By optimizing the search space, it significantly improves the accuracy of the final model.

\subsubsection{Automated search space optimization.}

\begin{figure*}[t]
    \centering
     \includegraphics[width=\textwidth]{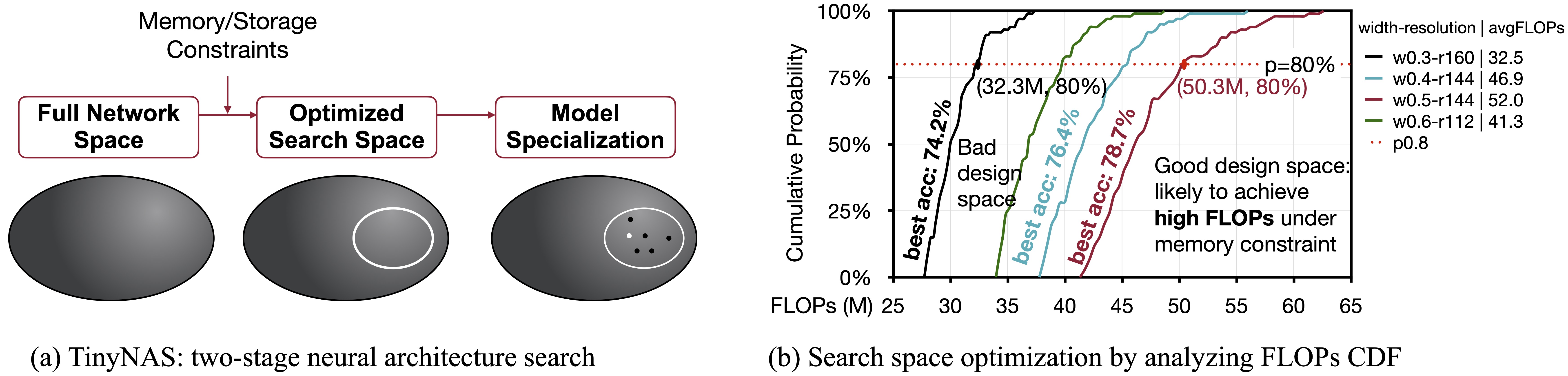}
    \caption{(a) TinyNAS is a two-stage neural architecture search method. It first specifies a sub-space according to the constraints, and then performs model specialization. (b) TinyNAS selects the best search space by analyzing the FLOPs CDF of different search spaces.
    Each curve represents a design space. Our insight is that the design space that is more likely to produce \textit{high FLOPs} models under the memory constraint gives higher model capacity, thus more likely to achieve high accuracy.
    }
    \label{fig:two_stage}
\end{figure*}
TinyNAS proposes to optimize the search space automatically at \emph{low cost} by analyzing the computation distribution of the satisfying models.
To fit the tiny and diverse resource constraints of different microcontrollers, 
TinyNAS scales the \emph{input resolution} and the \emph{width multiplier} of the mobile search space~\cite{tan2019mnasnet}. It chooses from an input resolution spanning $R=\{48, 64, 80, ..., 192, 208, 224\}$  and a width multiplier $W=\{0.2, 0.3, 0.4, ..., 1.0\}$ to cover a wide spectrum of resource constraints. This leads to $12\times9=108$ possible search space configurations $S=W\times R$. Each search space configuration contains $3.3\times 10^{25}$ possible sub-networks. The goal is to find the best search space configuration $S^*$ that contains the model with the highest accuracy while satisfying the resource constraints.

Finding $S^*$ is non-trivial. One way is to perform neural architecture search on each of the search spaces and compare the final results. But the computation would be astronomical. Instead, 
TinyNAS evaluates the quality of the search space by randomly sampling $m$ networks from the search space and comparing the distribution of satisfying networks. 
Instead of collecting the Cumulative Distribution Function (CDF) of each satisfying network's \textit{accuracy}~\cite{radosavovic2020designing}, which is computationally heavy due to tremendous training, it only collects the CDF of \textit{FLOPs} (see Figure~\ref{fig:two_stage}(b)). The intuition is that, within the same model family, the accuracy is usually positively related to the computation~\cite{canziani2016analysis,he2018amc}. A model with larger computation has a larger capacity, which is more likely to achieve higher accuracy. 

Take the study of the best search space for ImageNet-100 (a 100-class classification task taken from the original ImageNet) on STM32F746 as an example. We show the FLOPs distribution CDF of the top-10 search space configurations in Figure~\ref{fig:two_stage}(b). Only the models that satisfy the memory requirement at the best scheduling from TinyEngine are kept. For example, according to the experimental results on ImageNet-100, using the solid red space (average FLOPs 52.0M) achieves 2.3\% better accuracy compared to using the solid green space (average FLOPs 46.9M), showing the effectiveness of automated search space optimization. 

\subsubsection{Resource-constrained model specialization with Once-For-All NAS.}

\begin{figure*}[t]
    \centering
     \includegraphics[width=0.7\textwidth]{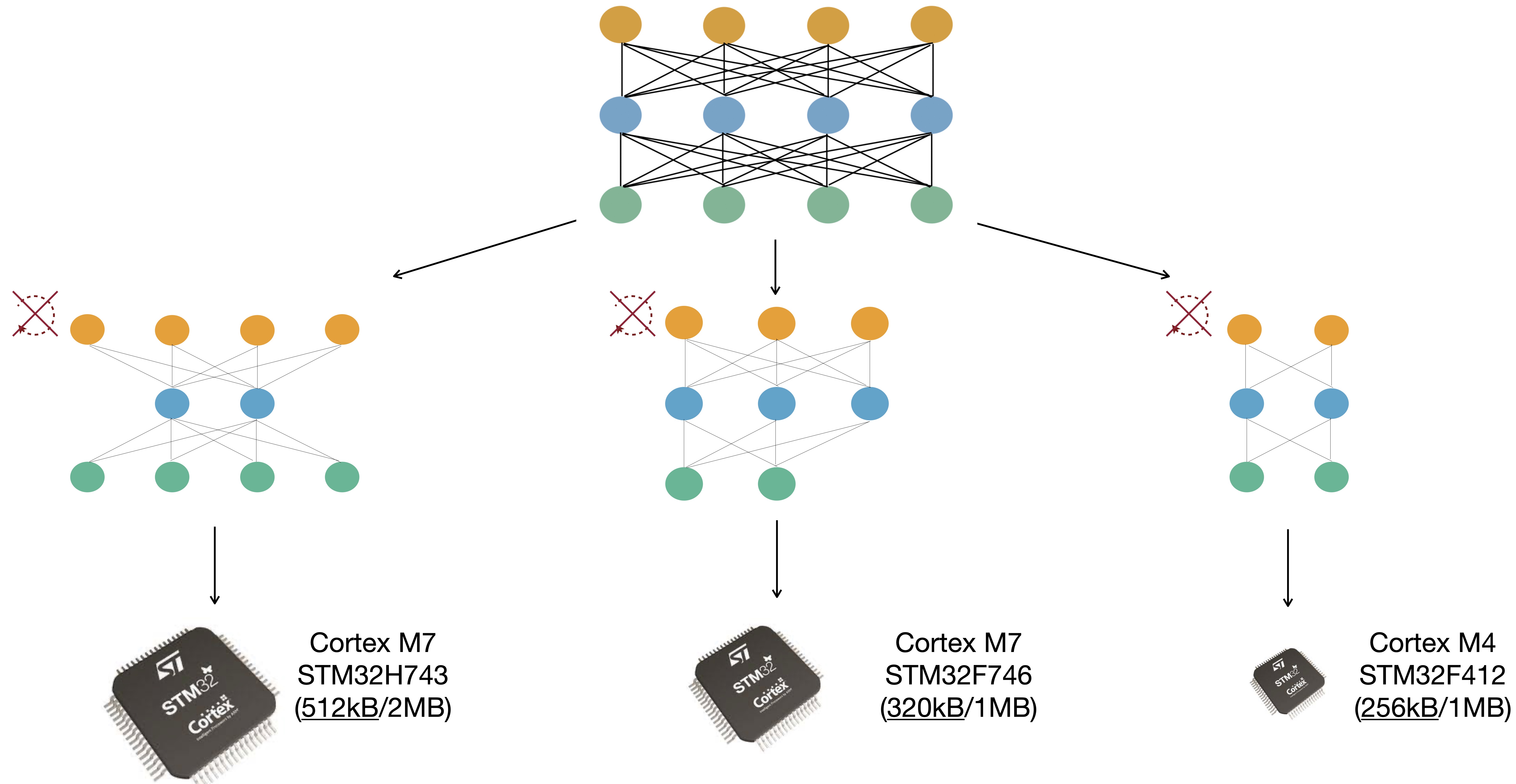}
    \caption{Once-For-All~\cite{cai2020once} trains one single super network that supports a wide range of sub-networks through weight sharing, and specializes different sub-network architectures for different MCU hardware. 
    }
    \label{fig:ofa_nas}
\end{figure*}

To specialize network architecture for various microcontrollers, we need to keep a low neural architecture search cost. 
Given an optimized search space, TinyNAS further performs one-shot neural architecture search~\cite{bender2018understanding, guo2019single} to efficiently find a good model.
Specifically, it follows Once-For-All (OFA) NAS~\cite{cai2020once} to perform network specialization (Figure~\ref{fig:ofa_nas}).
We train one super network that contains all the possible sub-networks through \emph{weight sharing} and use it to estimate the performance of each sub-network. 
The search space is based on the widely-used mobile search space~\cite{tan2019mnasnet, cai2019proxylessnas, wu2019fbnet, cai2020once} and supports variable kernel sizes for depth-wise convolution (3/5/7), variable expansion ratios for inverted bottleneck (3/4/6) and variable stage depths (2/3/4). The number of possible sub-networks that TinyNAS can cover in the search space is large: $2\times10^{19}$.
For each batch of data, it randomly samples 4 sub-networks, calculates the loss, backpropagates the gradients for each sub-network, and updates the corresponding weights. 
It then performs an evolution search to find the best model within the search space that meets the onboard resource constraints while achieving the highest accuracy. 
For each sampled network, it uses TinyEngine to optimize the memory scheduling to measure the optimal memory usage. With such a kind of co-design, we can efficiently fit the tiny memory budget. 

\subsection{TinyEngine: A Memory-Efficient Inference Library}
Researchers used to assume that using different deep learning frameworks (libraries) will only affect the  \textit{inference speed} but not the \textit{accuracy} . However, this is not the case for TinyML: the efficiency of the inference library matters a lot to both the latency and accuracy of the searched  model. Specifically, a good inference framework will make full use of the limited resources in MCUs, avoiding waste of memory, and allowing a larger search space for architecture search. With a larger degree of design freedom, TinyNAS is more likely to find a high-accuracy model. Thus, TinyNAS is co-designed with a memory-efficient inference library, TinyEngine.

\subsubsection{Code generation.} 
\begin{figure*}[t]
    \centering
    \includegraphics[width=0.85\textwidth]{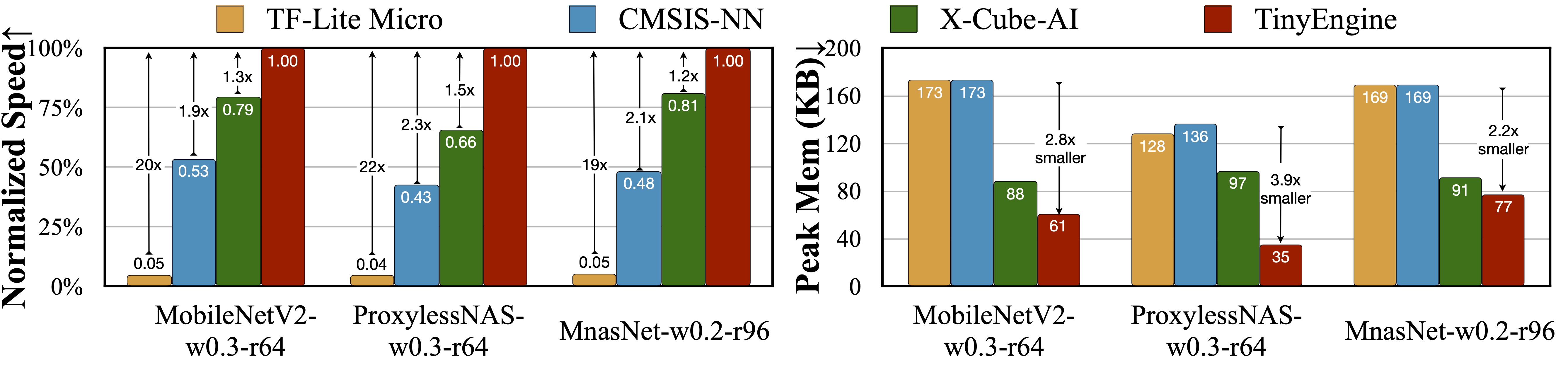}
    \caption{TinyEngine achieves higher inference efficiency than existing inference frameworks while reducing memory usage. 
     \textbf{Left}: TinyEngine is up to \textcolor{black}{22$\times$, 2.3$\times$, and 1.5$\times$ faster than TF-Lite Micro, CMSIS-NN, and X-Cube-AI, }respectively.
    \textbf{Right}: By reducing the memory usage, TinyEngine can run various model designs with tiny memory, enlarging the design space for TinyNAS under the limited memory of MCU.}
    \label{fig:speedup}
\end{figure*}

\begin{figure*}[t]
    \centering
    \includegraphics[width=\textwidth]{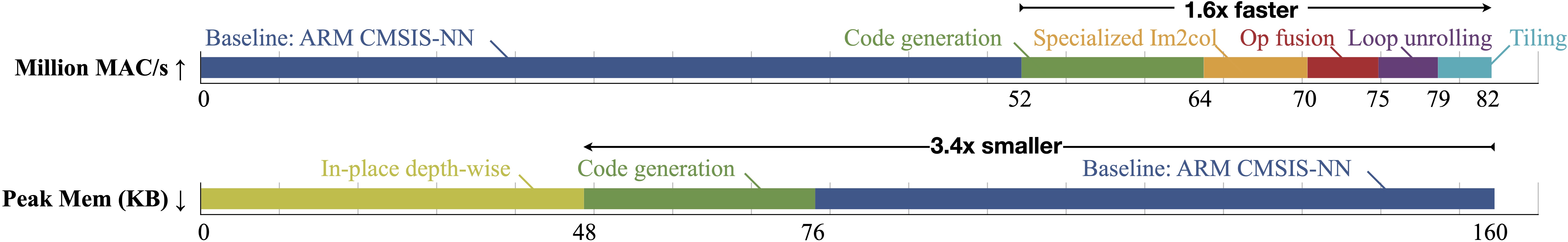}
    \caption{TinyEngine outperforms existing libraries by eliminating runtime overheads, specializing each optimization technique, and adopting in-place depth-wise convolution. This effectively enlarges design space for TinyNAS under a given latency/memory constraint.}
    \label{fig:speedup_memory_ablation}
\end{figure*}

Most existing inference libraries (\eg, TF-Lite Micro, CMSIS-NN) are interpreter-based. Though it is easy to support cross-platform development, it requires extra memory, the most expensive resource in MCU, to store the meta-information (such as model structure parameters). Instead, TinyEngine only focuses on MCU devices and adopts code generator-based compilation. This not only avoids the time for runtime interpretation, but also frees up the memory usage to allow design and inference of larger models. Compared to CMSIS-NN, TinyEngine reduced memory usage by 2.1$\times$ and improve inference efficiency by 22\% via code generation, as shown in Figures~\ref{fig:speedup} and ~\ref{fig:speedup_memory_ablation}.

\setlength{\columnsep}{3pt}%
\begin{wrapfigure}{r}{0.2\textwidth}
    \includegraphics[width=0.2\textwidth]{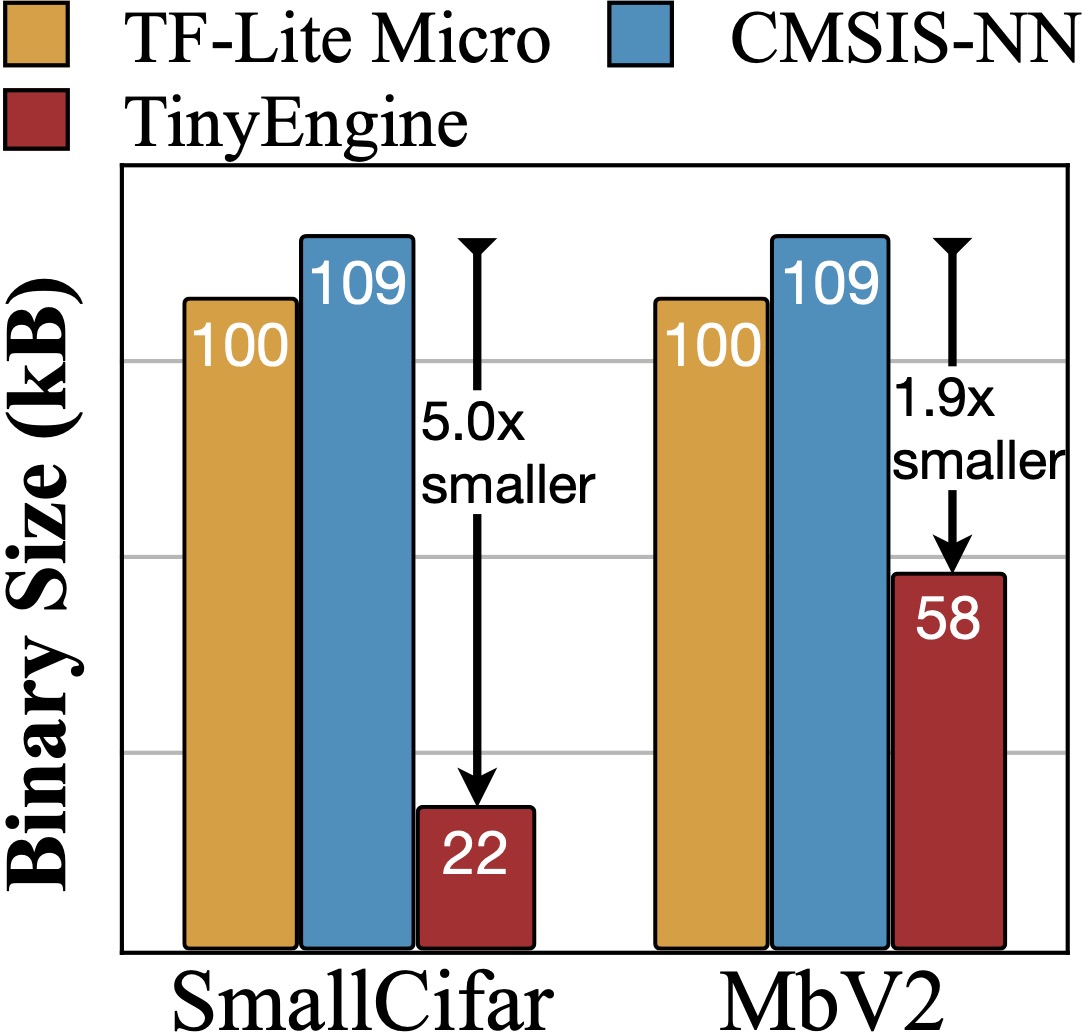}
    \caption{Binary size.}
    \label{fig:binary_size}
\end{wrapfigure}

The binary size of TinyEngine is lightweight, making it very memory-efficient for MCUs. 
\textcolor{black}{The model directly compiled by well-known programming languages for deep learning (\eg, Python, Cython, \etc) cannot be run on MCUs as the size of their dependencies and packages are already larger than the Flash size of MCUs, let alone the size of the compiled model.
Besides,} unlike interpreter-based TF-Lite Micro, which prepares the code for \textit{every} operation (\eg, conv, softmax) to support cross-model inference even if they are not used, which has high redundancy. TinyEngine only compiles the operations that are used by a given model into the binary. \textcolor{black}{That is, the reduction of binary size of the model compiled by TinyEngine comes from not only the benefit of compilation over interpretation but also the model-specific optimization/specialization.} As shown in Figure~\ref{fig:binary_size}, such model-adaptive compilation reduces code size by up to 4.5$\times$ and 5.0$\times$ compared to TF-Lite Micro and CMSIS-NN, respectively.

\begin{figure*}[t]
    \centering
     \includegraphics[width=0.75\textwidth]{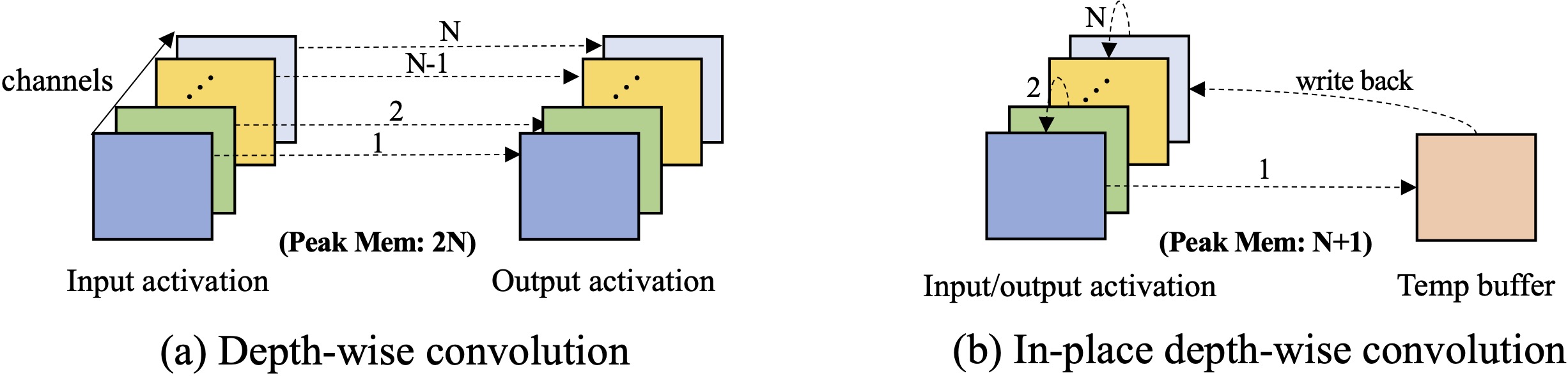}
    \caption{TinyEngine reduces peak memory by performing in-place depth-wise convolution. \textbf{Left:} Conventional depth-wise convolution requires 2N memory footprint for activations. \textbf{Right:} in-place depth-wise convolution reduces the memory of depth-wise convolutions to N+1. Specifically, the output activation of the first channel is stored in a temporary buffer. Then, for each following channel, the output activation overwrites the input activation of its previous channel. Finally, the output activation of the first channel stored in the buffer is written back to the input activation of the last channel.
    }
    \label{fig:inplaceDW}
\end{figure*}

\subsubsection{In-place depth-wise convolution}
TinyEngine supports \textit{in-place} depth-wise convolution to further reduce peak memory. Different from standard convolutions, depth-wise convolutions do not perform filtering across channels. Therefore, once the computation of a channel is completed, the input activation of the channel can be overwritten and used to store the output activation of another channel, allowing activation of depth-wise convolutions to be updated in-place as shown in Figure~\ref{fig:inplaceDW}. This method reduces the measured memory usage by 1.6$\times$ as shown in Figure~\ref{fig:speedup_memory_ablation}.

\subsubsection{Patched-based Inference}

TinyNAS and TinyEngine have significantly reduced the peak memory at the same level of accuracy. But we still notice a very imbalanced peak memory usage per block.

\begin{figure*}[t]
    \centering
     \includegraphics[width=1.0\textwidth]{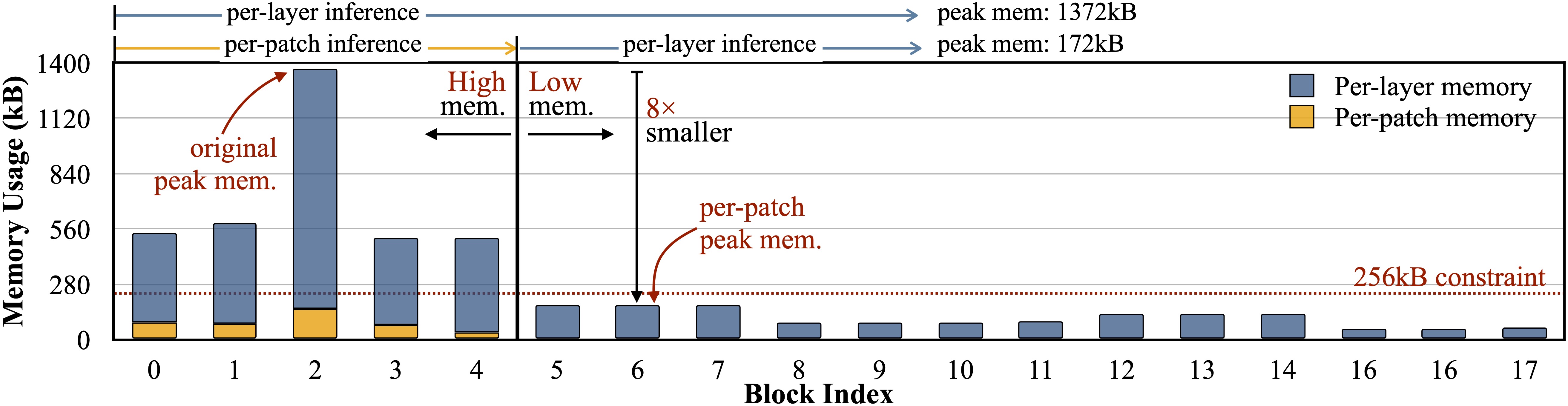}
    \caption{MobileNetV2~\cite{sandler2018mobilenetv2} has a very \emph{imbalanced memory usage distribution}. The peak memory is determined by the first 5 blocks with high peak memory, while the later blocks all share a small memory usage. By using per-patch inference ($4\times4$ patches), we are able to significantly reduce the memory usage of the first 5 blocks, and reduce the overall peak memory by 8$\times$, fitting MCUs with a 256kB memory budget. Notice that the model architecture and accuracy are not changed for the two settings. The memory usage is measured in \texttt{int8}. %
    }
    \label{fig:mbv2_mem}
\end{figure*}
\textbf{Imbalanced memory distribution.}
As an example, the per-block peak memory usage of MobileNetV2~\cite{sandler2018mobilenetv2} is shown in Figure~\ref{fig:mbv2_mem}. The profiling is done in \texttt{int8}. 
There is a clear pattern of \emph{imbalanced memory usage distribution}. The first 5 blocks have large peak memory, exceeding the memory constraints of MCUs, while the remaining 13 blocks easily fit 256KB memory constraints. The third block has $8\times$ larger memory usage than the rest of the network, becoming the memory bottleneck.
There are similar patterns for other efficient network designs, which is quite common across different CNN backbones, even for models specialized for memory-limited microcontrollers~\cite{lin2020mcunet}.

The phenomenon applies to most single-branch or residual CNN designs due to the hierarchical structure\footnote{some CNN designs have highly complicated branching structure (\eg, NASNet~\cite{zoph2018learning}), but they are generally less efficient for inference~\cite{ma2018shufflenet, tan2019mnasnet, cai2019proxylessnas}; thus not widely used for edge computing.}: after each stage, the image resolution is down-sampled by half, leading to 4$\times$ fewer pixels, while the channel number increases only by 2$\times$~\cite{simonyan2014very, he2016deep, howard2017mobilenets} or by an even smaller ratio~\cite{sandler2018mobilenetv2, howard2019searching, tan2019efficientnet}, resulting in a decreasing activation size. Therefore, the memory bottleneck tends to appear at the early stage of the network, after which the peak memory usage is much smaller. 


\begin{figure*}[t]
    \centering
     \includegraphics[width=1.0\textwidth]{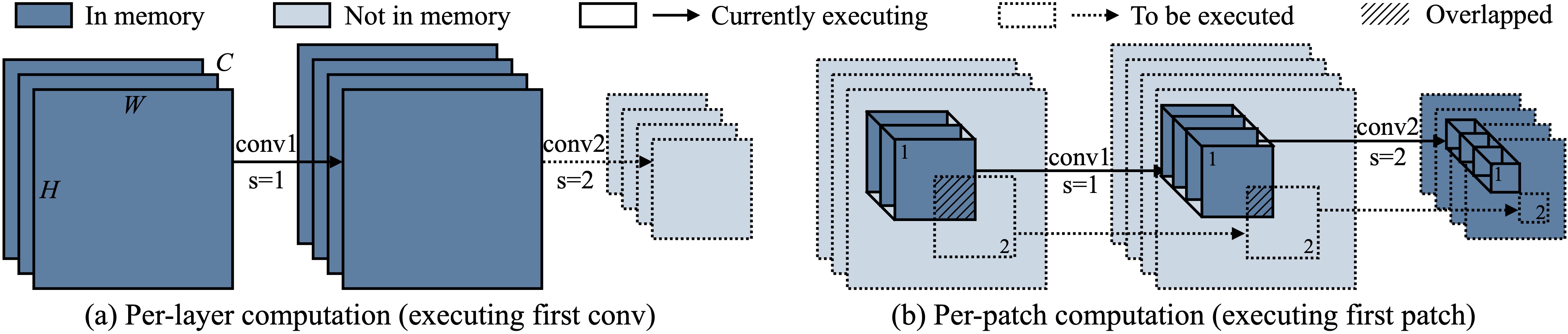}
    \caption{Per-patch inference can significantly reduce the peak memory required to execute a sequence of convolutional layers. We study two convolutional layers (stride 1 and 2). Under per-layer computation (a), the first convolution has a large input/output activation size, dominating the peak memory requirement. With per-patch computation (b), we allocate the buffer to host the final output activation, and compute the results \emph{patch-by-patch}. We only need to store the activation from \emph{one patch} but not the entire feature map, reducing the peak memory (the first input is the image, which can be partially decoded from a compressed format like JPEG).
    }
    \label{fig:spatial_partial}
\end{figure*}
\textbf{Breaking the Memory Bottleneck with Patch-based Inference.}
TinEngine breaks the memory bottleneck of the initial layers with \emph{patch-based inference} (Figure~\ref{fig:spatial_partial}).
Existing deep learning inference frameworks (\eg, TensorFlow Lite Micro~\cite{abadi2016tensorflow}, TinyEngine~\cite{lin2020mcunet}, microTVM~\cite{chen2018tvm}, \etc) use a \emph{layer-by-layer} execution. 
For each convolutional layer, the inference library first allocates the input and output activation buffer in SRAM, and releases the input buffer after the \emph{whole} layer computation is finished.
The patch-based inference runs the initial memory-intensive stage in a \emph{patch-by-patch} manner. For each time, it only runs the model on a small spatial region (>10$\times$ smaller than the whole area), which effectively cuts down the peak memory usage. After this stage is finished, the rest of the network with a small peak memory is executed in a normal layer-by-layer manner (upper notations in Figure~\ref{fig:mbv2_mem}).

An example of two convolutional layers (with stride 1 and 2) is shown in Figure~\ref{fig:spatial_partial}. For conventional per-layer computation, the first convolutional layer has large input and output activation size, leading to a high peak memory. With spatial partial computation, it allocates the buffer for the final output and computes its values \emph{patch-by-patch}. In this manner, it only needs to store the activation from \emph{one patch} instead of the \emph{whole feature map}. 

\begin{figure*}[t]
    \centering
     \includegraphics[width=0.9\textwidth]{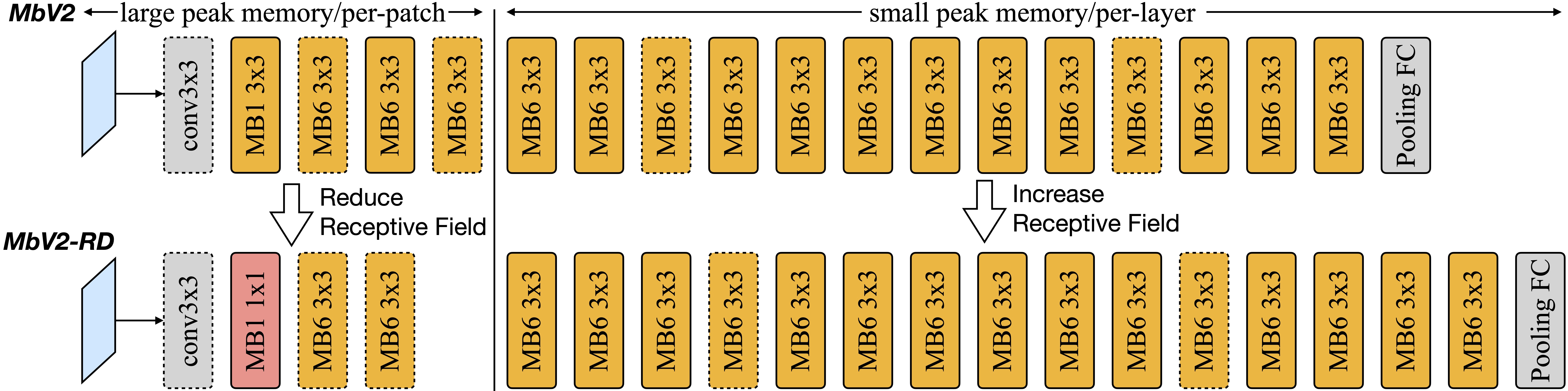}
    \caption{The redistributed MobileNetV2 (MbV2-RD) has reduced receptive field for the per-patch inference stage and increased receptive field for the per-layer stage. The two networks have the same level of performance, but MbV2-RD has a smaller overhead under patch-based inference. The mobile inverted block is denoted as \texttt{MB\{expansion ratio\} \{kernel size\}}. The dashed border means stride=2.
    }
    \label{fig:mbv2_vs_mbv2rd}
\end{figure*}

\textbf{Reducing Computation Overhead by Redistributing the Receptive Field.}
The significant memory saving comes at the cost of computation overhead. To maintain the same output results as per-layer inference, the non-overlapping output patches correspond to overlapping patches in the input image (the shadow area in Figure~\ref{fig:spatial_partial}(b)). This is because convolutional filters with kernel size >1 contribute to increasing receptive fields. 
The computation overhead is related to the receptive field of the patch-based initial stage. Consider the output of the patch-based stage, the larger receptive field it has on the input image, the larger resolution for each patch, leading to a larger overlapping area and repeated computation.
There are some focusing on addressing the issue from the hardware perspective~\cite{alwani2016fused}. However, since such practices may not be general to all devices, TinyEngine solves the problem from the network architecture side. 

MCUNet proposes to \emph{redistribute} the receptive field (RF) of the CNN to reduce computation overhead. The basic idea is: \emph{(1) reduce the receptive field of the patch-based initial stage;  (2) increase the receptive field of the later stage}. Reducing RF for the initial stage helps to reduce the size of each input patch and repeated computation. However, some tasks may have degraded performance if the overall RF is smaller (\eg, detecting large objects). Therefore, it further increases the RF of the later stage to compensate for the performance loss. 
A manually tuned example of MobileNetV2 is shown in Figure~\ref{fig:mbv2_vs_mbv2rd}. After redistributing the receptive field (``MbV2-RD''), the computation overhead is negligible.

MCUNet automates the process with joint search (introduced in the next section).

\subsection{Co-Design: Joint Neural Architecture and Inference Scheduling Search}
\label{sec:joint_search}

\subsubsection{Co-Design Loop}
The optimization algorithms for model architectures and inference engines are tightly coupled. For example, redistributing the receptive field allows us to enjoy the benefit of memory reduction at minimal computation/latency overhead, which allows larger freedom when designing the backbone architecture (\eg, we can now use a larger input resolution). To explore such a large design space, MCUNet jointly optimizes the \emph{neural architecture} and the \emph{inference scheduling} in an automated manner.
Given a certain dataset and hardware constraints (SRAM limit, Flash limit, latency limit, \etc), our goal is to achieve the highest accuracy while satisfying all the constraints.
For the model optimization, it uses NAS to find a good candidate network architecture; for the scheduling optimization, it optimizes the knobs like the patches number $p$ and the number of blocks $n$ to perform patch-based inference, and other knobs in TinyEngine~\cite{lin2020mcunet}.

There are some trade-offs during the co-design. For example, given the same constraints, it can choose to use a smaller model that fits per-layer execution ($p=1$, no computation overhead), or a larger model and per-patch inference ($p>1$, with a small computation overhead). Therefore, MCUNet puts both sides in the same loop and uses evolutionary search to find the best set of $(k_{[~]}, e_{[~]}, d_{[~]}, w_{[~]}, r, p, n)$ satisfying constraints. Therefore, The two dimensions are jointly searched in the same loop with evolutionary search.

\subsubsection{Experimental Results}

\myparagraph{Pushing the ImageNet record on MCUs.} 
\label{sec:exp_image_cls}

With joint optimization of neural architecture and inference scheduling, MCUNet significantly pushes the state-of-the-art results for MCU-based tiny image classification. 

\def\thisgray{\color{gray!75}}

\renewcommand \arraystretch{0.9}
\begin{table*}[t]
    \setlength{\tabcolsep}{5.5pt}
    \caption{\method significantly improves the ImageNet accuracy on microcontrollers, outperforming the state-of-the-arts by \textbf{4.6\%} under 256kB SRAM and \textbf{3.3\%} under 512kB. 
    Lower or mixed precisions (marked \textcolor{gray}{gray}) are orthogonal techniques, which we leave for future work. Out-of-memory (OOM) results are \sout{struck out}.  }
    \vspace{5pt}
    \label{tab:mcu_imagenet}
    \centering
    \small{
     \begin{tabu}{lcccccc}
    \toprule
    Model / Library & Quant. & MACs & SRAM & Flash & Top-1  & Top-5 \\  
  \midrule\midrule
  \multicolumn{7}{c}{\emph{STM32F412 (256kB SRAM, 1MB Flash)}} \\ 
  \midrule
    MbV2 0.35$\times$ ($r$=144)~\cite{sandler2018mobilenetv2} / TinyEngine~\cite{lin2020mcunet} & int8 & 24M & \sout{308kB} & 862kB   & 49.0\% & 73.8\% \\
    Proxyless 0.3$\times$ ($r$=176)~\cite{cai2019proxylessnas} / TinyEngine~\cite{lin2020mcunet} & int8 & 38M & \sout{292kB} & 892kB   & 56.2\% & 79.7\% \\
  \rowfont{\thisgray} MbV1 0.5$\times$ ($r$=192)~\cite{howard2017mobilenets} / Rusci \etal~\cite{rusci2019memory} & mixed & 110M & <256kB & <1MB  & 60.2\%\\
  MCUNet (TinyNAS / TinyEngine)~\cite{lin2020mcunet} & int8 & 68M  & 238kB & 1007kB  & 60.3\% & - \\ 
  \rowfont{\thisgray} MCUNet (TinyNAS / TinyEngine)~\cite{lin2020mcunet} & int4 & 134M & 233kB & 1008kB  & 62.0\% & - \\ 
    \midrule
    \method{-M4 (w/ patch)} & int8 & 119M &  196kB
    & 1010kB  & \textbf{64.9\%} & \textbf{86.2\%} \\
    \midrule \midrule
    \multicolumn{7}{c}{\emph{STM32H743 (512kB SRAM, 2MB Flash)}} \\
  \midrule
  \rowfont{\thisgray} MbV1 0.75$\times$ ($r$=224)~\cite{howard2017mobilenets} / Rusci \etal~\cite{rusci2019memory} & mixed & 317M &  <512kB & <2MB   & 68.0\%\\
   MCUNet (TinyNAS / TinyEngine)~\cite{lin2020mcunet} & int8 & 126M  & 452kB & 2014kB  & 68.5\% & - \\ 
  \rowfont{\thisgray} MCUNet (TinyNAS / TinyEngine)~\cite{lin2020mcunet} & int4 & 474M &  498kB & 2000kB  & 70.7\% & - \\ 
  \midrule
  \method{-H7 (w/ patch)} & int8 & 256M & 465kB & 2032kB & \textbf{71.8\%} & \textbf{90.7\%} \\
\bottomrule
\end{tabu}
}
\end{table*}

We compared MCUNet with existing state-of-the-art solutions on ImageNet classification under two hardware settings: 256kB SRAM/1MB Flash and 512kB SRAM/2MB Flash. The former represents a widely used Cortex-M4 microcontroller; the latter corresponds to a  higher-end Cortex-M7. 
The goal is to achieve the highest ImageNet accuracy on resource-constrained MCUs (Table~\ref{tab:mcu_imagenet}). 
\method significantly improves the ImageNet accuracy of tiny deep learning on microcontrollers. Under 256kB SRAM/1MB Flash, \method outperforms the state-of-the-art method~\cite{lin2020mcunet} by 4.6\% at 18\% lower peak SRAM. Under 512kB SRAM/2MB Flash, MCUNet achieves a new \emph{record} ImageNet accuracy of 71.8\% on commercial microcontrollers, which is 3.3\% compared to the best solution under the same quantization policy. 
Lower-bit (\texttt{int4}) or mixed-precision quantization can further improve the accuracy (marked in \textcolor{gray}{gray} in the table).  

\begin{figure*}[t]
    \centering
     \includegraphics[width=0.85\textwidth]{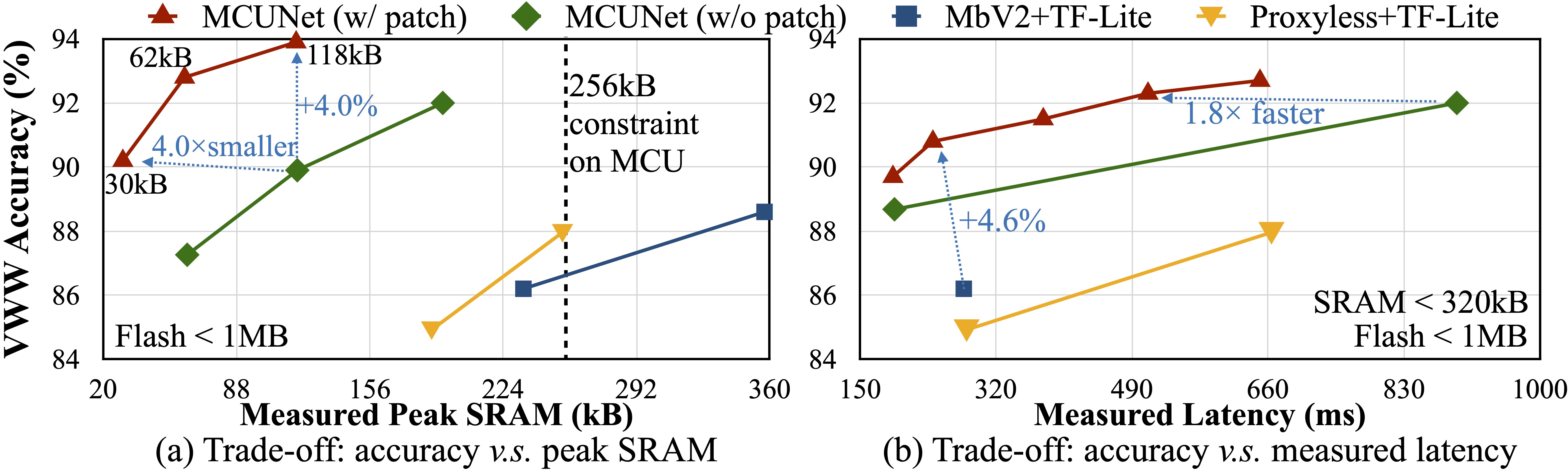}
    \caption{
    \textbf{Left}: \method has better visual wake word (VWW) accuracy \vs peak SRAM trade-off. Compared to MCUNet~\cite{lin2020mcunet}, \method achieves better accuracy at 4.0$\times$ smaller peak memory. It achieves >90\% accuracy under <32kB memory, facilitating deployment on extremely small hardware.
    \textbf{Right}: patch-based method expands the search space that can fit the MCU, allowing better accuracy \vs latency trade-off.%
    }
    \label{fig:vww_curves}
\end{figure*}

\myparagraph{Visual Wake Words under 32kKB SRAM.} Visual wake word (VWW) reflects the low-energy application of TinyML. \method allows running a VWW model with a modest memory requirement. As in Figure~\ref{fig:vww_curves}\footnote{Note that MCUNetV2 refers to the version w/ patch-based inference, while MCUNet refers to per-layer inference.}, \method outperforms state-of-the-art method~\cite{lin2020mcunet} for both accuracy \vs peak memory and accuracy \vs latency trade-off. Compared to per-layer inference, \method can achieve better accuracy using 4.0$\times$ smaller memory. Actually, it can achieve >90\% accuracy under 32kB SRAM requirement, allowing model deployment on low-end MCUs like STM32F410 costing only \$1.6. 
Per-patch inference also expands the search space, giving us more freedom to find models with better accuracy \vs latency trade-off.

\myparagraph{MCU-based detection on Pascal VOC.}
Object detection is sensitive to a smaller input resolution~\cite{lin2021mcunetv2}. 
Current state-of-the-art~\cite{lin2020mcunet} cannot achieve a decent detection performance on MCUs due to the resolution bottleneck. \method breaks the memory bottleneck for detectors and improves the mAP by double digits. 

\renewcommand \arraystretch{0.9}
\begin{table*}[t]
    \setlength{\tabcolsep}{4pt}
    \caption{\method significantly improves Pascal VOC~\cite{everingham2010pascal} object detection on MCU by allowing a higher input resolution. Under STM32H743 MCU constraints, \method{-H7} improves the mAP by 16.9\% compared to~\cite{lin2020mcunet}, achieving a record performance on MCU. It can also scale down to cheaper MCU STM32F412 with only 256kB SRAM while still improving mAP by 13.2\% at 1.9$\times$ smaller peak SRAM and a similar computation. %
    } %
    \label{tab:mcu_det}
    \centering
    \small{
     \begin{tabular}{lllcccccc}
    \toprule
  MCU Model & Constraint & Model  & \#Param  &  MACs & peak SRAM  & VOC mAP & Gain \\
    \midrule 
\multirow{3}{*}{H743 ($\sim$\$7)} & \multirow{3}{*}{\shortstack{SRAM\\<512kB}} & MbV2+CMSIS~\cite{lin2020mcunet}  &  0.87M & 34M & \sout{519kB} & 31.6\% & -\\ 
 & &MCUNet~\cite{lin2020mcunet}  & 1.20M &168M & 466kB & 51.4\%  & 0\%  \\ \cmidrule(lr){3-8}
& & \method{-H7}  &  0.67M & 343M &438kB  & \textbf{68.3\%} & +16.9\% \\  \midrule
F412 ($\sim$\$4) & <256kB & \method{-M4}    & 0.47M & 172M &\textbf{247kB} &  64.6\% & +13.2\% \\
      \bottomrule
     \end{tabular}
     }
\end{table*}

The object detection results on Pascal VOC trained with YOLOv3~\cite{redmon2018yolov3} are shown in Table~\ref{tab:mcu_det}, including results for M4 MCU with 256kB SRAM and H7 MCU with 512kB SRAM. On H7 MCU, \method{-H7} improves the mAP by 16.7\% compared to the state-of-the-art method MCUNet~\cite{lin2020mcunet}. It can also scale down to fit a cheaper commodity Cortex-M4 MCU with only 256kB SRAM, while still improving the mAP by 13.2\% at 1.9$\times$ smaller peak SRAM.
Note that \method{-M4} shares a similar computation with MCUNet  (172M \vs 168M) but a much better mAP. This is because the expanded search space from patch-based inference allows us to choose a better configuration of larger input resolution and smaller models.

\myparagraph{Memory-efficient face detection.}
\def\noparam{1}
\renewcommand \arraystretch{0.9}
\begin{table*}[t]
    \setlength{\tabcolsep}{5pt}
    \caption{\method outperforms existing methods for memory-efficient face detection on WIDER FACE~\cite{yang2016wider} dataset. 
    Compared to RNNPool-Face-C~\cite{saha2020rnnpool}, \method{-L} can achieve similar mAP at \textbf{3.4$\times$} smaller peak SRAM and \textbf{1.6$\times$} smaller computation.
The model statistics are profiled on $640\times480$ RGB input images following~\cite{saha2020rnnpool}.
}
    \label{tab:wider_face}
    \centering
    \small{
     \begin{tabular}{lcccccccccc}
    \toprule
  \multirow{2}{*}{Method}  & \multirow{2}{*}{MACs $\downarrow$} &
  \multirow{2}{*}{Peak Memory $\downarrow$}&  \multicolumn{3}{c}{mAP $\uparrow$} &  \multicolumn{3}{c}{mAP ($\leq$3 faces) $\uparrow$} \\  \cmidrule(lr){4-6}\cmidrule(lr){7-9}
  & & (\texttt{fp32}) & Easy & Medium & Hard &  Easy & Medium & Hard \\ \midrule\midrule
  EXTD~\cite{yoo2019extd} & 8.49G & \ifx\noparam\undefined 0.07M  & \fi 18.8MB (9.9$\times$) & 0.90 & 0.88 & 0.82 & 0.93 & 0.93 & 0.91\\
  LFFD~\cite{he2019lffd} & 9.25G & \ifx\noparam\undefined  2.15M & \fi  18.8MB (9.9$\times$) & 0.91 & 0.88 & 0.77 & 0.83 & 0.83 & 0.82 \\
  RNNPool-Face-C~\cite{saha2020rnnpool} & 1.80G & \ifx\noparam\undefined 1.52M & \fi 6.44MB (3.4$\times$)   &\textbf{0.92} & 0.89 & \textbf{0.70}  & \textbf{0.95} & \textbf{0.94} & \textbf{0.92} \\ \midrule
  \method{-L} & \textbf{1.10G} & \textbf{1.89MB} (1.0$\times$)  & \textbf{0.92} & \textbf{0.90} & \textbf{0.70} & 0.94 & 0.93 & \textbf{0.92} \\  \midrule\midrule
  EagleEye~\cite{zhao2019real} & \textbf{0.08G} & \ifx\noparam\undefined 0.23M & \fi 1.17MB (1.8$\times$) & 0.74 & 0.70 & 0.44 & 0.79 & 0.78 & 0.75 \\
  RNNPool-Face-A~\cite{saha2020rnnpool} & 0.10G & \ifx\noparam\undefined 0.06M & \fi 1.17MB (1.8$\times$) & 0.77 & 0.75 & 0.53 & 0.81 & 0.79 & 0.77\\
\midrule
  \method{-S} & 0.11G & \ifx\noparam\undefined 0.13M &\fi  \textbf{672kB} (1.0$\times$)  & \textbf{0.85} & \textbf{0.81} & \textbf{0.55} & \textbf{0.90} & \textbf{0.89} & \textbf{0.87}\\
    \bottomrule 
     \end{tabular}
     }
\end{table*}

The performance of \method for memory-efficient face detection on WIDER FACE~\cite{yang2016wider} dataset are shown in Table~\ref{tab:wider_face}. The analytic memory usage of the detector backbone in \texttt{fp32} is reported following~\cite{saha2020rnnpool}. The models are trained with  S3FD face detector~\cite{zhang2017s3fd} following~\cite{saha2020rnnpool} for a fair comparison. 
The reported mAP is calculated on samples with $\leq$ 3 faces, which is a more realistic setting for tiny devices.  
\method outperforms existing solutions under different scales.  \method{-L} achieves comparable performance at 3.4$\times$ smaller peak memory compared to RNNPool-Face-C~\cite{lin2020mcunet} and 9.9$\times$ smaller peak memory compared to LFFD~\cite{he2019lffd}. The computation is also 1.6$\times$ and 8.4$\times$ smaller.
\method{-S} consistently outperforms RNNPool-Face-A~\cite{saha2020rnnpool} and EagleEye~\cite{zhao2019real} at 1.8$\times$ smaller peak memory. 

\def\engine{Tiny Training Engine\xspace}
\def\engineshort{TTE\xspace}
\section{Tiny Training}
\label{tiny_training}

In addition to inference, tiny on-device training is a growing direction that allows us to \emph{adapt} the pre-trained model to newly collected sensory data \emph{after} deployment.  By training and adapting \emph{locally} on the edge, the model can learn to continuously improve its predictions and perform lifelong learning and user customization. By bringing training closer to the sensors, it also helps to protect user privacy when handling sensitive data (\eg, healthcare). 

However, on-device training on tiny edge devices is extremely challenging and fundamentally different from cloud training. Tiny IoT devices (\eg, microcontrollers) typically have a limited SRAM size like 256KB. Such a small memory budget is hardly enough for the \emph{inference} of deep learning models~\cite{lin2020mcunet, lin2021mcunetv2, banbury2021micronets, burrello2021dory, liberis2020mu, fedorov2019sparse, liberis2019neural, rusci2019memory}, let alone the \emph{training}, which requires extra computation for the backward and extra memory for intermediate activation~\cite{chen2016training}. On the other hand, modern deep training frameworks (\eg, PyTorch~\cite{pytorch2019}, TensorFlow~\cite{tensorflow2015}) are usually designed for cloud servers and require a large memory footprint (>300MB) even when training a small model (\eg, MobileNetV2-w0.35~\cite{sandler2018mobilenetv2}) with batch size 1. 
The huge gap (>1000$\times$) makes it impossible to run on tiny IoT devices. Furthermore, devices like microcontrollers are bare-metal and do not have an operational system and the runtime support needed by existing training frameworks. Therefore, we need to jointly design the \emph{algorithm} and the \emph{system} to enable tiny on-device training.

Deep learning training systems such as PyTorch~\cite{pytorch2019}, TensorFlow~\cite{tensorflow2015}, JAX~\cite{jax2018github}, MXNet~\cite{chen2015mxnet}, \etc do not consider the tight resources on edge devices. Edge deep learning inference frameworks like TVM~\cite{chen2018tvm}, TF-Lite~\cite{tflite}, NCNN~\cite{tensorRT}, \etc provide a slim runtime, but lack the support for back-propagation. There are low-cost efficient transfer learning algorithms like training only the final classifier layer, bias-only update~\cite{cai2020tinytl}, \etc. However, the existing training systems can not realize the theoretical saving into measured savings. The downstream accuracy of such update schemes is also low (Figure~\ref{fig:acc_vs_mem}). There is a need for training systems that can effectively utilize the limited resources on edge devices."

\begin{figure*}
    \centering
    \includegraphics[width=0.9\textwidth]{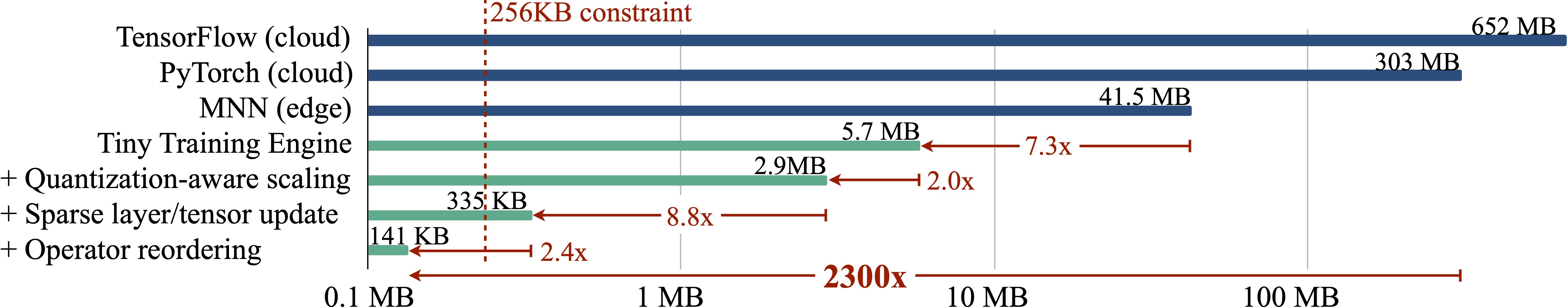}
    \caption{Algorithm and system co-design reduces the training memory from 303MB (PyTorch) to 149KB with the same transfer learning accuracy, leading to 2077$\times$ reduction. The numbers are measured with MobilenetV2-w0.35~\cite{sandler2018mobilenetv2}, batch size 1 and resolution 128$\times$128. It can be deployed to a microcontroller with 256KB SRAM.
    }
    \label{fig:teaser}
\end{figure*}

In order to bridge the gap and enable tiny on-device training with algorithm-system co-design, there are two unique challenges in tiny on-device training: (1) the model is quantized on edge devices. A \emph{real} quantized graph is difficult to optimize due to mixed-precision tensors and the lack of Batch Normalization layers~\cite{ioffe2015batch}; (2) the limited hardware resource (memory and computation) of tiny hardware does not allow full back-propagation, as the memory usage can easily exceed the SRAM of microcontrollers by more than an order of magnitude. 
To cope with the difficulties, TinyTraining proposes the following designs

\subsection{Quantization Aware Scaling}

Neural networks usually need to be quantized to fit the limited memory of edge devices~\cite{lin2020mcunet, jacob2018quantization}. 
For a \texttt{fp32} linear layer
$\mathbf{y}_{\texttt{fp32}} = \mathbf{W}_{\texttt{fp32}}\mathbf{x}_{\texttt{fp32}} + \mathbf{b}_{\texttt{fp32}}$, the \texttt{int8} quantized counterpart is:
\begin{equation}  \label{eq:quantize_forward}
\mathbf{\bar{y}}_{\texttt{int8}} = \texttt{cast2int8}[s_{\texttt{fp32}} \cdot(\mathbf{\bar{W}}_{\texttt{int8}}\mathbf{\bar{x}}_{\texttt{int8}} + \mathbf{\bar{b}}_{\texttt{int32}})],
\end{equation}\useshortskip
where $\bar{\cdot}$ denotes the tensor being quantized to fixed-point numbers, and $s$ is a floating-point scaling factor to project the results back into \texttt{int8} range. 
The gradient update for the weights can be presented as:
$\mathbf{\bar{W}^\prime}_{\texttt{int8}} = \texttt{cast2int8}(\mathbf{\bar{W}_{\texttt{int8}}} - \alpha\cdot \mathbf{G_\mathbf{\bar{W}}})$,
where $\alpha$ is the learning rate, and $\mathbf{G_\mathbf{\bar{W}}}$ is the gradient of the weights. After applying the gradient update, the weights are rounded back to 8-bit integers.


\begin{figure*}[b]
    \centering
     \includegraphics[width=0.85\textwidth]{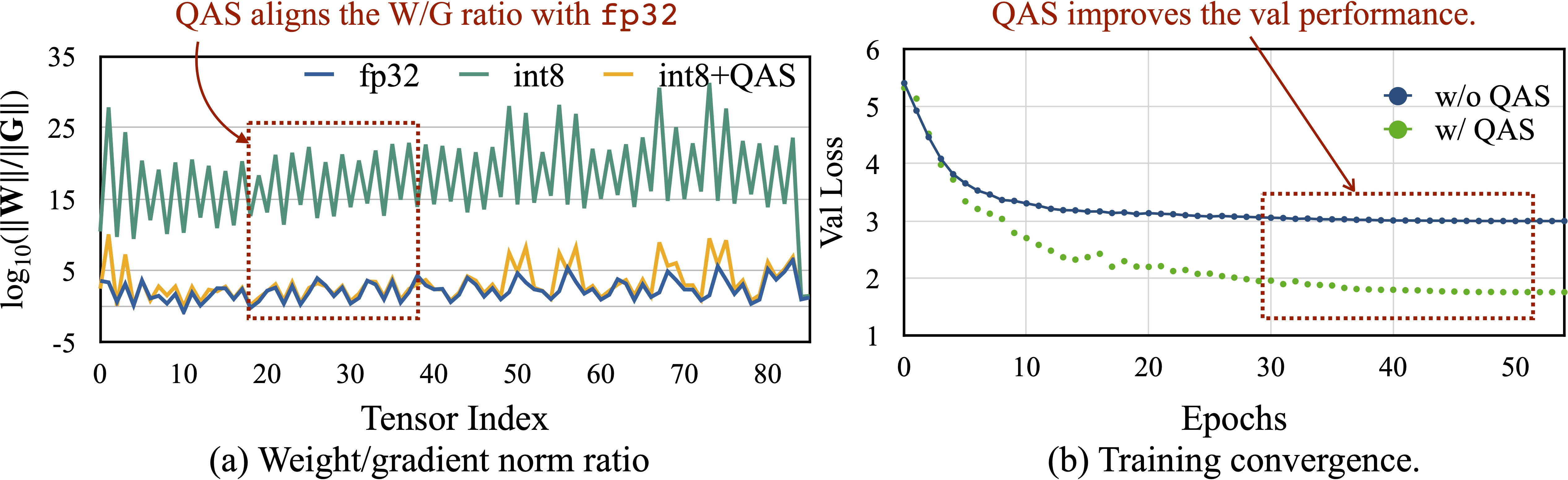}
    \caption{
    \textit{Left}: The quantized model has a very different weight/gradient norm ratio (\ie, $\lVert\mathbf{W}\rVert/\lVert\mathbf{G}\rVert$) compared to the floating-point model at training time. QAS stabilizes the $\lVert\mathbf{W}\rVert/\lVert\mathbf{G}\rVert$  ratio and helps optimization. 
    \textit{Right}: The validation loss curves w/ and w/o QAS. QAS effectively helps convergence, leading to better accuracy. The results are from updating the last two blocks of the MCUNet model on the Cars dataset.
    }
    \label{fig:per_layer_wg_rate}
    \label{fig:training_curves_short}
\end{figure*}
\noindent\textbf{Gradient Scale Mismatch}:
\label{sec:opt_mixed_prec}
Unlike fine-tuning floating-point model on the cloud, training with \textit{a real} quantized graph\footnote{Note that this is contrary to the \textit{fake} quantization 
graph, which is widely used in quantization-aware training~\cite{jacob2018quantization}.} is difficult: the quantized graph has tensors of different bit-precisions (\texttt{int8}, \texttt{int32}, \texttt{fp32}, shown in Equation~\ref{eq:quantize_forward})
and lacks Batch Normalization~\cite{ioffe2015batch} layers (fused), leading to unstable gradient update.

Optimizing a quantized graph often leads to lower accuracy compared to the floating-point counterpart.. A possible hypothesis is that the quantization process distorts the gradient update. 
To verify the idea, Figure~\ref{fig:per_layer_wg_rate} plot the ratio between weight norm and gradient norm (\ie, $\lVert \mathbf{W}\rVert/\lVert\mathbf{G}\rVert$) for each tensor at the beginning of the training on the CIFAR dataset~\cite{krizhevsky2009learning}. The ratio curve is very different after quantization: (1) the ratio is much larger (could be addressed by adjusting the learning rate); (2) the ratio has a different pattern after quantization. Take the highlighted area (red box) as an example, the quantized ratios have a zigzag pattern, differing from the floating-point curve.
If all the tensors are updated with a fixed learning rate, then the update speed of each tensor would be very different compared to the floating-point case, leading to inferior accuracy. Even adaptive-learning rate optimizers like Adam~\cite{kingma2014adam} cannot fully address the issue, as shown in Table~\ref{tab:optimizer_study}.

\begin{table*}[t]
    \caption{Updating real quantized graphs (\texttt{int8}) with SGD is difficult: the transfer learning accuracy falls behind the floating-point counterpart (\texttt{fp32}), even with adaptive learning rate optimizers like Adam~\cite{kingma2014adam} and LARS~\cite{you2017large}. QAS helps to bridge the accuracy gap without memory overhead (slightly higher). The numbers are for updating the last two blocks of MCUNet-5FPS~\cite{lin2020mcunet} model.
    }
    \label{tab:optimizer_study}
    \centering
    \small{
     \begin{tabular}{llccccccccc}
    \toprule
 \multirow{2}{*}{Precision} & \multirow{2}{*}{Optimizer} &  \multicolumn{8}{c}{Accuracy (\%) (MCUNet backbone: 23M MACs, 0.48M Param )} & \multirow{2}{*}{\shortstack{Avg\\Acc.}} \\ \cmidrule(lr){3-10}
 & & Cars & CF10 & CF100 & CUB & Flowers & Food & Pets & VWW \\
\midrule
\texttt{fp32} & SGD-M & 56.7 & 86.0 & 63.4 & 56.2 & 88.8 & 67.1 & 79.5 & 88.7 & 73.3 \\ \midrule 
 \multirow{4}{*}{\texttt{int8}} & SGD-M &  31.2 & 75.4 & 54.5 & 55.1 & 84.5 & 52.5 & 81.0 & 85.4 & 64.9 \\ %
 & Adam~\cite{kingma2014adam} &  54.0 & 84.5 & 61.0 & 58.5 & 87.2 & 62.6 & 80.1 & 86.5 & 71.8  \\
 & LARS~\cite{you2017large} & 5.1 & 64.8 & 39.5 & 9.6 & 28.8 & 46.5 & 39.1 & 85.0 & 39.8  \\  \cmidrule(lr){2-11}
& SGD-M+QAS & 55.2 & 86.9 & 64.6 & 57.8 & 89.1 & 64.4 & 80.9 & 89.3 & \textbf{73.5} \\
    \bottomrule
     \end{tabular}
     }
     \vspace{-10pt}
\end{table*}

\noindent\textbf{Hyperparameter-Free Gradient Scaling.} 
To address the problem, a hyper-parameter-free learning rate scaling rule, QAS, is proposed.
Consider a 2D weight matrix of a linear layer $\mathbf{W}\in \mathds{R}^{c_1\times c_2}$, where $c_1, c_2$ are the input and output channel. To perform per-tensor quantization\footnote{For simplicity. In practice, per-channel quantization~\cite{jacob2018quantization} is used and the scaling factor is a vector of size $c_2$.}, a scaling rate $s_\mathbf{W}\in\mathds{R}$ is computed, such that $\mathbf{\bar{W}}$'s largest magnitude is $2^7-1=127$: %
\begin{equation}
\mathbf{W} = s_\mathbf{W}\cdot(\mathbf{W} / s_\mathbf{W}) \stackrel{\text{quantize}}{\approx} s_\mathbf{W} \cdot \mathbf{\bar{W}}, \quad \mathbf{G_{\bar{W}}} \approx s_\mathbf{W}\cdot \mathbf{G_W},
\end{equation}
The process (roughly) preserves the mathematical functionality during the forward (Equation~\ref{eq:quantize_forward}), but it distorts the magnitude ratio between the weight and its corresponding gradient:
\begin{equation}\label{eq:wg_ratio}
  \lVert \mathbf{\bar{W}} \rVert / \lVert \mathbf{G_{\bar{W}}} \rVert \approx \lVert \mathbf{W}  /s_\mathbf{W} \rVert / \lVert s_\mathbf{W}\cdot \mathbf{G_W}\rVert = s_\mathbf{W}^{-2} \cdot \lVert\mathbf{W}\rVert / \lVert\mathbf{G}\rVert.
\end{equation}
The weight and gradient ratios are off by  $s_\mathbf{W}^{-2}$, leading to the distorted pattern in Figure~\ref{fig:per_layer_wg_rate}: (1) the scaling factor is far smaller than 1, making the weight-gradient ratio much larger; (2) weights and biases have different data type (\texttt{int8} \vs \texttt{int32}) and thus have scaling factors of very different magnitude, leading to the zigzag pattern.
To solve the issue, Quantization-Aware Scaling (QAS) is proposed by compensating the gradient of the quantized graph according to Equation~\ref{eq:wg_ratio}:
\begin{equation} \label{eq:qas}
    \mathbf{\Tilde{G}_{\bar{W}}} = \mathbf{G_{\bar{W}}} \cdot s_\mathbf{W}^{-2}, \quad 
    \mathbf{\Tilde{G}_{\bar{b}}} = \mathbf{G_{\bar{b}}} \cdot s_\mathbf{W}^{-2} \cdot s_{\mathbf{x}}^{-2} = \mathbf{G_{\bar{b}}} \cdot s^{-2}
\end{equation}
where $s_{\mathbf{X}}^{-2}$ is the scaling factor for quantizing input $\mathbf{x}$ (a scalar following~\cite{jacob2018quantization}, note that $s=s_\mathbf{W} \cdot s_{\mathbf{x}}$ in Equation~\ref{eq:quantize_forward}).
 $\lVert \mathbf{W}\rVert/\lVert\mathbf{G}\rVert$ curve with QAS is plotted in Figure~\ref{fig:per_layer_wg_rate} (int8+scale). After scaling, the gradient ratios  match the floating-point counterpart. It also improves transfer learning accuracy (Table~\ref{tab:optimizer_study}), matching the accuracy of the floating-point counterpart without incurring memory overhead.   

\noindent\textbf{Experiment Results.}
The last two blocks in Table~\ref{tab:optimizer_study} show the fine-tuning results (simulating low-cost fine-tuning) of MCUNet on various downstream datasets. With momentum SGD, the training accuracy of the quantized model (\texttt{int8}) falls behind the floating-point counterpart due to the difficulty in optimization.
Adaptive learning rate optimizers like Adam~\cite{kingma2014adam} can improve the accuracy, but it is still lower than the \texttt{fp32} fine-tuning results. It also consumes 3 times more memory due to second-order momentum, which is not desired in TinyML settings. LARS~\cite{you2017large} does not converge well on most datasets despite extensive hyperparameter tuning (of both the learning rate and the "trust coefficient"). The aggressive gradient scaling rule of LARS makes the training unstable.
The accuracy gap is closed when applying QAS, achieving the same accuracy as floating-point training with no extra memory cost.
Figure~\ref{fig:training_curves_short} shows the training curve of TinyTraining on the Cars dataset with and without QAS. QAS effectively improves optimization.

\subsection{Memory-Efficient Sparse  Update} 
\label{sec:method:sparse_update}
\begin{figure*}
    \centering
    \includegraphics[width=0.85\linewidth]{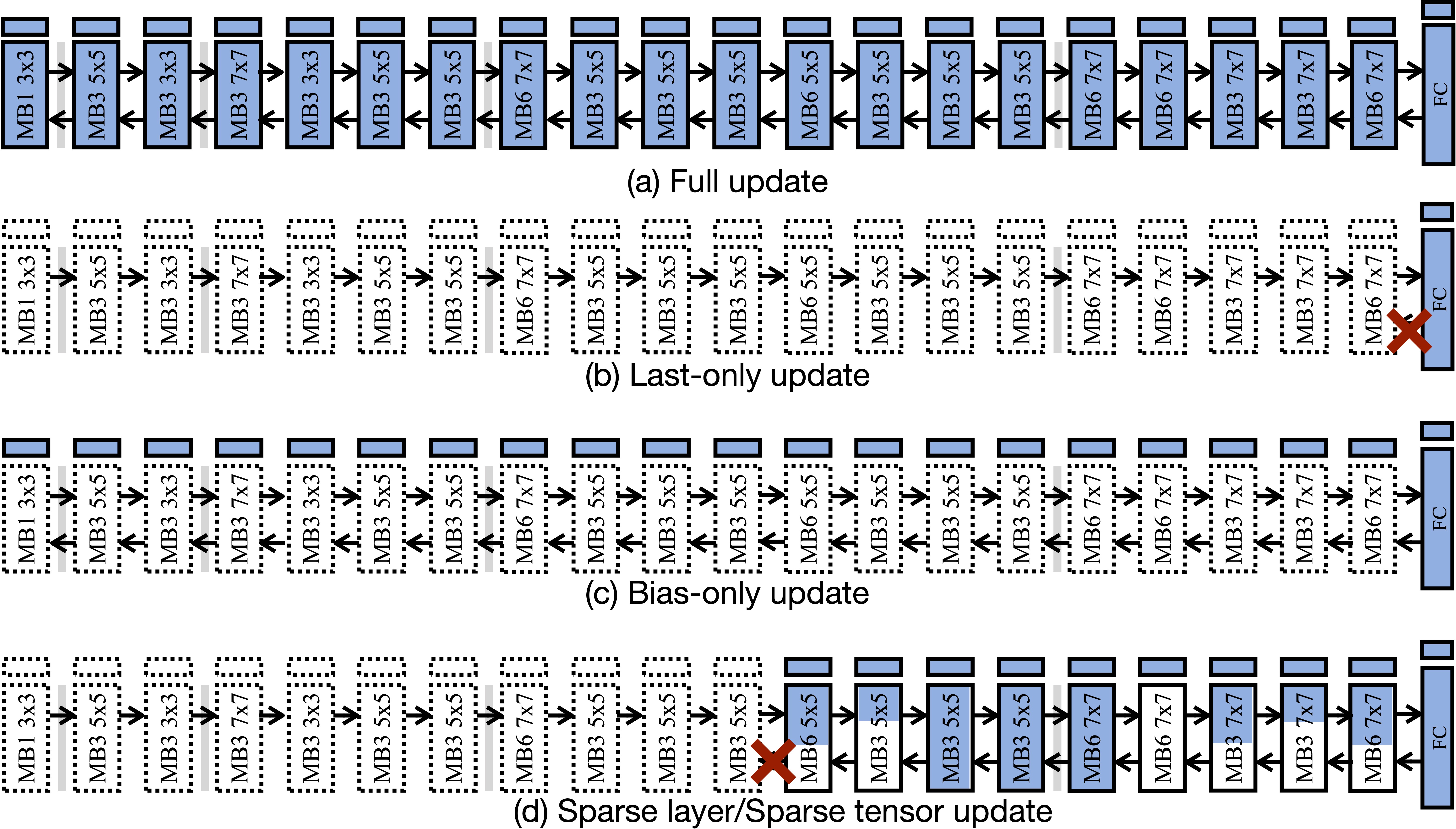}
    \caption{
        Different learning schemes on ProxylessNAS-Mobile~\cite{cai2019proxylessnas}. Full update (a) consumes a lot of memory thus cannot fit TinyML. Efficient learning methods like last-only (b) / bias-only (c) save the memory but cannot match the baseline performance. Sparse update (d) only performs partial back-propagation, leading to less memory usage and computation with comparable accuracy on downstream tasks.
    }
    \label{fig:different_update_scheme}
\end{figure*}


\begin{figure*}[t]
    \centering
     \includegraphics[width=0.85\textwidth]{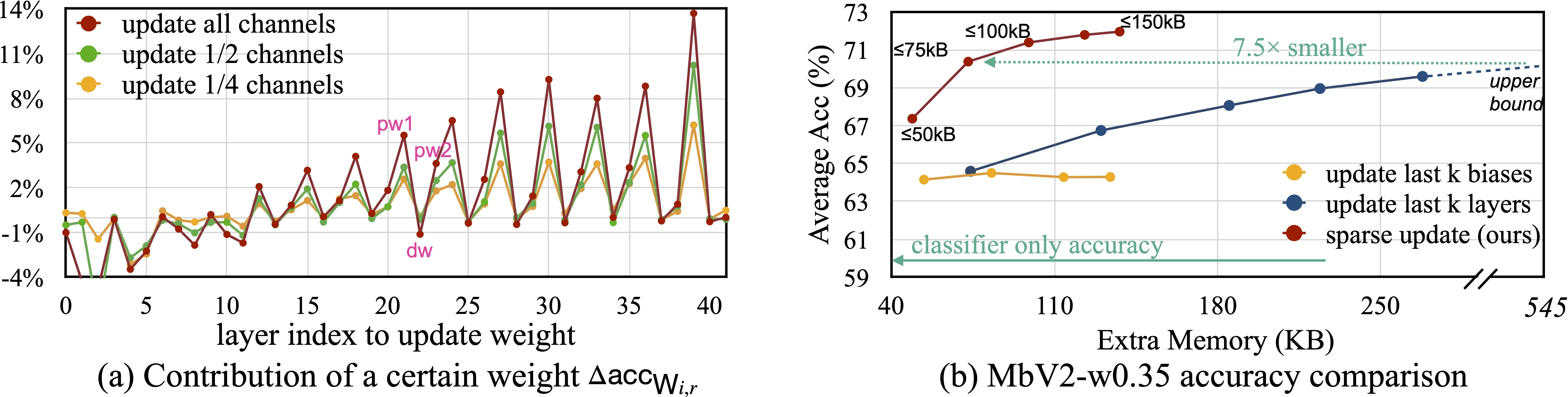}
    \caption{
    \textit{Left}: Contribution analysis of updating biases and weights.  For updating the weight of a specific layer, the later layers appear to be more important; the first point-wise conv (pw1) in an inverted bottleneck block~\cite{sandler2018mobilenetv2} appears to be more important; and the gains are bigger with more channels updated.
    \textit{Right}: Sparse update can achieve higher transfer learning accuracy using 4.5-7.5$\times$ smaller extra memory (analytic) compared to updating the last $k$ layers. For classifier-only update, the accuracy is low due to limited capacity. Bias-only update can achieve a higher accuracy but plateaus soon. 
    }
    \label{fig:sensitivity_curve}
    \label{fig:acc_vs_mem}
\end{figure*}

Though QAS makes optimizing a quantized model possible, updating the whole model (or even the last several blocks) requires a large amount of memory, which is not affordable for the TinyML setting. To address this, sparsely updating the layers and the tensors is proposed

\noindent\textbf{Sparse Layer/Tensor Update.} 
Pruning techniques prove to be quite successful for achieving sparsity and reducing model size~\cite{han2016deep, he2018amc, lin2017runtime, he2017channel, liu2017learning, liu2019metapruning}. 
Instead of pruning \emph{weights} for inference, the \emph{gradient} during backpropagation, and updating the model sparsely are pruned. Given a tight memory budget,  updates of the \emph{less important} parameters are skipped to reduce memory usage and computation cost. 
When updating a linear layer $\mathbf{y}=\mathbf{W}\mathbf{x}+\mathbf{b}$ (similar analysis applies to convolutions), the gradient update is $\mathbf{G_W}=f_1(\mathbf{G_y}, \mathbf{x})$ and $\mathbf{G_b}=f_2(\mathbf{G_y})$, given the output gradient $\mathbf{G_y}$ from the later layer
Notice that updating the biases does not require saving the intermediate activation $\mathbf{x}$, leading to a lighter memory footprint~\cite{cai2020tinytl}\footnote{If many layers are updated, the intermediate activation could consume a large memory~\cite{chen2016training}.}; while updating the weights is more memory-intensive but also more expressive. For hardware like microcontrollers,  an extra copy is needed for the updated parameters since the original ones are stored in read-only FLASH~\cite{lin2020mcunet}.
Given the different natures of updating rules, three aspects of the sparse update rule are considered  (Figure~\ref{fig:different_update_scheme}):
(1) \emph{Bias update}: how many layers should be backpropagated to and update the biases (bias update is cheap,  the bias terms can be always updated if the layer is  backpropagated).
(2) \emph{Sparse layer update}: select a subset of layers to update the corresponding weights.
(3) \emph{Sparse tensor update}:  further allow updating a subset of weight channels to reduce the cost. %

However, finding the right sparse update scheme under a memory budget is challenging due to the large combinational space. For MCUNet~\cite{lin2020mcunet} model with 43 convolutional layers and weight update ratios from \{0, 1/8, 1/4, 1/2, 1\},
the combination is about $10^{30}$, making exhaustive search impossible.

\noindent\textbf{Automated Selection with Contribution Analysis.} 
\emph{contribution analysis} is proposed to automatically derive the sparse update scheme by counting the contribution of each parameter (weight/bias) to the downstream accuracy.
Given a convolutional neural network with $l$ layers, the accuracy improvement is measured from (1) biases: the improvement of updating \emph{last} $k$ biases $\textbf{b}_{l}, \textbf{b}_{l-1},..., \textbf{b}_{l-k+1}$ (bias-only update) compared to only updating the classifier, defined as $\Delta\text{acc}_{\textbf{b}[:k]}$; (2) weights: the improvement of updating the weight of one extra layer $\textbf{W}_i$ (with a channel update ratio $r$) compared to bias-only update, defined as $\Delta\text{acc}_{\textbf{W}i, r}$. An example of the contribution analysis can be found in Figure~\ref{fig:sensitivity_curve} Left (MCUNet on Cars~\cite{krause20133d} dataset; 
After finding $\Delta\text{acc}_{\textbf{b}[:k]}$ and $\Delta\text{acc}_{\textbf{W}i}$ ($1\leq k, i\leq l$), an optimization problem is solved to find:
\begin{align}
\begin{split}
    k^*, \mathbf{i}^*, \mathbf{r}^* &= \max_{k, \mathbf{i}, \mathbf{r}} (\Delta\text{acc}_{\textbf{b}[:k]} + \sum_{i\in\mathbf{i}, r\in\mathbf{r}} \Delta\text{acc}_{\textbf{W}i, r}) \quad \\
    &\text{s.t. Memory}(k, \mathbf{i}, \mathbf{r}) \leq \text{constraint},
\end{split}
\end{align}
where $\mathbf{i}$ is a collection of layer indices whose weights are updated, and $\mathbf{r}$ is the corresponding update ratios (1/8, 1/4, 1/2, 1). Intuitively, by solving this optimization problem,  the combination of (\#layers for bias update is found, the subset of weights to update), such that the total contribution is maximized  while the memory overhead does not exceed the constraint. The problem can be efficiently solved with the evolutionary search. 
Sparse update assumes that the accuracy contribution of each tensor ($\Delta\text{acc}$) can be summed up. Such an approximation proves to be quite effective in our experiments.

\noindent\textbf{Sparse Update Obtains Better Accuracy at Lower Memory.}
The performance of our searched sparse update schemes is compared to two baseline methods: fine-tuning only the biases of the last $k$ layers and fine-tuning the weights and biases of the last $k$ layers. For each configuration, the average accuracy is measured on 8 downstream datasets, and the extra memory usage is calculated analytically. Figure~\ref{fig:acc_vs_mem} compares the results with a simple baseline of fine-tuning only the classifier. The accuracy of classifier-only update is low due to the limited learning capacity. Updating only the classifier is not enough; the backbone also needs updates. Bias-only update outperforms classifier-only update, but the accuracy quickly plateaus and does not improve even when more biases are tuned. For updating the last $k$ layers, the accuracy generally improves as more layers are tuned; however, it has a very large memory footprint. For example, updating the last two blocks of MCUNet leads to an extra memory usage exceeding 256KB, making it infeasible for IoT devices/microcontrollers.
Our sparse update scheme can achieve higher downstream accuracy at a much lower memory cost. Compared to updating the last $k$ layers, the sparse update can achieve higher downstream accuracy with 4.5-7.5 times smaller memory overhead
The highest accuracy is achievable by updating the last $k$ layers\footnote{Note that fine-tuning the entire model does not always lead to the best accuracy. The best $k$ on Cars dataset is obtained via grid search: $k=$36 for MobileNetV2, 39 for ProxylessNAS, 12 for MCUNet, and apply it to all datasets.} as the baseline upper bound (denoted as "upper bound"). Interestingly, our sparse update achieves a better downstream accuracy compared to the baseline best statistics. The sparse update scheme alleviates over-fitting or makes momentum-free optimization easier.

\subsection{\engine (\engineshort)}
\label{sec:engine}
\definecolor{fig5gray}{rgb}{0.80, 0.85, 0.89}
\definecolor{fig5red}{rgb}{1.00, 0.59, 0.55}
\definecolor{fig5yellow}{rgb}{1.00, 0.68, 0.00}

\begin{figure*}[t]
    \centering
     \includegraphics[width=0.9\textwidth]{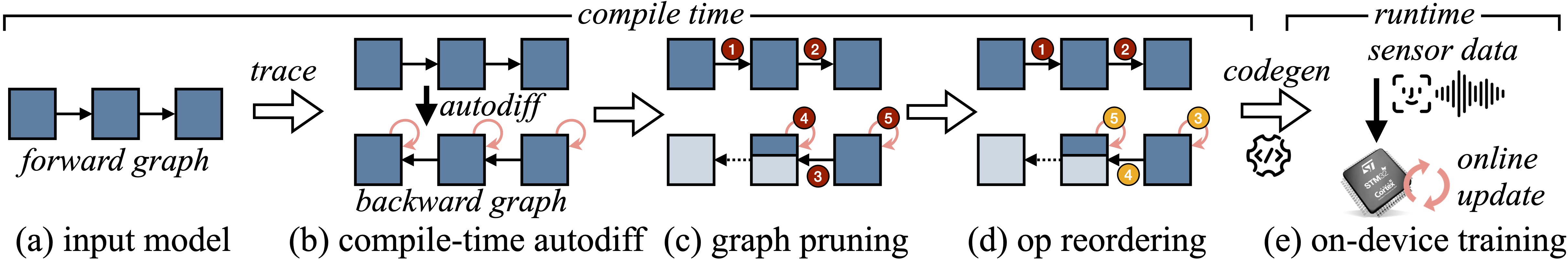}
    \caption{
    The workflow of our \engine (\engineshort). 
    \textbf{(a,b)} Our engine traces the forward graph for a given model and derives the corresponding backward graph at compile time. The {\color{fig5red} red} cycles denote the gradient descent operators.
    \textbf{(c)} To reduce memory requirements, nodes related with frozen weights (colored in {\color{fig5gray} light blue}) are pruned from backward computation.
    \textbf{(d)} To minimize memory footprint, the gradient descent operators are re-ordered to be interlaced with backward computations (colored in {\color{fig5yellow} yellow}).
    \textbf{(e)} \engineshort compiles forward and backward graphs using code generation and deploys training on tiny IoT devices (best viewed in colors).
    }
    \label{fig:compiler_stack}
\end{figure*}

\begin{figure*}[t]
    \centering
     \includegraphics[width=\textwidth]{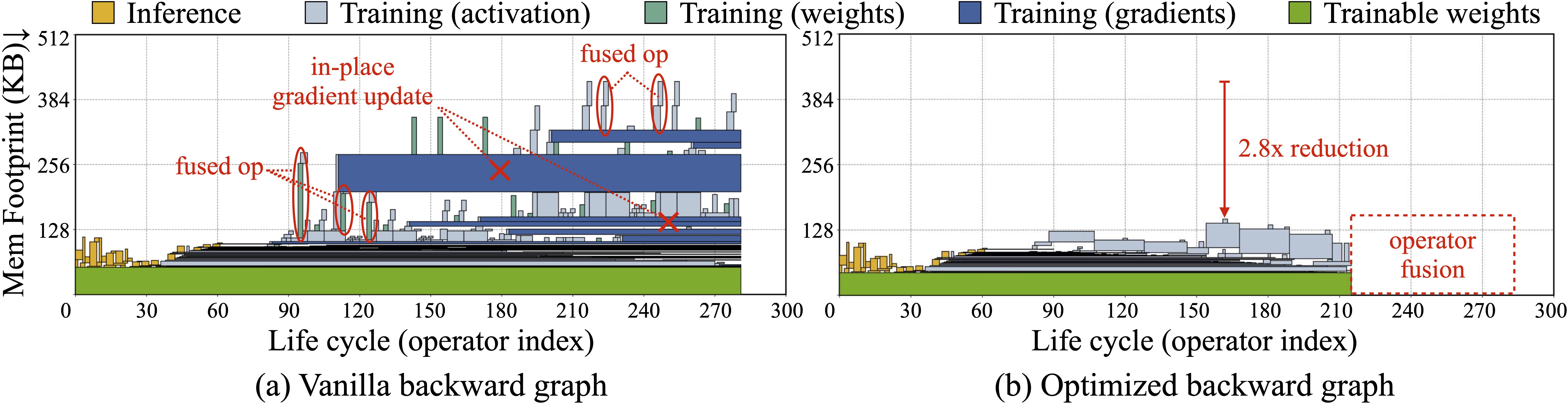}
     \vspace{-15pt}
    \caption{Memory footprint reduction by operator reordering. 
    With operator reordering, \engineshort can apply in-place gradient update and perform operator fusion to avoid large intermediate tensors to reduce memory footprint. We profiled MobileNetV2-w0.35 in this figure (same as Figure~\ref{fig:teaser}).
     }
    \label{fig:mem_footprint}
\end{figure*}

The theoretical saving from real quantized training and sparse update does not translate to measured memory saving in existing deep learning frameworks, due to the redundant runtime and the lack of graph pruning.
MCUNetV3 co-designed an efficient training system, \engine (\engineshort), to transform the above algorithms into slim binary codes (Figure~\ref{fig:compiler_stack}). %

\noindent\textbf{Compile-time Differentiation and Code Generation.}
\engineshort offloads the auto-differentiation from the runtime to the compile-time, generating a static backward graph that can be pruned and optimized (see below) to reduce the memory and computation. \engineshort is based on code generation: it compiles the optimized graphs to executable binaries on the target hardware, which minimizes the runtime library size and removes the need for host languages like Python (typically uses Megabytes of memory).

\begin{figure*}[t]
    \centering
     \includegraphics[width=\textwidth]{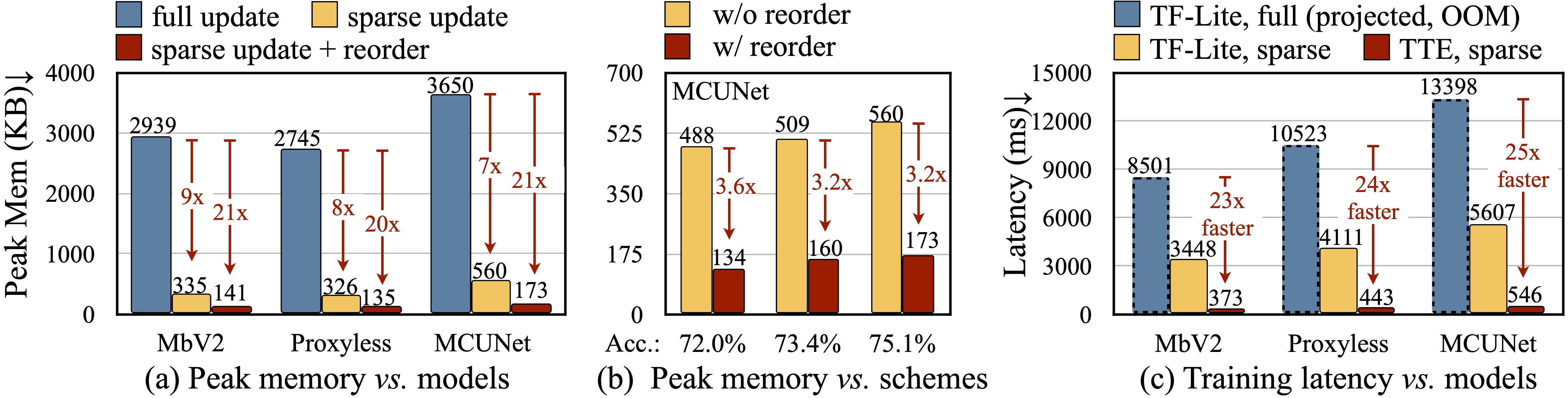}
     \vspace{-15pt}
    \caption{\emph{Measured} peak memory and latency: \textbf{(a)} Sparse update with TTE graph optimization can reduce the measured peak memory by 20-21$\times$ for different models, making training feasible on tiny edge devices.
    \textbf{(b)} Graph optimization consistently reduces the peak memory for different sparse update schemes (denoted by different average transfer learning accuracies).
    \textbf{(c)} Sparse update with TTE operators achieves 23-25$\times$ faster training speed compared to the full update with TF-Lite Micro operators, leading to less energy usage.
    \emph{Note}: for sparse update, we choose the config that achieves the same accuracy as full update.
    }
    \label{fig:latency_peakmem_comparison}
\end{figure*}

\noindent\textbf{Backward Graph Pruning for Sparse Update.}
\engineshort prune the redundant nodes  in the backward graph before compiling it to binary codes.
For sparse layer update, \engineshort prune away the gradient nodes of the frozen weights, only keeping the nodes for bias update. Afterward, \engineshort traverses the graph to find unused intermediate nodes due to pruning (\eg, saved input activation) and apply dead-code elimination (DCE) to remove the redundancy. 
For sparse tensor update, \engineshort introduces a \emph{sub-operator slicing} mechanism to split a layer's weights into trainable and frozen parts; the backward graph of the frozen subset is removed. 
\engineshort's compilation translates the sparse update algorithm into the measured memory saving, reducing the training memory by 7-9$\times$ without losing accuracy (Figure~\ref{fig:latency_peakmem_comparison}(a)).

\noindent\textbf{Operator Reordering and Graph Optimization.}
The execution order of different operations affects the life cycle of tensors and the overall memory footprint. This has been well-studied for inference~\cite{ahn2020ordering, liberis2019neural} but not for training due to the extra complexity.
Traditional training frameworks usually derive the gradients of all the trainable parameters before applying the update. Such a practice leads to significant memory waste for storing the gradients. By reordering operators, the gradient update to a specific tensor can immediately be applied (in-place update) before back-propagating to earlier layers, so that the gradient can be released. As such, \engineshort trace the dependency of all tensors (weights, gradients, activation) and reorder the operators, so that some operators can be fused to reduce memory footprint (by 2.4-3.2$\times$, Figure~\ref{fig:latency_peakmem_comparison}(a)). Figure~\ref{fig:mem_footprint} provides an example to reflect the memory saving from reordering.

\noindent\textbf{Memory Saving \& Faster Training}
Figure~\ref{fig:latency_peakmem_comparison}(a)) shows the training memory of three models on STM32F746 MCU to compare the memory saving from \engineshort.
The sparse update effectively reduces peak memory by 7-9$\times$ compared to the full update thanks to the graph pruning mechanism, while achieving the same or higher transfer learning accuracy (compare the data points connected by arrows in Figure~\ref{fig:acc_vs_mem}). The memory is further reduced with operator reordering, leading to 20-21$\times$ total memory saving. With both techniques, the training of all 3 models fits 256KB SRAM.

The training latency per image on the STM32F746 MCU is measured in Figure~\ref{fig:latency_peakmem_comparison}(c). 
By graph optimization and exploiting multiple compiler optimization approaches (such as loop unrolling and tiling), our sparse update + TTE kernels can significantly enhance the training speed by 23-25$\times$ compared to the full update + TF-Lite Micro kernels, leading to energy saving and making training practical. 

\section{Conclusion and Outlook}
\label{tiny_future}
In conclusion, TinyML is a rapidly evolving field that enables deep learning on resource-constrained devices. It fosters a wide range of customized and private AI applications on edge devices, which can process the data collected from the sensors right at the source. 
We point out several unique challenges of TinyML. First, we need to redesign the model design space since deep models designed for mobile and other platforms do not work well for TinyML. Second, we need to redesign backpropagation schemes and investigate new learning algorithms since directly adapting models for inference does not work for tiny training. Third, co-design is necessary for TinyML.
We summarize the related works aiming to overcome the challenges from the algorithm and the system perspectives.
Furthermore, we introduce the TinyML techniques that not only enable practical AI applications on a wide range of IoT platforms for inference, but also allow AI to be continuously trained over time, adapting to a world that is changing fast.
Looking to the future, TinyML will continue to be an active and rapidly growing area, which requires continued efforts to improve the performance and energy efficiency. We discuss several possible directions for the future development of TinyML.

\textbf{More applications and modalities.} 
This review mainly focuses on convolutional neural networks (CNNs) as computer vision is widely adapted to tiny devices. However, TinyML has a broad range of applications beyond computer vision, including but not limited to audio processing, language processing, anomaly detection, \etc, with sensor inputs from temperature/humidity sensors, accelerometers, current/voltage sensors, among others. 
TinyML enables local devices to process multiple-sensor inputs to handle multi-task workloads, opening up future avenues for numerous potential applications. We will leave further exploration of these possibilities for future work.

\textbf{Self-supervised learning.}
Obtaining accurately labeled data for on-device learning on the edge can be challenging. In some cases, like keyboard typing, we can use the next input word as the prediction target for the model. However, this is not always practical for most applications, such as domain adaptation for vision tasks (e.g., segmentation, detection), obtaining supervision can be expensive and difficult. One potential solution is to design self-supervised learning tasks for on-device training, as has been proposed in recent research~\cite{sun2020test}.

\textbf{Relationship between TinyML and LargeML.}
TinyML and LargeML both aim to develop efficient models under resource constraints such as memory, computation, engineering effort, and data. While TinyML is primarily focusing on making models run efficiently on small devices, many of its techniques can also be applied in cloud environments for large-scale machine learning scenarios. For example, quantization techniques have been effective in both TinyML~\cite{lin2020mcunet, lin2021mcunetv2} and LargeML settings~\cite{dettmers2022llm, xiao2022smoothquant}, and the concept of sparse learning has been used in both scenarios to run models efficiently with limited resources~\cite{google2022palm, lin2022ondevice_mcunetv3}. These efficient techniques are generally applicable and should not be limited to TinyML settings.

The concept of TinyML is constantly evolving and expanding. When ResNet-50~\cite{he2016deep} was first introduced in 2016, it was considered as a large model with $25M$ parameters and $4G$ MACs. However, 6 years later, with the rapid advances in hardware, it can now achieve sub-millisecond inference on a smartphone DSP (Qualcomm Snapdragon 8Gen1).
As hardware continues to improve, what was once considered a ``large'' model may be considered ``tiny'' in the future. The scope of TinyML should evolve and adapt over time.


\section{Acknowledgement}
We thank MIT AI Hardware Program, National Science Foundation, NVIDIA Academic Partnership Award, MIT-IBM Watson AI Lab, Amazon and MIT Science Hub, Qualcomm Innovation Fellowship, Microsoft Turing Academic Program for supporting this research.




\bibliographystyle{IEEEtran}
\bibliography{main}

\begin{thebibliography}{100}
\providecommand{\url}[1]{#1}
\csname url@samestyle\endcsname
\providecommand{\newblock}{\relax}
\providecommand{\bibinfo}[2]{#2}
\providecommand{\BIBentrySTDinterwordspacing}{\spaceskip=0pt\relax}
\providecommand{\BIBentryALTinterwordstretchfactor}{4}
\providecommand{\BIBentryALTinterwordspacing}{\spaceskip=\fontdimen2\font plus
\BIBentryALTinterwordstretchfactor\fontdimen3\font minus
  \fontdimen4\font\relax}
\providecommand{\BIBforeignlanguage}[2]{{%
\expandafter\ifx\csname l@#1\endcsname\relax
\typeout{** WARNING: IEEEtran.bst: No hyphenation pattern has been}%
\typeout{** loaded for the language `#1'. Using the pattern for}%
\typeout{** the default language instead.}%
\else
\language=\csname l@#1\endcsname
\fi
#2}}
\providecommand{\BIBdecl}{\relax}
\BIBdecl

\bibitem{liu2020datamix}
Z.~Liu, Z.~Wu, C.~Gan, L.~Zhu, and S.~Han, ``Datamix: Efficient
  privacy-preserving edge-cloud inference,'' in \emph{European Conference on
  Computer Vision}.\hskip 1em plus 0.5em minus 0.4em\relax Springer, 2020, pp.
  578--595.

\bibitem{singh2019detailed}
A.~Singh, P.~Vepakomma, O.~Gupta, and R.~Raskar, ``Detailed comparison of
  communication efficiency of split learning and federated learning,''
  \emph{arXiv preprint arXiv:1909.09145}, 2019.

\bibitem{konevcny2016federated}
J.~Kone{\v{c}}n{\`y}, H.~B. McMahan, F.~X. Yu, P.~Richt{\'a}rik, A.~T. Suresh,
  and D.~Bacon, ``Federated learning: Strategies for improving communication
  efficiency,'' \emph{arXiv preprint arXiv:1610.05492}, 2016.

\bibitem{sandler2018mobilenetv2}
M.~Sandler, A.~Howard, M.~Zhu, A.~Zhmoginov, and L.-C. Chen, ``Mobilenetv2:
  Inverted residuals and linear bottlenecks,'' in \emph{Conference on Computer
  Vision and Pattern Recognition (CVPR)}, 2018.

\bibitem{howard2017mobilenets}
A.~G. Howard, M.~Zhu, B.~Chen, D.~Kalenichenko, W.~Wang, T.~Weyand,
  M.~Andreetto, and H.~Adam, ``Mobilenets: Efficient convolutional neural
  networks for mobile vision applications,'' \emph{arXiv preprint
  arXiv:1704.04861}, 2017.

\bibitem{ma2018shufflenet}
N.~Ma, X.~Zhang, H.-T. Zheng, and J.~Sun, ``Shufflenet v2: Practical guidelines
  for efficient cnn architecture design,'' in \emph{ECCV}, 2018.

\bibitem{zhang2018shufflenet}
X.~Zhang, X.~Zhou, M.~Lin, and J.~Sun, ``Shufflenet: An extremely efficient
  convolutional neural network for mobile devices,'' in \emph{CVPR}, 2018.

\bibitem{lin2020mcunet}
J.~Lin, W.-M. Chen, Y.~Lin, J.~Cohn, C.~Gan, and S.~Han, ``Mcunet: Tiny deep
  learning on iot devices,'' in \emph{NeurIPS}, 2020.

\bibitem{lin2021mcunetv2}
J.~Lin, W.-M. Chen, H.~Cai, C.~Gan, and S.~Han, ``Mcunetv2: Memory-efficient
  patch-based inference for tiny deep learning,'' \emph{arXiv preprint
  arXiv:2110.15352}, 2021.

\bibitem{he2016deep}
K.~He, X.~Zhang, S.~Ren, and J.~Sun, ``Deep residual learning for image
  recognition,'' in \emph{Conference on Computer Vision and Pattern Recognition
  (CVPR)}, 2016.

\bibitem{tsoukas2021healthcare}
V.~Tsoukas, E.~Boumpa, G.~Giannakas, and A.~Kakarountas, ``A review of machine
  learning and tinyml in healthcare,'' in \emph{25th Pan-Hellenic Conference on
  Informatics}, 2021, pp. 69--73.

\bibitem{rana2022cough}
A.~Rana, Y.~Dhiman, and R.~Anand, ``Cough detection system using tinyml,'' in
  \emph{2022 International Conference on Computing, Communication and Power
  Technology (IC3P)}.\hskip 1em plus 0.5em minus 0.4em\relax IEEE, 2022, pp.
  119--122.

\bibitem{dsouza2022healthcare}
\BIBentryALTinterwordspacing
O.~D’Souza, S.~C. Mukhopadhyay, and M.~Sheng, ``Health, security and fire
  safety process optimisation using intelligence at the edge,'' \emph{Sensors},
  vol.~22, no.~21, 2022. [Online]. Available:
  \url{https://www.mdpi.com/1424-8220/22/21/8143}
\BIBentrySTDinterwordspacing

\bibitem{shumba2022healthcare}
\BIBentryALTinterwordspacing
A.-T. Shumba, T.~Montanaro, I.~Sergi, L.~Fachechi, M.~De~Vittorio, and
  L.~Patrono, ``Leveraging iot-aware technologies and ai techniques for
  real-time critical healthcare applications,'' \emph{Sensors}, vol.~22,
  no.~19, 2022. [Online]. Available:
  \url{https://www.mdpi.com/1424-8220/22/19/7675}
\BIBentrySTDinterwordspacing

\bibitem{vuletic2021healthcare}
M.~Vuletic, V.~Mujagic, N.~Milojevic, and D.~Biswas, ``Edge ai framework for
  healthcare applications.''

\bibitem{wong2020tinyspeech}
A.~Wong, M.~Famouri, M.~Pavlova, and S.~Surana, ``Tinyspeech: Attention
  condensers for deep speech recognition neural networks on edge devices,''
  \emph{arXiv preprint arXiv:2008.04245}, 2020.

\bibitem{mazumder2021kws}
M.~Mazumder, C.~Banbury, J.~Meyer, P.~Warden, and V.~J. Reddi, ``Few-shot
  keyword spotting in any language,'' \emph{arXiv preprint arXiv:2104.01454},
  2021.

\bibitem{hardy2021voice}
E.~Hardy and F.~Badets, ``An ultra-low power rnn classifier for always-on voice
  wake-up detection robust to real-world scenarios,'' \emph{arXiv preprint
  arXiv:2103.04792}, 2021.

\bibitem{lu2020edgecameras}
C.-H. Lu and X.-Z. Lin, ``Toward direct edge-to-edge transfer learning for
  iot-enabled edge cameras,'' \emph{IEEE Internet of Things Journal}, vol.~8,
  no.~6, pp. 4931--4943, 2020.

\bibitem{giordano2020facedetection}
M.~Giordano, P.~Mayer, and M.~Magno, ``A battery-free long-range wireless smart
  camera for face detection,'' in \emph{Proceedings of the 8th International
  Workshop on Energy Harvesting and Energy-Neutral Sensing Systems}, 2020, pp.
  29--35.

\bibitem{luukkonen2021cough}
T.~Luukkonen, A.~Colley, T.~Sepp{\"a}nen, and J.~H{\"a}kkil{\"a}, ``Cough
  activated dynamic face visor,'' in \emph{Augmented Humans Conference 2021},
  2021, pp. 295--297.

\bibitem{mohan2021facemaskdetection}
P.~Mohan, A.~J. Paul, and A.~Chirania, ``A tiny cnn architecture for medical
  face mask detection for resource-constrained endpoints,'' in
  \emph{Innovations in Electrical and Electronic Engineering}.\hskip 1em plus
  0.5em minus 0.4em\relax Springer, 2021, pp. 657--670.

\bibitem{wong2020attendnets}
A.~Wong, M.~Famouri, and M.~J. Shafiee, ``Attendnets: tiny deep image
  recognition neural networks for the edge via visual attention condensers,''
  \emph{arXiv preprint arXiv:2009.14385}, 2020.

\bibitem{benatti2019handgesture}
S.~Benatti, F.~Montagna, V.~Kartsch, A.~Rahimi, D.~Rossi, and L.~Benini,
  ``Online learning and classification of emg-based gestures on a parallel
  ultra-low power platform using hyperdimensional computing,'' \emph{IEEE
  transactions on biomedical circuits and systems}, vol.~13, no.~3, pp.
  516--528, 2019.

\bibitem{moin2021handgesture}
A.~Moin, A.~Zhou, A.~Rahimi, A.~Menon, S.~Benatti, G.~Alexandrov, S.~Tamakloe,
  J.~Ting, N.~Yamamoto, Y.~Khan \emph{et~al.}, ``A wearable biosensing system
  with in-sensor adaptive machine learning for hand gesture recognition,''
  \emph{Nature Electronics}, vol.~4, no.~1, pp. 54--63, 2021.

\bibitem{zhou2021handgesture}
A.~Zhou, R.~Muller, and J.~Rabaey, ``Memory-efficient, limb position-aware hand
  gesture recognition using hyperdimensional computing,'' \emph{arXiv preprint
  arXiv:2103.05267}, 2021.

\bibitem{bian2021handgesture}
S.~Bian and P.~Lukowicz, ``Capacitive sensing based on-board hand gesture
  recognition with tinyml,'' in \emph{Adjunct Proceedings of the 2021 ACM
  International Joint Conference on Pervasive and Ubiquitous Computing and
  Proceedings of the 2021 ACM International Symposium on Wearable Computers},
  2021, pp. 4--5.

\bibitem{paul2020signlanguage}
A.~J. Paul, P.~Mohan, and S.~Sehgal, ``Rethinking generalization in american
  sign language prediction for edge devices with extremely low memory
  footprint,'' in \emph{2020 IEEE Recent Advances in Intelligent Computational
  Systems (RAICS)}.\hskip 1em plus 0.5em minus 0.4em\relax IEEE, 2020, pp.
  147--152.

\bibitem{de2021autonomous}
M.~de~Prado, M.~Rusci, A.~Capotondi, R.~Donze, L.~Benini, and N.~Pazos,
  ``Robustifying the deployment of tinyml models for autonomous
  mini-vehicles,'' \emph{Sensors}, vol.~21, no.~4, p. 1339, 2021.

\bibitem{roshan2021autonomous}
A.~N. Roshan, B.~Gokulapriyan, C.~Siddarth, and P.~Kokil, ``Adaptive traffic
  control with tinyml,'' in \emph{2021 Sixth International Conference on
  Wireless Communications, Signal Processing and Networking (WiSPNET)}.\hskip
  1em plus 0.5em minus 0.4em\relax IEEE, 2021, pp. 451--455.

\bibitem{bao2021vehicularIoT}
W.~Bao, C.~Wu, S.~Guleng, J.~Zhang, K.-L.~A. Yau, and Y.~Ji, ``Edge
  computing-based joint client selection and networking scheme for federated
  learning in vehicular iot,'' \emph{China Communications}, vol.~18, no.~6, pp.
  39--52, 2021.

\bibitem{ying2021manufacturing}
J.~Ying, J.~Hsieh, D.~Hou, J.~Hou, T.~Liu, X.~Zhang, Y.~Wang, and Y.-T. Pan,
  ``Edge-enabled cloud computing management platform for smart manufacturing,''
  in \emph{2021 IEEE International Workshop on Metrology for Industry 4.0 \&
  IoT (MetroInd4.0\&IoT)}, 2021, pp. 682--686.

\bibitem{siang2021anomaly}
Y.~Y. Siang, M.~R. Ahamd, and M.~S.~Z. Abidin, ``Anomaly detection based on
  tiny machine learning: A review,'' \emph{Open International Journal of
  Informatics}, vol.~9, no. Special Issue 2, pp. 67--78, 2021.

\bibitem{roth2022anomaly}
K.~Roth, L.~Pemula, J.~Zepeda, B.~Sch{\"o}lkopf, T.~Brox, and P.~Gehler,
  ``Towards total recall in industrial anomaly detection,'' in
  \emph{Proceedings of the IEEE/CVF Conference on Computer Vision and Pattern
  Recognition}, 2022, pp. 14\,318--14\,328.

\bibitem{alongi2020environment}
F.~Alongi, N.~Ghielmetti, D.~Pau, F.~Terraneo, and W.~Fornaciari, ``Tiny neural
  networks for environmental predictions: an integrated approach with miosix,''
  in \emph{2020 IEEE International Conference on Smart Computing
  (SMARTCOMP)}.\hskip 1em plus 0.5em minus 0.4em\relax IEEE, 2020, pp.
  350--355.

\bibitem{vuppalapati2020agriculture1}
C.~Vuppalapati, A.~Ilapakurti, K.~Chillara, S.~Kedari, and V.~Mamidi,
  ``Automating tiny ml intelligent sensors devops using microsoft azure,'' in
  \emph{2020 ieee international conference on big data (big data)}.\hskip 1em
  plus 0.5em minus 0.4em\relax IEEE, 2020, pp. 2375--2384.

\bibitem{vuppalapati2020agriculture2}
C.~Vuppalapati, A.~Ilapakurti, S.~Kedari, J.~Vuppalapati, S.~Kedari, and
  R.~Vuppalapati, ``Democratization of ai, albeit constrained iot devices \&
  tiny ml, for creating a sustainable food future,'' in \emph{2020 3rd
  International Conference on Information and Computer Technologies
  (ICICT)}.\hskip 1em plus 0.5em minus 0.4em\relax IEEE, 2020, pp. 525--530.

\bibitem{nakhle2021phenomics}
F.~Nakhle and A.~L. Harfouche, ``Ready, steady, go ai: A practical tutorial on
  fundamentals of artificial intelligence and its applications in phenomics
  image analysis,'' \emph{Patterns}, vol.~2, no.~9, p. 100323, 2021.

\bibitem{curnick2022smallsats}
D.~J. Curnick, A.~J. Davies, C.~Duncan, R.~Freeman, D.~M. Jacoby, H.~T.
  Shelley, C.~Rossi, O.~R. Wearn, M.~J. Williamson, and N.~Pettorelli,
  ``Smallsats: a new technological frontier in ecology and conservation?''
  \emph{Remote Sensing in Ecology and Conservation}, vol.~8, no.~2, pp.
  139--150, 2022.

\bibitem{nicolas2022agriculture}
C.~Nicolas, B.~Naila, and R.-C. Amar, ``Tinyml smart sensor for energy saving
  in internet of things precision agriculture platform,'' in \emph{2022
  Thirteenth International Conference on Ubiquitous and Future Networks
  (ICUFN)}.\hskip 1em plus 0.5em minus 0.4em\relax IEEE, 2022, pp. 256--259.

\bibitem{lai2018cmsis}
L.~Lai, N.~Suda, and V.~Chandra, ``Cmsis-nn: Efficient neural network kernels
  for arm cortex-m cpus,'' \emph{arXiv preprint arXiv:1801.06601}, 2018.

\bibitem{X-Cube-AI}
STMicroelectronics, ``X-cube-ai: Ai expansion pack for stm32cubemx,''
  \url{https://www.st.com/en/embedded-software/x-cube-ai.html}.

\bibitem{microTVM}
``microtvm: Tvm on bare-metal,''
  \url{https://tvm.apache.org/docs/topic/microtvm/index.html}.

\bibitem{liberis2019neural}
E.~Liberis and N.~D. Lane, ``Neural networks on microcontrollers: saving memory
  at inference via operator reordering,'' \emph{arXiv preprint
  arXiv:1910.05110}, 2019.

\bibitem{rusci2019memory}
M.~Rusci, A.~Capotondi, and L.~Benini, ``Memory-driven mixed low precision
  quantization for enabling deep network inference on microcontrollers,'' in
  \emph{MLSys}, 2020.

\bibitem{capotondi2020cmix}
A.~Capotondi, M.~Rusci, M.~Fariselli, and L.~Benini, ``Cmix-nn: Mixed
  low-precision cnn library for memory-constrained edge devices,'' \emph{IEEE
  Transactions on Circuits and Systems II: Express Briefs}, vol.~67, no.~5, pp.
  871--875, 2020.

\bibitem{david2021_tflitemicro}
R.~David, J.~Duke, A.~Jain, V.~Janapa~Reddi, N.~Jeffries, J.~Li, N.~Kreeger,
  I.~Nappier, M.~Natraj, T.~Wang, P.~Warden, and R.~Rhodes, ``Tensorflow lite
  micro: Embedded machine learning for tinyml systems,'' in \emph{Proceedings
  of Machine Learning and Systems}, vol.~3, 2021, pp. 800--811.

\bibitem{banbury2021micronets}
C.~Banbury, C.~Zhou, I.~Fedorov, R.~Matas, U.~Thakker, D.~Gope,
  V.~Janapa~Reddi, M.~Mattina, and P.~Whatmough, ``Micronets: Neural network
  architectures for deploying tinyml applications on commodity
  microcontrollers,'' \emph{Proceedings of Machine Learning and Systems},
  vol.~3, 2021.

\bibitem{Sadiq2022TinyOps}
S.~Sadiq, J.~Hare, P.~Maji, S.~Craske, and G.~V. Merrett, ``Tinyops: Imagenet
  scale deep learning on microcontrollers,'' in \emph{2022 IEEE/CVF Conference
  on Computer Vision and Pattern Recognition Workshops (CVPRW)}, 2022, pp.
  2701--2705.

\bibitem{TinyMaix}
``Tinymaix,'' \url{https://github.com/sipeed/TinyMaix}.

\bibitem{fedorov2022udc}
I.~Fedorov, R.~Matas, H.~Tann, C.~Zhou, M.~Mattina, and P.~Whatmough, ``{UDC}:
  Unified {DNAS} for compressible tiny{ML} models for neural processing
  units,'' in \emph{Advances in Neural Information Processing Systems}, 2022.

\bibitem{cai2020tinytl}
H.~Cai, C.~Gan, L.~Zhu, and S.~Han, ``Tinytl: Reduce activations, not trainable
  parameters for efficient on-device learning,'' \emph{arXiv preprint
  arXiv:2007.11622}, 2020.

\bibitem{ren2021tinyol}
H.~Ren, D.~Anicic, and T.~A. Runkler, ``Tinyol: Tinyml with online-learning on
  microcontrollers,'' in \emph{2021 International Joint Conference on Neural
  Networks (IJCNN)}.\hskip 1em plus 0.5em minus 0.4em\relax IEEE, 2021, pp.
  1--8.

\bibitem{patil2022poet}
S.~G. Patil, P.~Jain, P.~Dutta, I.~Stoica, and J.~Gonzalez, ``Poet: Training
  neural networks on tiny devices with integrated rematerialization and
  paging,'' in \emph{International Conference on Machine Learning}.\hskip 1em
  plus 0.5em minus 0.4em\relax PMLR, 2022, pp. 17\,573--17\,583.

\bibitem{profentzas2022minilearn}
C.~Profentzas, M.~Almgren, and O.~Landsiedel, ``Minilearn: On-device learning
  for low-power iot devices,'' in \emph{Proceedings of the 2022 International
  Conference on Embedded Wireless Systems and Networks (Linz,
  Austria)(EWSN’22). Junction Publishing, USA}, 2022.

\bibitem{lin2022ondevice_mcunetv3}
J.~Lin, L.~Zhu, W.-M. Chen, W.-C. Wang, C.~Gan, and S.~Han, ``On-device
  training under 256kb memory,'' 2022.

\bibitem{han2015learning}
S.~Han, J.~Pool, J.~Tran, and W.~Dally, ``Learning both weights and connections
  for efficient neural network,'' in \emph{NeurIPS}, 2015.

\bibitem{he2017channel}
Y.~He, X.~Zhang, and J.~Sun, ``Channel pruning for accelerating very deep
  neural networks,'' in \emph{ICCV}, 2017.

\bibitem{lin2017runtime}
J.~Lin, Y.~Rao, J.~Lu, and J.~Zhou, ``Runtime neural pruning,'' in
  \emph{NeurIPS}, 2017.

\bibitem{liu2017learning}
Z.~Liu, J.~Li, Z.~Shen, G.~Huang, S.~Yan, and C.~Zhang, ``Learning efficient
  convolutional networks through network slimming,'' in \emph{ICCV}, 2017.

\bibitem{he2018amc}
Y.~He, J.~Lin, Z.~Liu, H.~Wang, L.-J. Li, and S.~Han, ``Amc: Automl for model
  compression and acceleration on mobile devices,'' in \emph{ECCV}, 2018.

\bibitem{liu2019metapruning}
Z.~Liu, H.~Mu, X.~Zhang, Z.~Guo, X.~Yang, K.-T. Cheng, and J.~Sun,
  ``{MetaPruning: Meta Learning for Automatic Neural Network Channel
  Pruning},'' in \emph{ICCV}, 2019.

\bibitem{han2016deep}
S.~Han, H.~Mao, and W.~J. Dally, ``Deep compression: Compressing deep neural
  networks with pruning, trained quantization and huffman coding,'' in
  \emph{ICLR}, 2016.

\bibitem{zhu2016trained}
C.~Zhu, S.~Han, H.~Mao, and W.~J. Dally, ``Trained ternary quantization,'' in
  \emph{ICLR}, 2017.

\bibitem{rastegari2016xnor}
M.~Rastegari, V.~Ordonez, J.~Redmon, and A.~Farhadi, ``Xnor-net: Imagenet
  classification using binary convolutional neural networks,'' in \emph{ECCV},
  2016.

\bibitem{zhou2016dorefa}
S.~Zhou, Y.~Wu, Z.~Ni, X.~Zhou, H.~Wen, and Y.~Zou, ``Dorefa-net: Training low
  bitwidth convolutional neural networks with low bitwidth gradients,''
  \emph{arXiv preprint arXiv:1606.06160}, 2016.

\bibitem{courbariaux2016binarynet}
M.~Courbariaux and Y.~Bengio, ``Binarynet: Training deep neural networks with
  weights and activations constrained to+ 1 or-1,'' \emph{arXiv preprint
  arXiv:1602.02830}, 2016.

\bibitem{choi2018pact}
J.~Choi, Z.~Wang, S.~Venkataramani, P.~I.-J. Chuang, V.~Srinivasan, and
  K.~Gopalakrishnan, ``Pact: Parameterized clipping activation for quantized
  neural networks,'' \emph{arXiv preprint arXiv:1805.06085}, 2018.

\bibitem{wang2019haq}
K.~Wang, Z.~Liu, Y.~Lin, J.~Lin, and S.~Han, ``{HAQ: Hardware-Aware Automated
  Quantization with Mixed Precision},'' in \emph{CVPR}, 2019.

\bibitem{langroudi2021tent}
H.~F. Langroudi, V.~Karia, T.~Pandit, and D.~Kudithipudi, ``Tent: Efficient
  quantization of neural networks on the tiny edge with tapered fixed point,''
  \emph{arXiv preprint arXiv:2104.02233}, 2021.

\bibitem{lebedev2014speeding}
V.~Lebedev, Y.~Ganin, M.~Rakhuba, I.~Oseledets, and V.~Lempitsky, ``Speeding-up
  convolutional neural networks using fine-tuned cp-decomposition,''
  \emph{arXiv preprint arXiv:1412.6553}, 2014.

\bibitem{gong2014compressing}
Y.~Gong, L.~Liu, M.~Yang, and L.~Bourdev, ``Compressing deep convolutional
  networks using vector quantization,'' \emph{arXiv preprint arXiv:1412.6115},
  2014.

\bibitem{kim2015compression}
Y.-D. Kim, E.~Park, S.~Yoo, T.~Choi, L.~Yang, and D.~Shin, ``Compression of
  deep convolutional neural networks for fast and low power mobile
  applications,'' \emph{arXiv preprint arXiv:1511.06530}, 2015.

\bibitem{hinton2015distilling}
G.~Hinton, O.~Vinyals, and J.~Dean, ``Distilling the knowledge in a neural
  network,'' \emph{arXiv preprint arXiv:1503.02531}, 2015.

\bibitem{park2019distillation}
W.~Park, D.~Kim, Y.~Lu, and M.~Cho, ``Relational knowledge distillation,'' in
  \emph{Proceedings of the IEEE/CVF Conference on Computer Vision and Pattern
  Recognition}, 2019, pp. 3967--3976.

\bibitem{tung2019distillation}
F.~Tung and G.~Mori, ``Similarity-preserving knowledge distillation,'' in
  \emph{Proceedings of the IEEE/CVF International Conference on Computer
  Vision}, 2019, pp. 1365--1374.

\bibitem{mirzadeh2020distillation}
S.~I. Mirzadeh, M.~Farajtabar, A.~Li, N.~Levine, A.~Matsukawa, and
  H.~Ghasemzadeh, ``Improved knowledge distillation via teacher assistant,'' in
  \emph{Proceedings of the AAAI conference on artificial intelligence},
  vol.~34, no.~04, 2020, pp. 5191--5198.

\bibitem{wang2021distillation}
L.~Wang and K.-J. Yoon, ``Knowledge distillation and student-teacher learning
  for visual intelligence: A review and new outlooks,'' \emph{IEEE Transactions
  on Pattern Analysis and Machine Intelligence}, 2021.

\bibitem{yang2022distillation}
Z.~Yang, Z.~Li, X.~Jiang, Y.~Gong, Z.~Yuan, D.~Zhao, and C.~Yuan, ``Focal and
  global knowledge distillation for detectors,'' in \emph{Proceedings of the
  IEEE/CVF Conference on Computer Vision and Pattern Recognition}, 2022, pp.
  4643--4652.

\bibitem{zhao2022distillation}
B.~Zhao, Q.~Cui, R.~Song, Y.~Qiu, and J.~Liang, ``Decoupled knowledge
  distillation,'' in \emph{Proceedings of the IEEE/CVF Conference on Computer
  Vision and Pattern Recognition}, 2022, pp. 11\,953--11\,962.

\bibitem{beyer2022distillation}
L.~Beyer, X.~Zhai, A.~Royer, L.~Markeeva, R.~Anil, and A.~Kolesnikov,
  ``Knowledge distillation: A good teacher is patient and consistent,'' in
  \emph{Proceedings of the IEEE/CVF Conference on Computer Vision and Pattern
  Recognition}, 2022, pp. 10\,925--10\,934.

\bibitem{zoph2017neural}
B.~Zoph and Q.~V. Le, ``Neural architecture search with reinforcement
  learning,'' in \emph{ICLR}, 2017.

\bibitem{zoph2018learning}
B.~Zoph, V.~Vasudevan, J.~Shlens, and Q.~V. Le, ``Learning transferable
  architectures for scalable image recognition,'' in \emph{CVPR}, 2018.

\bibitem{liu2019darts}
H.~Liu, K.~Simonyan, and Y.~Yang, ``Darts: Differentiable architecture
  search,'' in \emph{ICLR}, 2019.

\bibitem{cai2019proxylessnas}
\BIBentryALTinterwordspacing
H.~Cai, L.~Zhu, and S.~Han, ``Proxyless{NAS}: Direct neural architecture search
  on target task and hardware,'' in \emph{ICLR}, 2019. [Online]. Available:
  \url{https://arxiv.org/pdf/1812.00332.pdf}
\BIBentrySTDinterwordspacing

\bibitem{tan2019mnasnet}
M.~Tan, B.~Chen, R.~Pang, V.~Vasudevan, M.~Sandler, A.~Howard, and Q.~V. Le,
  ``Mnasnet: Platform-aware neural architecture search for mobile,'' in
  \emph{CVPR}, 2019.

\bibitem{wu2019fbnet}
B.~Wu, X.~Dai, P.~Zhang, Y.~Wang, F.~Sun, Y.~Wu, Y.~Tian, P.~Vajda, Y.~Jia, and
  K.~Keutzer, ``Fbnet: Hardware-aware efficient convnet design via
  differentiable neural architecture search,'' in \emph{CVPR}, 2019.

\bibitem{radosavovic2020designing}
I.~Radosavovic, R.~P. Kosaraju, R.~Girshick, K.~He, and P.~Doll{\'a}r,
  ``Designing network design spaces,'' \emph{arXiv preprint arXiv:2003.13678},
  2020.

\bibitem{pytorch2019}
A.~Paszke, S.~Gross, F.~Massa, A.~Lerer, J.~Bradbury, G.~Chanan, T.~Killeen,
  Z.~Lin, N.~Gimelshein, L.~Antiga \emph{et~al.}, ``Pytorch: An imperative
  style, high-performance deep learning library,'' \emph{Advances in neural
  information processing systems}, vol.~32, 2019.

\bibitem{abadi2016tensorflow}
M.~Abadi, P.~Barham, J.~Chen, Z.~Chen, A.~Davis, J.~Dean, M.~Devin,
  S.~Ghemawat, G.~Irving, M.~Isard \emph{et~al.}, ``Tensorflow: A system for
  large-scale machine learning,'' in \emph{OSDI}, 2016.

\bibitem{chen2015mxnet}
T.~Chen, M.~Li, Y.~Li, M.~Lin, N.~Wang, M.~Wang, T.~Xiao, B.~Xu, C.~Zhang, and
  Z.~Zhang, ``Mxnet: A flexible and efficient machine learning library for
  heterogeneous distributed systems,'' \emph{arXiv preprint arXiv:1512.01274},
  2015.

\bibitem{jax2018github}
\BIBentryALTinterwordspacing
J.~Bradbury, R.~Frostig, P.~Hawkins, M.~J. Johnson, C.~Leary, D.~Maclaurin,
  G.~Necula, A.~Paszke, J.~Vander{P}las, S.~Wanderman-{M}ilne, and Q.~Zhang,
  ``{JAX}: composable transformations of {P}ython+{N}um{P}y programs,'' 2018.
  [Online]. Available: \url{http://github.com/google/jax}
\BIBentrySTDinterwordspacing

\bibitem{chen2018tvm}
T.~Chen, T.~Moreau, Z.~Jiang, L.~Zheng, E.~Yan, H.~Shen, M.~Cowan, L.~Wang,
  Y.~Hu, L.~Ceze \emph{et~al.}, ``$\{$TVM$\}$: An automated end-to-end
  optimizing compiler for deep learning,'' in \emph{OSDI}, 2018.

\bibitem{tflite}
``Tensorflow lite,'' \url{https://www.tensorflow.org/lite}.

\bibitem{jiang2020mnn}
X.~Jiang, H.~Wang, Y.~Chen, Z.~Wu, L.~Wang, B.~Zou, Y.~Yang, Z.~Cui, Y.~Cai,
  T.~Yu, C.~Lv, and Z.~Wu, ``Mnn: A universal and efficient inference engine,''
  in \emph{MLSys}, 2020.

\bibitem{ncnn}
``Ncnn : A high-performance neural network inference computing framework
  optimized for mobile platforms,'' \url{https://github.com/Tencent/ncnn}.

\bibitem{tensorRT}
``Nvidia tensorrt, an sdk for high-performance deep learning inference,''
  \url{https://developer.nvidia.com/tensorrt}.

\bibitem{vaswani2017attention}
A.~Vaswani, N.~Shazeer, N.~Parmar, J.~Uszkoreit, L.~Jones, A.~N. Gomez,
  {\L}.~Kaiser, and I.~Polosukhin, ``Attention is all you need,'' in
  \emph{NeurIPS}, 2017.

\bibitem{chen2018learning}
T.~Chen, L.~Zheng, E.~Yan, Z.~Jiang, T.~Moreau, L.~Ceze, C.~Guestrin, and
  A.~Krishnamurthy, ``Learning to optimize tensor programs,'' in
  \emph{NeurIPS}, 2018.

\bibitem{stoutchinin2019optimally}
A.~Stoutchinin, F.~Conti, and L.~Benini, ``Optimally scheduling cnn
  convolutions for efficient memory access,'' \emph{arXiv preprint
  arXiv:1902.01492}, 2019.

\bibitem{ahn2020ordering}
B.~H. Ahn, J.~Lee, J.~M. Lin, H.-P. Cheng, J.~Hou, and H.~Esmaeilzadeh,
  ``Ordering chaos: Memory-aware scheduling of irregularly wired neural
  networks for edge devices,'' \emph{arXiv preprint arXiv:2003.02369}, 2020.

\bibitem{miao2021enabling}
H.~Miao and F.~X. Lin, ``Enabling large neural networks on tiny
  microcontrollers with swapping,'' \emph{arXiv preprint arXiv:2101.08744},
  2021.

\bibitem{alwani2016fused}
M.~Alwani, H.~Chen, M.~Ferdman, and P.~Milder, ``Fused-layer cnn
  accelerators,'' in \emph{2016 49th Annual IEEE/ACM International Symposium on
  Microarchitecture (MICRO)}.\hskip 1em plus 0.5em minus 0.4em\relax IEEE,
  2016, pp. 1--12.

\bibitem{goetschalckx2019breaking}
K.~Goetschalckx and M.~Verhelst, ``Breaking high-resolution cnn bandwidth
  barriers with enhanced depth-first execution,'' \emph{IEEE Journal on
  Emerging and Selected Topics in Circuits and Systems}, vol.~9, no.~2, pp.
  323--331, 2019.

\bibitem{saha2020rnnpool}
O.~Saha, A.~Kusupati, H.~V. Simhadri, M.~Varma, and P.~Jain, ``Rnnpool:
  Efficient non-linear pooling for ram constrained inference,'' \emph{arXiv
  preprint arXiv:2002.11921}, 2020.

\bibitem{tan2019efficientnet}
M.~Tan and Q.~Le, ``Efficientnet: Rethinking model scaling for convolutional
  neural networks,'' in \emph{International Conferences on Machine Learning
  (ICML)}.\hskip 1em plus 0.5em minus 0.4em\relax PMLR, 2019, pp. 6105--6114.

\bibitem{tan2021efficientnetv2}
\BIBentryALTinterwordspacing
M.~Tan and Q.~V. Le, ``Efficientnetv2: Smaller models and faster training,''
  \emph{CoRR}, vol. abs/2104.00298, 2021. [Online]. Available:
  \url{https://arxiv.org/abs/2104.00298}
\BIBentrySTDinterwordspacing

\bibitem{audrunas2016nips}
A.~Gruslys, R.~Munos, I.~Danihelka, M.~Lanctot, and A.~Graves,
  ``Memory-efficient backpropagation through time,'' in \emph{NeurIPS}, 2016,
  p. 4132–4140.

\bibitem{chen2016training}
T.~Chen, B.~Xu, C.~Zhang, and C.~Guestrin, ``Training deep nets with sublinear
  memory cost,'' \emph{arXiv preprint arXiv:1604.06174}, 2016.

\bibitem{klaus2017iclr}
\BIBentryALTinterwordspacing
K.~Greff, R.~K. Srivastava, and J.~Schmidhuber, ``Highway and residual networks
  learn unrolled iterative estimation,'' in \emph{ICLR}, 2017. [Online].
  Available: \url{https://arxiv.org/pdf/1604.06174.pdf}
\BIBentrySTDinterwordspacing

\bibitem{liu2019dynamic}
L.~Liu, L.~Deng, X.~Hu, M.~Zhu, G.~Li, Y.~Ding, and Y.~Xie, ``Dynamic sparse
  graph for efficient deep learning,'' in \emph{ICLR}, 2019.

\bibitem{wang2019e2train}
\BIBentryALTinterwordspacing
Y.~Wang, Z.~Jiang, X.~Chen, P.~Xu, Y.~Zhao, Y.~Lin, and Z.~Wang, ``E2-train:
  Training state-of-the-art cnns with over 80\% energy savings,'' 2019.
  [Online]. Available: \url{https://arxiv.org/abs/1910.13349}
\BIBentrySTDinterwordspacing

\bibitem{wang2018training}
N.~Wang, J.~Choi, D.~Brand, C.-Y. Chen, and K.~Gopalakrishnan, ``Training deep
  neural networks with 8-bit floating point numbers,'' in \emph{NeurIPS}, 2018.

\bibitem{sun2019hybrid}
X.~Sun, J.~Choi, C.-Y. Chen, N.~Wang, S.~Venkataramani, V.~V. Srinivasan,
  X.~Cui, W.~Zhang, and K.~Gopalakrishnan, ``Hybrid 8-bit floating point (hfp8)
  training and inference for deep neural networks,'' in \emph{NeurIPS}, 2019.

\bibitem{krizhevsky2012imagenet}
A.~Krizhevsky, I.~Sutskever, and G.~E. Hinton, ``Imagenet classification with
  deep convolutional neural networks,'' in \emph{Advances in Neural Information
  Processing Systems (NIPS)}, 2012.

\bibitem{cui2018large}
Y.~Cui, Y.~Song, C.~Sun, A.~Howard, and S.~Belongie, ``Large scale fine-grained
  categorization and domain-specific transfer learning,'' in \emph{CVPR}, 2018.

\bibitem{kornblith2019better}
S.~Kornblith, J.~Shlens, and Q.~V. Le, ``Do better imagenet models transfer
  better?'' in \emph{CVPR}, 2019.

\bibitem{kolesnikov2020big}
A.~Kolesnikov, L.~Beyer, X.~Zhai, J.~Puigcerver, J.~Yung, S.~Gelly, and
  N.~Houlsby, ``Big transfer (bit): General visual representation learning,''
  in \emph{European conference on computer vision}.\hskip 1em plus 0.5em minus
  0.4em\relax Springer, 2020, pp. 491--507.

\bibitem{chatfield2014return}
K.~Chatfield, K.~Simonyan, A.~Vedaldi, and A.~Zisserman, ``Return of the devil
  in the details: Delving deep into convolutional nets,'' in \emph{BMVC}, 2014.

\bibitem{donahue2014decaf}
J.~Donahue, Y.~Jia, O.~Vinyals, J.~Hoffman, N.~Zhang, E.~Tzeng, and T.~Darrell,
  ``Decaf: A deep convolutional activation feature for generic visual
  recognition,'' in \emph{International Conferences on Machine Learning
  (ICML)}, 2014.

\bibitem{gan2015devnet}
C.~Gan, N.~Wang, Y.~Yang, D.-Y. Yeung, and A.~G. Hauptmann, ``{DevNet:} a deep
  event network for multimedia event detection and evidence recounting,'' in
  \emph{CVPR}, 2015, pp. 2568--2577.

\bibitem{sharif2014cnn}
A.~Sharif~Razavian, H.~Azizpour, J.~Sullivan, and S.~Carlsson, ``Cnn features
  off-the-shelf: an astounding baseline for recognition,'' in \emph{CVPR
  Workshops}, 2014.

\bibitem{wang2022melon}
Q.~Wang, M.~Xu, C.~Jin, X.~Dong, J.~Yuan, X.~Jin, G.~Huang, Y.~Liu, and X.~Liu,
  ``Melon: Breaking the memory wall for resource-efficient on-device machine
  learning,'' in \emph{Proceedings of the 20th Annual International Conference
  on Mobile Systems, Applications and Services}, ser. MobiSys '22, 2022, p.
  450–463.

\bibitem{xu2022mandheling}
D.~Xu, M.~Xu, Q.~Wang, S.~Wang, Y.~Ma, K.~Huang, G.~Huang, X.~Jin, and X.~Liu,
  ``Mandheling: Mixed-precision on-device dnn training with dsp offloading,''
  in \emph{Proceedings of the 28th Annual International Conference on Mobile
  Computing And Networking}, ser. MobiCom '22, 2022, p. 214–227.

\bibitem{gim2022trainingonmobile}
I.~Gim and J.~Ko, ``Memory-efficient dnn training on mobile devices,'' in
  \emph{Proceedings of the 20th Annual International Conference on Mobile
  Systems, Applications and Services}, ser. MobiSys '22, 2022, p. 464–476.

\bibitem{frankle2020training}
J.~Frankle, D.~J. Schwab, and A.~S. Morcos, ``Training batchnorm and only
  batchnorm: On the expressive power of random features in cnns,'' \emph{arXiv
  preprint arXiv:2003.00152}, 2020.

\bibitem{mudrakarta2019k}
P.~K. Mudrakarta, M.~Sandler, A.~Zhmoginov, and A.~Howard, ``K for the price of
  1: Parameter efficient multi-task and transfer learning,'' in \emph{ICLR},
  2019.

\bibitem{canziani2016analysis}
A.~Canziani, A.~Paszke, and E.~Culurciello, ``An analysis of deep neural
  network models for practical applications,'' \emph{arXiv preprint
  arXiv:1605.07678}, 2016.

\bibitem{cai2020once}
\BIBentryALTinterwordspacing
H.~Cai, C.~Gan, T.~Wang, Z.~Zhang, and S.~Han, ``Once for all: Train one
  network and specialize it for efficient deployment,'' in \emph{ICLR}, 2020.
  [Online]. Available: \url{https://arxiv.org/pdf/1908.09791.pdf}
\BIBentrySTDinterwordspacing

\bibitem{bender2018understanding}
G.~Bender, P.-J. Kindermans, B.~Zoph, V.~Vasudevan, and Q.~Le, ``Understanding
  and simplifying one-shot architecture search,'' in \emph{International
  Conferences on Machine Learning (ICML)}, 2018.

\bibitem{guo2019single}
Z.~Guo, X.~Zhang, H.~Mu, W.~Heng, Z.~Liu, Y.~Wei, and J.~Sun, ``Single path
  one-shot neural architecture search with uniform sampling,'' \emph{arXiv
  preprint arXiv:1904.00420}, 2019.

\bibitem{simonyan2014very}
K.~Simonyan and A.~Zisserman, ``Very deep convolutional networks for
  large-scale image recognition,'' in \emph{International Conference on
  Learning Representations (ICLR)}, 2015.

\bibitem{howard2019searching}
A.~Howard, M.~Sandler, G.~Chu, L.-C. Chen, B.~Chen, M.~Tan, W.~Wang, Y.~Zhu,
  R.~Pang, V.~Vasudevan \emph{et~al.}, ``Searching for mobilenetv3,'' in
  \emph{ICCV}, 2019.

\bibitem{everingham2010pascal}
M.~Everingham, L.~Van~Gool, C.~K. Williams, J.~Winn, and A.~Zisserman, ``The
  pascal visual object classes (voc) challenge,'' \emph{International journal
  of computer vision}, vol.~88, no.~2, pp. 303--338, 2010.

\bibitem{redmon2018yolov3}
J.~Redmon and A.~Farhadi, ``{YOLOv3: An Incremental Improvement},''
  \emph{arXiv}, 2018.

\bibitem{yang2016wider}
S.~Yang, P.~Luo, C.~C. Loy, and X.~Tang, ``Wider face: A face detection
  benchmark,'' in \emph{IEEE Conference on Computer Vision and Pattern
  Recognition (CVPR)}, 2016.

\bibitem{yoo2019extd}
Y.~Yoo, D.~Han, and S.~Yun, ``Extd: Extremely tiny face detector via iterative
  filter reuse,'' \emph{arXiv preprint arXiv:1906.06579}, 2019.

\bibitem{he2019lffd}
Y.~He, D.~Xu, L.~Wu, M.~Jian, S.~Xiang, and C.~Pan, ``Lffd: A light and fast
  face detector for edge devices,'' \emph{arXiv preprint arXiv:1904.10633},
  2019.

\bibitem{zhao2019real}
X.~Zhao, X.~Liang, C.~Zhao, M.~Tang, and J.~Wang, ``Real-time multi-scale face
  detector on embedded devices,'' \emph{Sensors}, vol.~19, no.~9, p. 2158,
  2019.

\bibitem{zhang2017s3fd}
S.~Zhang, X.~Zhu, Z.~Lei, H.~Shi, X.~Wang, and S.~Z. Li, ``S3fd: Single shot
  scale-invariant face detector,'' in \emph{Proceedings of the IEEE
  international conference on computer vision}, 2017, pp. 192--201.

\bibitem{burrello2021dory}
A.~Burrello, A.~Garofalo, N.~Bruschi, G.~Tagliavini, D.~Rossi, and F.~Conti,
  ``Dory: Automatic end-to-end deployment of real-world dnns on low-cost iot
  mcus,'' \emph{IEEE Transactions on Computers}, vol.~70, no.~8, pp.
  1253--1268, 2021.

\bibitem{liberis2020mu}
E.~Liberis, {\L}.~Dudziak, and N.~D. Lane, ``$\mu$nas: Constrained neural
  architecture search for microcontrollers,'' \emph{arXiv preprint
  arXiv:2010.14246}, 2020.

\bibitem{fedorov2019sparse}
I.~Fedorov, R.~P. Adams, M.~Mattina, and P.~Whatmough, ``Sparse: Sparse
  architecture search for cnns on resource-constrained microcontrollers,'' in
  \emph{NeurIPS}, 2019.

\bibitem{tensorflow2015}
\BIBentryALTinterwordspacing
M.~Abadi, A.~Agarwal, P.~Barham, E.~Brevdo, Z.~Chen, C.~Citro, G.~S. Corrado,
  A.~Davis, J.~Dean, M.~Devin, S.~Ghemawat, I.~Goodfellow, A.~Harp, G.~Irving,
  M.~Isard, Y.~Jia, R.~Jozefowicz, L.~Kaiser, M.~Kudlur, J.~Levenberg,
  D.~Man\'{e}, R.~Monga, S.~Moore, D.~Murray, C.~Olah, M.~Schuster, J.~Shlens,
  B.~Steiner, I.~Sutskever, K.~Talwar, P.~Tucker, V.~Vanhoucke, V.~Vasudevan,
  F.~Vi\'{e}gas, O.~Vinyals, P.~Warden, M.~Wattenberg, M.~Wicke, Y.~Yu, and
  X.~Zheng, ``{TensorFlow}: Large-scale machine learning on heterogeneous
  systems,'' 2015, software available from tensorflow.org. [Online]. Available:
  \url{http://tensorflow.org/}
\BIBentrySTDinterwordspacing

\bibitem{ioffe2015batch}
S.~Ioffe and C.~Szegedy, ``Batch normalization: Accelerating deep network
  training by reducing internal covariate shift,'' in \emph{International
  Conferences on Machine Learning (ICML)}, 2015.

\bibitem{jacob2018quantization}
B.~Jacob, S.~Kligys, B.~Chen, M.~Zhu, M.~Tang, A.~Howard, H.~Adam, and
  D.~Kalenichenko, ``Quantization and training of neural networks for efficient
  integer-arithmetic-only inference,'' in \emph{CVPR}, 2018, pp. 2704--2713.

\bibitem{krizhevsky2009learning}
A.~Krizhevsky and G.~Hinton, ``Learning multiple layers of features from tiny
  images,'' \emph{Master's thesis, Department of Computer Science, University
  of Toronto}, 2009.

\bibitem{kingma2014adam}
D.~P. Kingma and J.~Ba, ``Adam: A method for stochastic optimization,''
  \emph{arXiv preprint arXiv:1412.6980}, 2014.

\bibitem{you2017large}
Y.~You, I.~Gitman, and B.~Ginsburg, ``Large batch training of convolutional
  networks,'' \emph{arXiv preprint arXiv:1708.03888}, 2017.

\bibitem{krause20133d}
J.~Krause, M.~Stark, J.~Deng, and L.~Fei-Fei, ``3d object representations for
  fine-grained categorization,'' in \emph{Proceedings of the IEEE International
  Conference on Computer Vision Workshops}, 2013.

\bibitem{sun2020test}
Y.~Sun, X.~Wang, Z.~Liu, J.~Miller, A.~Efros, and M.~Hardt, ``Test-time
  training with self-supervision for generalization under distribution
  shifts,'' in \emph{International conference on machine learning}.\hskip 1em
  plus 0.5em minus 0.4em\relax PMLR, 2020, pp. 9229--9248.

\bibitem{dettmers2022llm}
T.~Dettmers, M.~Lewis, Y.~Belkada, and L.~Zettlemoyer, ``Llm. int8 (): 8-bit
  matrix multiplication for transformers at scale,'' \emph{arXiv preprint
  arXiv:2208.07339}, 2022.

\bibitem{xiao2022smoothquant}
G.~Xiao, J.~Lin, M.~Seznec, J.~Demouth, and S.~Han, ``Smoothquant: Accurate and
  efficient post-training quantization for large language models,'' \emph{arXiv
  preprint arXiv:2211.10438}, 2022.

\bibitem{google2022palm}
A.~Chowdhery, S.~Narang, J.~Devlin, M.~Bosma, G.~Mishra, A.~Roberts, P.~Barham,
  H.~W. Chung, C.~Sutton, S.~Gehrmann, P.~Schuh, K.~Shi, S.~Tsvyashchenko,
  J.~Maynez, A.~Rao, P.~Barnes, Y.~Tay, N.~Shazeer, V.~Prabhakaran, E.~Reif,
  N.~Du, B.~Hutchinson, R.~Pope, J.~Bradbury, J.~Austin, M.~Isard, G.~Gur-Ari,
  P.~Yin, T.~Duke, A.~Levskaya, S.~Ghemawat, S.~Dev, H.~Michalewski, X.~Garcia,
  V.~Misra, K.~Robinson, L.~Fedus, D.~Zhou, D.~Ippolito, D.~Luan, H.~Lim,
  B.~Zoph, A.~Spiridonov, R.~Sepassi, D.~Dohan, S.~Agrawal, M.~Omernick, A.~M.
  Dai, T.~S. Pillai, M.~Pellat, A.~Lewkowycz, E.~Moreira, R.~Child, O.~Polozov,
  K.~Lee, Z.~Zhou, X.~Wang, B.~Saeta, M.~Diaz, O.~Firat, M.~Catasta, J.~Wei,
  K.~Meier-Hellstern, D.~Eck, J.~Dean, S.~Petrov, and N.~Fiedel, ``Palm:
  Scaling language modeling with pathways,'' in \emph{Machine Learning and
  Systems (MLSys)}, 2022.

\end{thebibliography}


\begin{IEEEbiography}[{\includegraphics[width=1in,height=1.25in,clip,keepaspectratio]{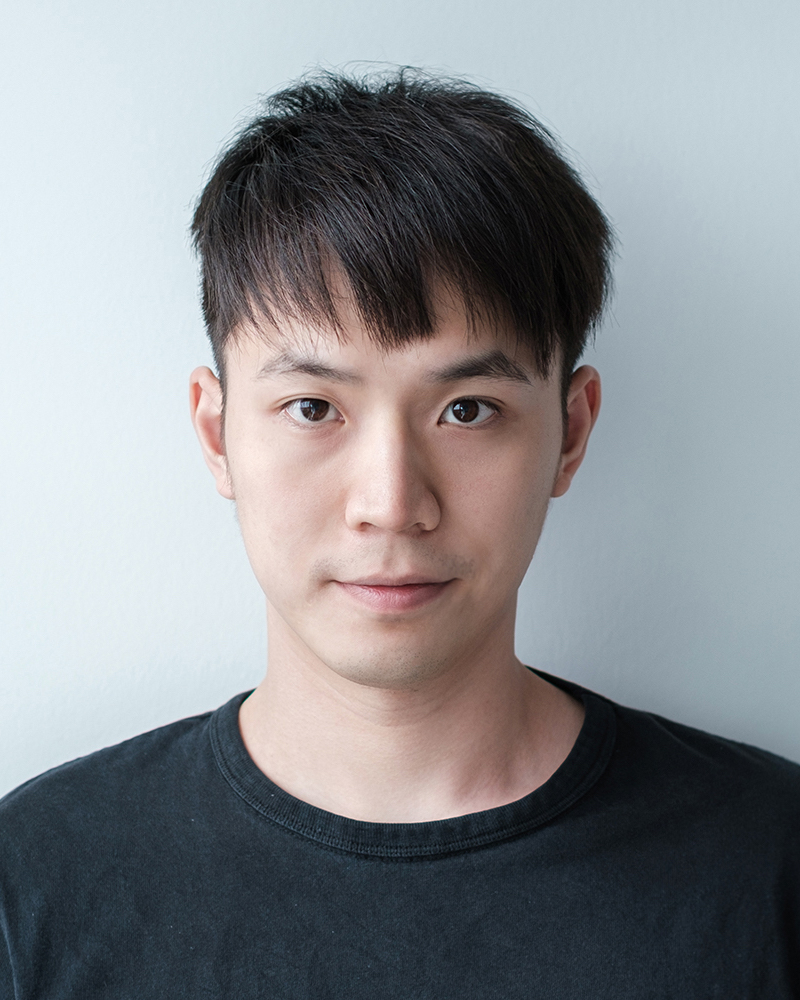}}]{Ji Lin}
is a PhD student at MIT EECS. Previously, he graduated from Department of Electronic Engineering, Tsinghua University. His research interests lie in efficient and hardware-friendly machine learning, model compression and acceleration.
\end{IEEEbiography}

\begin{IEEEbiography}[{\includegraphics[width=1in,height=1.25in,clip,keepaspectratio]{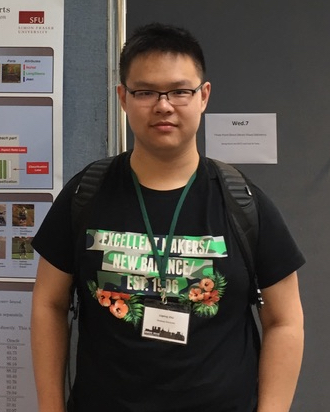}}]{Ligeng Zhu}
is a PhD student at MIT EECS, supervised by Professor Song Han.  His study focuses on efficient and accelerated deep learning systems and algorithms.
\end{IEEEbiography}

\begin{IEEEbiography}[{\includegraphics[width=1in,height=1.25in,clip,keepaspectratio]{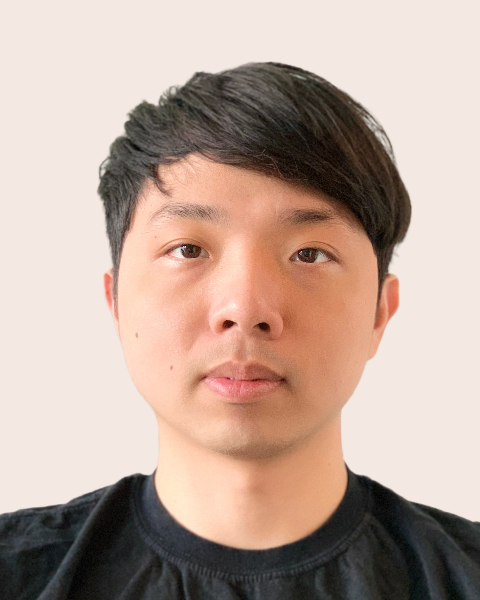}}]{Wei-Ming Chen}
is a Postdoctoral Associate at MIT EECS. Dr. Chen received his Master's and Doctorate degrees in computer science and information engineering from National Taiwan University in 2015 and 2020, respectively. His research interests include TinyML and embedded systems with a focus on enabling efficient deep learning on edge devices.
\end{IEEEbiography}

\begin{IEEEbiography}[{\includegraphics[width=1in,height=1.25in,clip,keepaspectratio]{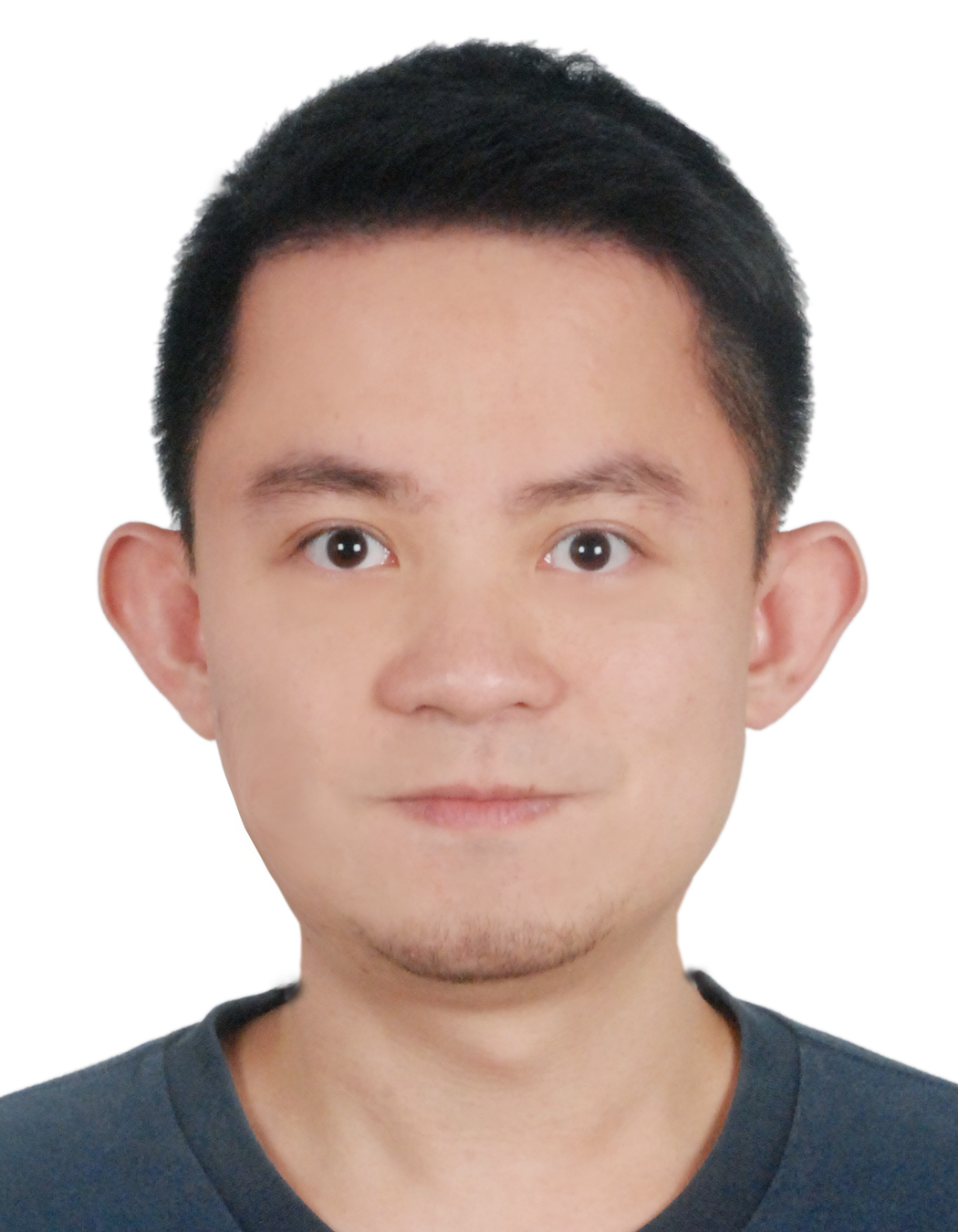}}]{Wei-Chen Wang}
is a Postdoctoral Associate at MIT EECS. Dr. Wang received his Ph.D. degree in Computer Science from the Department of Computer Science and Information Engineering at National Taiwan University in 2021. 
His current research interests include efficient deep learning, model compression, TinyML, and embedded systems.
\end{IEEEbiography}

\begin{IEEEbiography}[{\includegraphics[width=1in,height=1.25in,clip,keepaspectratio]{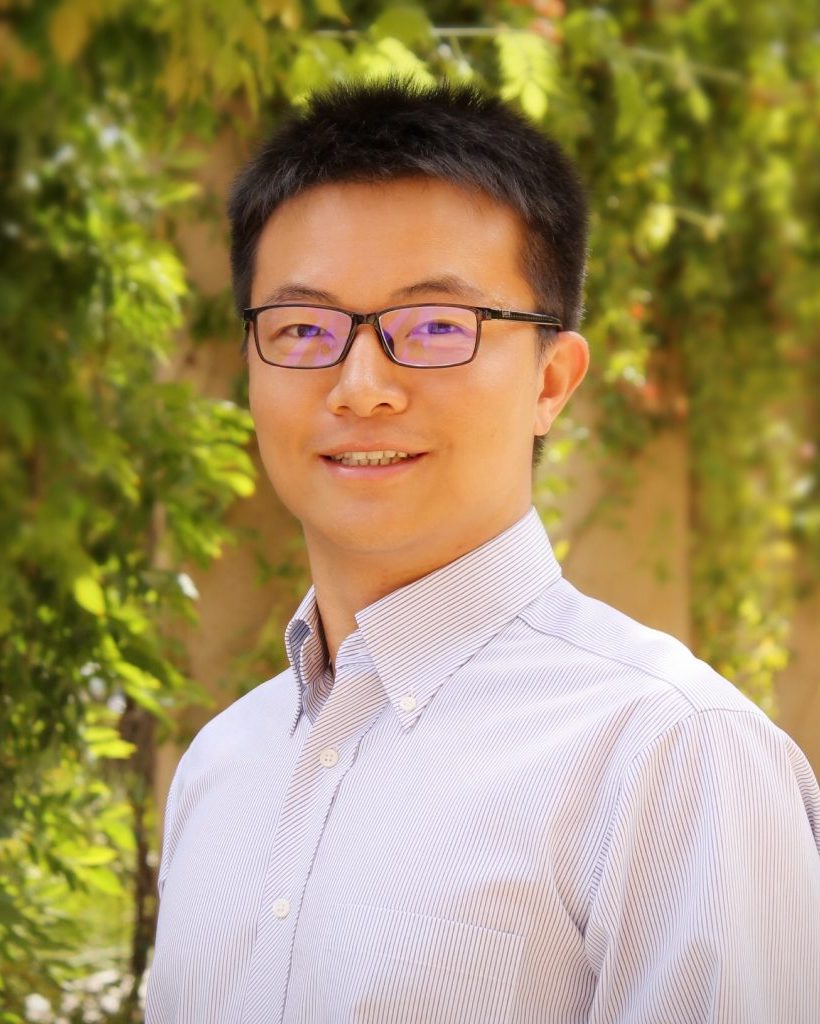}}]{Song Han}
is an associate professor at MIT EECS. Dr. Han received the Ph.D. degree from Stanford University. Dr. Han’s research focuses on efficient deep learning computing at the intersection between machine learning and computer architecture. He proposed ``Deep Compression'' and the ``Efficient Inference Engine'' that widely impacted the industry. He is a recipient of NSF CAREER Award, Sloan Research Fellowship, MIT Technology Review Innovators Under 35, best paper awards at the ICLR and FPGA, and faculty awards from Amazon, Facebook, NVIDIA, Samsung and SONY.
\end{IEEEbiography}

\vfill

\end{document}